# Simultaneous Clustering and Estimation of Heterogeneous Graphical Models

Botao Hao[*], Will Wei Sun[†], Yufeng Liu[‡], Guang Cheng[§]

**Abstract**

We consider joint estimation of multiple graphical models arising from heterogeneous and high-dimensional observations. Unlike most previous approaches which assume that the cluster structure is given in advance, an appealing feature of our method is to learn cluster structure while estimating heterogeneous graphical models. This is achieved via a high dimensional version of Expectation Conditional Maximization (ECM) algorithm (Meng and Rubin, 1993). A joint graphical lasso penalty is imposed on the conditional maximization step to extract both homogeneity and heterogeneity components across all clusters. Our algorithm is computationally efficient due to fast sparse learning routines and can be implemented without unsupervised learning knowledge. The superior performance of our method is demonstrated by extensive experiments and its application to a Glioblastoma cancer dataset reveals some new insights in understanding the Glioblastoma cancer. In theory, a non-asymptotic error bound is established for the output directly from our high dimensional ECM algorithm, and it consists of two quantities: *statistical error* (statistical accuracy) and *optimization error* (computational complexity). Such a result gives a theoretical guideline in terminating our ECM iterations.

[*]Ph.D student, Department of Statistics, Purdue University, West Lafayette, IN 47906. E-mail: hao22@purdue.edu.

[†]Assistant Professor, Department of Management Science, University of Miami, Coral Gables, FL 33146. Email: wsun@bus.miami.edu.

[‡]Professor, Department of Statistics and Operations Research, Department of Genetics, Department of Biostatistics, Carolina Center for Genome Sciences, Lineberger Comprehensive Cancer Center, University of North Carolina at Chapel Hill, NC 27599. E-mail: yfliu@email.unc.edu. Research Sponsored in part by NIH/NCI grant R01 GM126550 and NSF grants DMS-1407241 and IIS-1632951.

[§]Professor, Department of Statistics, Purdue University, West Lafayette, IN 47906. E-mail: chengg@purdue.edu. Research Sponsored by NSF CAREER Award DMS-1151692, DMS-1418042, DMS-1712919 and ONR N00014-15-1-2331.

# 1 Introduction

Graphical models have been widely employed to represent conditional dependence relationships among a set of variables. The structure recovery of an undirected Gaussian graph is known to be equivalent to recovering the support of its corresponding precision matrix (Lauritzen, 1996). In the situation where data dimension is comparable to or much larger than the sample size, the penalized likelihood method is proven to be an effective way to learn the structure of graphical models (Yuan and Lin, 2007; Friedman et al., 2008; Shojaie and Michailidis, 2010a,b). When observations come from several distinct subpopulations, a naive way is to estimate each graphical model separately. However, separate estimation ignores the information of common structure shared across different subpopulations, and thus can be inefficient in some real applications. For instance, in the glioblastoma multiforme (GBM) cancer dataset from The Cancer Genome Atlas Research Network (TCGA, 2008), Verhaak et al. (2010) showed that GBM cancer could be classified into four subtypes. Based on this cluster structure, it has been suggested that although the graphs across four subtypes differ in some edges, they share many common structures. In this case, the naive procedure can be suboptimal (Danaher et al., 2014; Lee and Liu, 2015). Such applications have motivated recent studies on joint estimation methods (Guo et al., 2011; Danaher et al., 2014; Lee and Liu, 2015; Qiu et al., 2016; Wang, 2015; Cai et al., 2016a; Peterson et al., 2015) that encourage common structure in estimating heterogeneous graphical models. However, all aforementioned approaches crucially rely on an assumption that the class label of each sample is known in advance.

For certain problems, prior knowledge of the class membership may be available. But this may not be the case for the massive data with complex and unknown population structures. For instance, in online advertising, an important task is to find the most suitable advertisement (ad) for a given user in a specific online context. This could increase the chance of users' favorable actions (e.g., click the ad, inquire about or purchase a product). In recent years, user clustering has gained increasing attention due to its superior performance of ad targeting. This is because users with similar attributes, such as gender, age, income, geographic information, and online behaviors, tend to behave similarly to the same ad (Yan et al., 2009). Moreover, it is very important to understand conditional dependence relationships among user attributes in order to improve ad targeting accuracy (Wang et al., 2015a). Such conditional dependence relationships are expected to share commonality across different groups (user homogeneity) while maintaining some levels of uniqueness within each group (user heterogeneity) (Jeziorski and Segal, 2015). In this online advertising application, previously mentioned joint estimation methods are no longer applicable as they need to know the user cluster structure in advance. Furthermore, with the data being continuously collected, the number of underlying user clusters grows with the sample size (Chen et al., 2009). This provides another reason for simultaneously conducting user clustering and joint graphical model estimation, which is much needed in the era of big data.

Our contributions in this paper are two-fold. On the methodological side, we propose a general framework of **S**imultaneous **C**lustering **A**nd estimatio**N** of heterogeneous graphical models (SCAN). SCAN is a likelihood based method which treats the underlying class label as a latent variable. Based on a high-dimensional version of Expectation Conditional Maximization (ECM) algorithm (Meng and Rubin, 1993), we are able to conduct clustering and sparse graphical model learning at the same time. In each iteration of the ECM algorithm, the expectation step performs cluster analysis by estimating missing labels and the conditional maximization step conducts feature selection and joint estimation of heterogeneous graphical models via a penalization procedure. With an iteratively updating process, the estimation for both cluster structure and sparse precision matrices becomes more and more refined. Our algorithm is computationally efficient by taking advantage of the fast sparse learning in the conditional maximization step. Moreover, it can be implemented in a user-friendly fashion, without the need of additional unsupervising learning knowledge.

As a promising application, we apply the SCAN method on the GBM cancer dataset to simultaneously cluster the GBM patients and construct the gene regulatory network of each subtype. Our method greatly outperforms the competitors in clustering accuracy and delivers new insights in understanding the GBM disease. Figure 1 reports four gene networks estimated from the SCAN method. The black lines are links shared in all four subtypes, and the color lines are uniquely presented in some subtypes. Our findings generally agree with the GBM disease literature (Verhaak et al., 2010). Besides common edges of all subtypes, we have discovered some unique gene connections that were not found through separate estimation (Danaher et al., 2014; Lee and Liu, 2015). This new finding suggests further investigation on their possible impact on the GBM disease. See Section 4.5 for more discussions.

On the theoretical side, we develop non-asymptotic statistical analysis for the output directly from the high dimensional ECM algorithm. This is nontrivial due to the non-convexity of the likelihood function. In this case, there is no guarantee that the sample-based estimator is close to the maximum likelihood estimator. Hence, we need to directly evaluate the estimation error in each iteration. Let $\boldsymbol{\Theta}$ represent vectorized cluster means $\boldsymbol{\mu}_k$ and precision matrices $\boldsymbol{\Omega}_k$, see (2.2) for a formal definition. Given an appropriate initialization $\boldsymbol{\Theta}^{(0)}$, the finite sample error bound of the $t$-th step solution $\boldsymbol{\Theta}^{(t)}$ consists of two parts:

$$\left\|\boldsymbol{\Theta}^{(t)} - \boldsymbol{\Theta}^*\right\|_2 \leq \underbrace{C \cdot \varepsilon\left(n, p, K, \Psi(\mathcal{M})\right)}_{\text{Statistical Error(SE)}} + \underbrace{\kappa^t \left\|\boldsymbol{\Theta}^{(0)} - \boldsymbol{\Theta}^*\right\|_2}_{\text{Optimization Error(OE)}}, \tag{1.1}$$

with high probability. Here, $K$ is the number of clusters, $\Psi(\mathcal{M})$ measures the sparsity of cluster means and precision matrices, and $\kappa \in (0, 1)$ is a contraction coefficient. The above theoretical analysis is applicable to any decomposable penalty used in the conditional maximization step.

Figure 1: Estimated gene networks corresponding to the Classical, Mesenchymal, Neural and Proneural clusters from our SCAN method applying to the Glioblastoma Cancer Data. In each network, the black lines are the links shared in all four groups. The color lines are the edges shared by some subtypes.

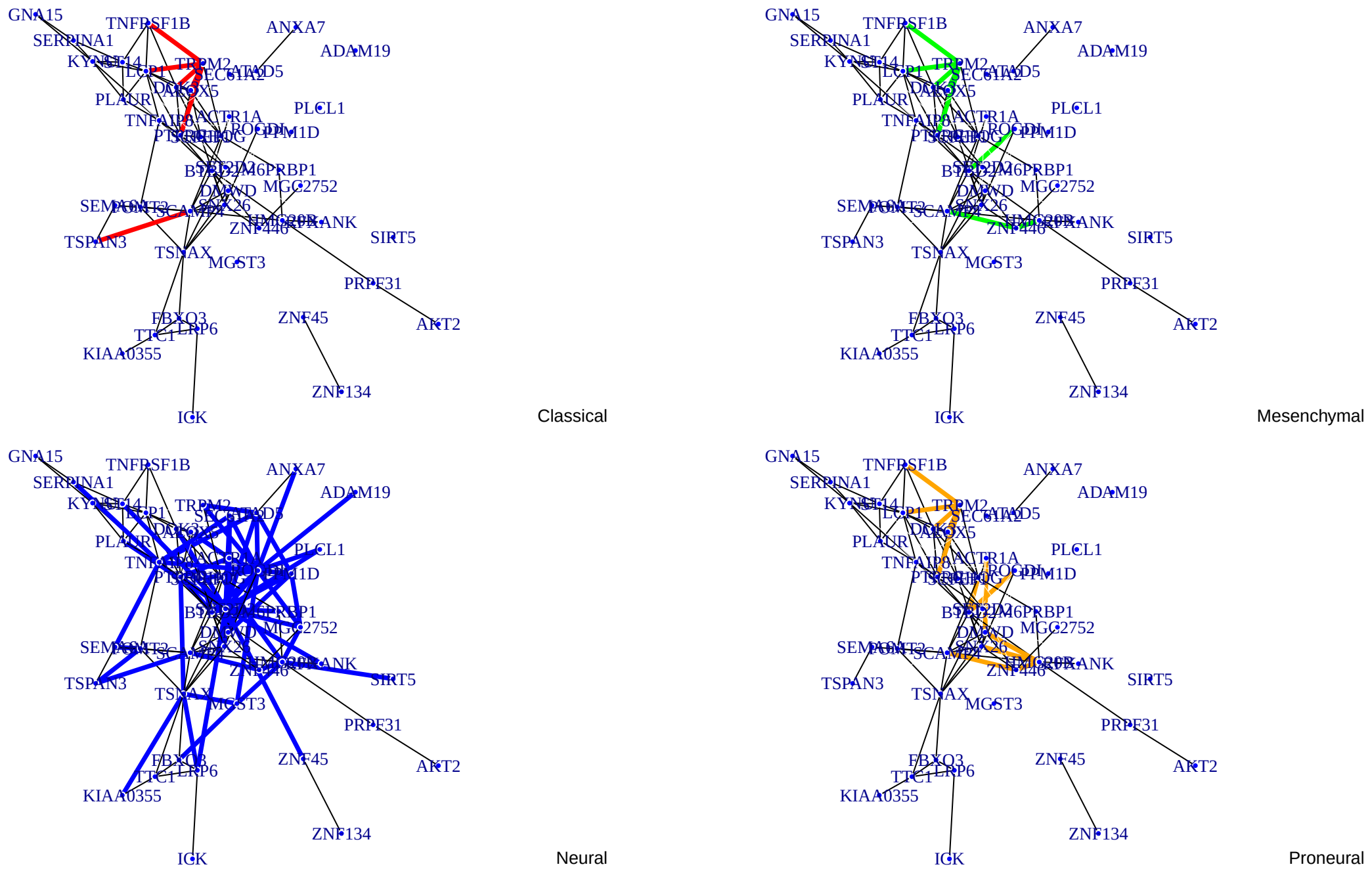

The error bound (1.1) enables us to monitor the dynamics of estimation error in each iteration. Specifically, the optimization error decays geometrically with the iteration number $t$, while the statistical error remains the same when $t$ grows. Therefore, the maximal number of iterations $T$ is implied, beyond which the optimization error is dominated by the statistical error such that consequently the whole error bound is in the same order as the statistical error. In particular,

$$\sum_{k=1}^{K} \Big( \Big\| \boldsymbol{\mu}_k^{(T)} - \boldsymbol{\mu}_k^* \Big\|_2 + \Big\| \boldsymbol{\Omega}_k^{(T)} - \boldsymbol{\Omega}_k^* \Big\|_F \Big) = O_P \left( \underbrace{\sqrt{\frac{K^5 d \log p}{n}}}_{\text{Cluster means error}} + \underbrace{\sqrt{\frac{K^3(Ks + p)\log p}{n}}}_{\text{Precision matrices error}} \right),$$

where $d$ and $s$ are the sparsity for a single cluster mean and precision matrix. This result indicates that, after $T$ steps, the SCAN estimator will fall within statistical precision of the true parameter $\{\boldsymbol{\mu}_k^*, \boldsymbol{\Omega}_k^*\}$. It is worth mentioning that our theory allows the number of clusters $K$ to diverge polynomially with the sample size, reflecting a typical big data scenario. When $K$ is fixed, our statistical rate for the precision matrix estimation under the Frobenius norm, i.e., $O_P(\sqrt{(s+p)\log p/n})$, achieves the optimal rate established in Theorem 7 of Cai et al. (2016b), which is the best rate we could obtain even when the true cluster structure is given.

In the literature, a related line of research focuses on methodological developments of high-dimensional clustering. Pan and Shen (2007) and Sun et al. (2012) introduced regularized model-based clustering and regularized $K$-means clustering, and Zhou et al. (2009) proposed a network-based clustering approach by imposing a graphical lasso to each individual precision matrix estimation. However, the regularized model-based clustering assumes an identical covariance matrix in each cluster, while the network-based clustering treats each graphical model estimation separately. As pointed out in Danaher et al. (2014) and Lee and Liu (2015), ignoring the network information of other clusters may lead to suboptimal graphical model estimation. During the submission of our paper, we became aware of an independent work by Gao et al. (2016) who also considered the multiple precision matrices estimation via a Gaussian mixture model. Different from ours, Gao et al. (2016) did not enforce the sparsity in the cluster means, which would inevitably lead to sub-optimal estimators in high-dimensional clustering (Yi and Caramanis, 2015; Wang et al., 2015b). Most importantly, no theoretical guarantee was provided in Zhou et al. (2009) and Gao et al. (2016). On the other hand, our SCAN method is more general than these existing methods since we allow the sparsity in both cluster means and precision matrices, and our theoretical analysis of the general SCAN framework sheds some lights on the behavior of these existing method, See Remark 2.1 for more discussions. In addition, in terms of the heterogeneous graphical model estimation, Saegusa and Shojaie (2016) proposed an interesting two-stage method which used hierarchical clustering to obtain cluster memberships and then estimated the multiple graphical models based on the attained cluster assignments. Despite its simplicity, it is unclear how the performance of clustering in the first stage could affect the performance of precision matrix estimation in the second stage. In comparison, our approach unifies clustering and parameter estimation into one optimization framework, which allows us to quantify both estimation errors in each iteration.

Another line of related work is the theoretical analysis of EM algorithm (Balakrishnan et al., 2016; Yi and Caramanis, 2015; Wang et al., 2015b). Specifically, Balakrishnan et al. (2016) studied the low-dimensional Gaussian mixture model, while Wang et al. (2015b) and Yi and Caramanis (2015) considered its high dimensional extensions. However, their methods are not applicable for the estimation of heterogeneous graphical models due to the assumed identity covariance matrix. In fact, our consideration of the general covariance matrix demands more challenging technical analysis since simultaneous estimation of cluster means and covariance matrices induces a bi-convex optimization beyond the non-convexity of the EM algorithm itself. This also explains why ECM is needed instead of EM. To address these technical issues, key ingredients of our theoretical analysis are to bound the dual norm of the gradient of an auxiliary $Q$-function and employ nice properties of bi-convex optimization (Boyd et al., 2011) in the regularized M-estimation framework (Negahban et al., 2012). See Section 3 for more details.

In terms of notation, we use $[K]$ to denote the set $\{1, 2, \ldots, K\}$. For a vector $\boldsymbol{\mu} \in \mathbb{R}^p$, $\|\boldsymbol{\mu}\|_2$ is its Euclidean norm. For a matrix $\boldsymbol{X} \in \mathbb{R}^{p_1 \times p_2}$, we denote $\|\boldsymbol{X}\|_F$ and $\|\boldsymbol{X}\|_2$ as its Frobenius norm and spectral norm, respectively, and define its matrix max norm as $\|\boldsymbol{X}\|_{\max} = \max_{i,j} |X_{ij}|$ and its max induced norm as $\|\boldsymbol{X}\|_\infty = \max_{i=1,\ldots,p_1} \sum_{j=1}^{p_2} |X_{ij}|$, which is simply the maximum absolute row sum of the matrix. For a square matrix $\boldsymbol{A} \in \mathbb{R}^{p \times p}$, let $\sigma_{\min}(\boldsymbol{A})$ and $\sigma_{\max}(\boldsymbol{A})$ be its smallest and largest eigenvalue respectively and $|\boldsymbol{A}|$ be its determinant. For a sub-Gaussian random variable $Z$, we use $\|Z\|_{\psi_2}$ and $\|Z\|_{\psi_1}$ to denote its Orlicz norm. Specifically, $\|Z\|_{\psi_2} = \sup_{p \geq 1} p^{-1/2} (\mathbb{E}|Z|^p)^{1/p}$ and $\|Z\|_{\psi_1} = \sup_{p \geq 1} p^{-1} (\mathbb{E}|Z|^p)^{1/p}$. For two sequences $\{a_n\}$ and $\{b_n\}$ of positive numbers, $a_n \lesssim b_n$ refers to the case that $a_n \leq C b_n$ for some uniform constant $C$. We write $\mathbb{1}(\cdot)$ as an indicator function. Throughout this paper, we use $C, C_1, C_2, \ldots D, D_1, D_2, \ldots$ to denote generic absolute constants, whose values may vary at different places.

The rest of this article is organized as follows. Section 2 introduces heterogeneous graphical models and the SCAN method. Section 3 provides some statistical guarantees for the output directly from the SCAN method. Section 4 shows some simulation results as well as a real data analysis on the Glioblastoma cancer data. Section 5 gives some discussions for future works. The appendix is devoted to the technical details of the main theorems, and the online supplementary material contains all the supporting lemmas and their proofs.

# 2 Methodology

In this section, we introduce the SCAN method that simultaneously conducts high-dimensional clustering and estimation of heterogeneous graphical models.

## 2.1 Heterogeneous Graphical Models

We start our discussions from heterogeneous graphical models with known labels. Assume we are given $K$ groups of data sets $\mathcal{A}_1, \ldots, \mathcal{A}_K$ and the samples in the $k$-th group are generated i.i.d. from the following Gaussian distribution:

$$f_k(\boldsymbol{x}; \boldsymbol{\mu}_k, \boldsymbol{\Sigma}_k) = (2\pi)^{-p/2} |\boldsymbol{\Sigma}_k|^{-1/2} \exp\left\{ -\frac{1}{2} (\boldsymbol{x} - \boldsymbol{\mu}_k)^\top \boldsymbol{\Sigma}_k^{-1} (\boldsymbol{x} - \boldsymbol{\mu}_k) \right\}, k = 1, \ldots, K. \tag{2.1}$$

Let $\mathbf{\Omega}_k = \mathbf{\Sigma}_k^{-1}$ be the $k$-th precision matrix with the $ij$-th entry $\omega_{kij}$. For the $k$-th pair of parameters $(\boldsymbol{\mu}_k, \mathbf{\Omega}_k)$, i.e.,

$$\boldsymbol{\mu}_k = \begin{pmatrix} \mu_{k1} \\ \vdots \\ \mu_{kp} \end{pmatrix}, \mathbf{\Omega}_k = \begin{pmatrix} \omega_{k11} & \cdots & \omega_{k1p} \\ \vdots & \ddots & \vdots \\ \omega_{kp1} & \cdots & \omega_{kpp} \end{pmatrix},$$

we write $\mathbf{\Theta}_k := \text{vec}(\boldsymbol{\mu}_k, \mathbf{\Omega}_k) = (\mu_{k1}, \ldots, \mu_{kp}, \omega_{k11}, \ldots, \omega_{kp1}, \ldots, \omega_{k1p}, \ldots, \omega_{kpp}) \in \mathbb{R}^{p^2+p}$ as its vectorized representation, and write the parameter of interest $\mathbf{\Theta}$ as

$$\mathbf{\Theta} = \big(\mathbf{\Theta}_1, \ldots, \mathbf{\Theta}_K\big)^\top \in \mathbb{R}^{K(p^2+p)}. \tag{2.2}$$

Note that the degrees of freedom of $\mathbf{\Theta}$ are $K(0.5p^2 + 1.5p)$, including $K$ sets of $p$ means, $p$ variances, as well as $p(p-1)/2$ covariances.

In some cases, there may also exist some common structure across $K$ precision matrices. Danaher et al. (2014) formulated the joint estimation of heterogeneous graphical models as

$$\underset{\mathbf{\Omega}_1,\ldots,\mathbf{\Omega}_K \succ 0}{\text{argmax}} \sum_{k=1}^{K} \sum_{\boldsymbol{x} \in \mathcal{A}_k} \log f_k(\boldsymbol{x}; \mathbf{\Theta}_k) - \mathcal{P}(\mathbf{\Omega}_1, \ldots, \mathbf{\Omega}_K), \tag{2.3}$$

where $\mathcal{P}(\mathbf{\Omega}_1, \ldots, \mathbf{\Omega}_K)$ is an entry-wise penalty which encourages both sparsity of each individual precision matrix and similarity among all precision matrices.

In practice, the cluster label is not always available. A probabilistic model is thus needed to accommodate the latent structure in the data. Assume the observation $\boldsymbol{x}_i; i = 1, \ldots, n$, from unlabeled heterogeneous population has the underlying density

$$f(\boldsymbol{x}, \mathbf{\Theta}) = \sum_{k=1}^{K} \pi_k f_k(\boldsymbol{x}; \mathbf{\Theta}_k), \tag{2.4}$$

where $\pi_k$ is the probability that an observation $\boldsymbol{x}_i$ belongs to the $k$-th subpopulation. Here, for simplicity we assume the number of cluster $K$ is identifiable. In order to ensure the identifiability of fixed-dimensional Gaussian graphical models, some sufficient conditions such as the strong identifiability condition (Chen, 1995; Nguyen, 2013) was imposed on the density functions. However these conditions are hard to verify in practice. In fact, the identifiability issue for high dimensional mixture model is still an open problem (Ho and Nguyen, 2015) and is beyond the scope of this paper.

Consider the penalized log-likelihood function for the *observed data*

$$\log \mathcal{L}(\mathbf{\Theta}|\boldsymbol{X}) := \frac{1}{n} \sum_{i=1}^{n} \log \left( \sum_{k=1}^{K} \pi_k f_k\Big(\boldsymbol{x}_i; \boldsymbol{\mu}_k, (\mathbf{\Omega}_k)^{-1}\Big) \right) - \mathcal{R}(\mathbf{\Theta}).$$

Our **S**imultaneous **C**lustering **A**nd estimatio**N** (SCAN) method aims to solve

$$\max_{\pi_k,\boldsymbol{\mu}_k,\boldsymbol{\Omega}_k} \log \mathcal{L}(\boldsymbol{\Theta}|\boldsymbol{X}). \tag{2.5}$$

For an illustration, we take

$$\mathcal{R}(\boldsymbol{\Theta}) = \lambda_1 \underbrace{\sum_{k=1}^{K}\sum_{j=1}^{p}|\mu_{kj}|}_{\mathcal{P}_1(\boldsymbol{\Theta})} + \lambda_2 \underbrace{\sum_{k=1}^{K}\sum_{i\neq j}|\omega_{kij}|}_{\mathcal{P}_2(\boldsymbol{\Theta})} + \lambda_3 \underbrace{\sum_{i\neq j}(\sum_{k=1}^{K}\omega_{kij}^2)^{1/2}}_{\mathcal{P}_3(\boldsymbol{\Theta})}, \tag{2.6}$$

where $\mathcal{P}_1(\boldsymbol{\Theta})$ and $\mathcal{P}_2(\boldsymbol{\Theta})$ impose sparsity of the estimated cluster mean and precision matrix, and $\mathcal{P}_3(\boldsymbol{\Theta})$ encourages similarity among all estimated precision matrices. The above three tuning parameters can be tuned efficiently via adaptive BIC. More details can be found in Section 4.1.

**Remark 2.1.** It is worth mentioning that our SCAN method is applicable to penalty functions other than (2.6). For instance, the cluster mean penalty can be replaced by the group lasso penalty in Sun et al. (2012) or the $\ell_0$-norm penalty in Shen et al. (2012). The group graphical lasso penalty for the precision matrix estimation can be substituted by the structural pursuit penalty in Zhu et al. (2014) or the weighted bridge penalty in Rothman and Forzani (2014). As shown in Section 2.2, only a slight modification of our algorithm is needed to accommodate other penalty functions. We also note that SCAN reduces to the regularized model-based clustering (Pan and Shen, 2007) when $\lambda_2 = \lambda_3 = 0$, reduces to the method by Zhou et al. (2009) when $\lambda_3 = 0$, and reduces to the method by Gao et al. (2016) when $\lambda_1 = 0$. Consequently, the technical tools developed for the SCAN estimator in Section 3 are also applicable to these special cases.

## 2.2 ECM Algorithm

In this subsection, we introduce an efficient ECM algorithm to solve the general non-convex optimization problem in (2.5). The ECM replaces each M-step with an conditional maximization (CM) step in which each parameter $\pi_k, \boldsymbol{\mu}_k, \boldsymbol{\Omega}_k$ is maximized separately, by fixing other parameters.

Denote the latent cluster assignment matrix as $\boldsymbol{L}$, where $L_{ik} = \mathbb{1}(\boldsymbol{x}_i \in \mathcal{A}_k)$; $i = 1,\ldots,n$, $k = 1,\ldots,K$. If the cluster label $L_{ik}$ is available, the penalized log-likelihood function for the *complete data* can be formulated as

$$\log \mathcal{L}(\boldsymbol{\Theta}|\boldsymbol{X},\boldsymbol{L}) := \frac{1}{n}\sum_{i=1}^{n}\sum_{k=1}^{K} L_{ik}\Big[\log \pi_k + \log f_k(\boldsymbol{x}_i;\boldsymbol{\Theta}_k)\Big] - \mathcal{R}(\boldsymbol{\Theta}).$$

In the expectation step, the conditional expectation of the penalized log-likelihood function is computed as

$$\mathbb{E}_{\boldsymbol{L}|\boldsymbol{X},\boldsymbol{\Theta}^{(t-1)}}\Big[\log \mathcal{L}(\boldsymbol{\Theta}|\boldsymbol{X},\boldsymbol{L})\Big] = Q_n(\boldsymbol{\Theta}|\boldsymbol{\Theta}^{(t-1)}) - \mathcal{R}(\boldsymbol{\Theta}), \tag{2.7}$$

where $\mathcal{R}(\boldsymbol{\Theta})$ is the penalty in (2.6) and

$$Q_n(\boldsymbol{\Theta}|\boldsymbol{\Theta}^{(t-1)}) := \frac{1}{n}\sum_{i=1}^{n}\sum_{k=1}^{K} L_{\boldsymbol{\Theta}^{(t-1)},k}(\boldsymbol{x}_i)\Big[\log \pi_k + \log f_k(\boldsymbol{x}_i;\boldsymbol{\Theta}_k)\Big], \tag{2.8}$$

with the class label being computed based on the parameter $\boldsymbol{\Theta}^{(t-1)}$ and $\pi_k^{(t-1)}$ obtained at the previous iteration, that is,

$$L_{\boldsymbol{\Theta}^{(t-1)},k}(\boldsymbol{x}_i) = \frac{\pi_k^{(t-1)} f_k(\boldsymbol{x}_i;\boldsymbol{\Theta}_k^{(t-1)})}{\sum_{k=1}^{K}\pi_k^{(t-1)} f_k(\boldsymbol{x}_i;\boldsymbol{\Theta}_k^{(t-1)})}. \tag{2.9}$$

In the conditional maximization step, maximizing (2.7) with respect to $\pi_k$, $\boldsymbol{\mu}_k$, $\boldsymbol{\Omega}_k$ yields the update of parameters. In particular, the update of $\pi_k$ is given as

$$\pi_k^{(t)} = \sum_{i=1}^{n}\frac{L_{\boldsymbol{\Theta}^{(t-1)},k}(\boldsymbol{x}_i)}{n}, \tag{2.10}$$

and the update of $\boldsymbol{\mu}_k$ is given in the following Lemma.

**Lemma 2.2.** Let $\boldsymbol{\mu}_k^{(t)} := \arg\max_{\boldsymbol{\mu}_k} Q_n(\boldsymbol{\Theta}|\boldsymbol{\Theta}^{(t-1)}) - \mathcal{R}(\boldsymbol{\Theta})$ and denote $n_k := \sum_{i=1}^{n} L_{\boldsymbol{\Theta}^{(t-1)},k}(\boldsymbol{x}_i)$. We have, for $j = 1,\ldots,p$,

$$\mu_{kj}^{(t)} = \begin{cases} g_{1,j}(\boldsymbol{x};\boldsymbol{\Theta}_k^{(t-1)}) - \frac{n\lambda_1}{n_k\omega_{kjj}^{(t-1)}}\text{sign}(\mu_{kj}^{(t-1)}) & \text{if} \left|\sum_{i=1}^{n} g_{2,j}(\boldsymbol{x}_i;\boldsymbol{\Theta}_k^{(t-1)})\right| > \lambda_1; \\ 0 & \text{otherwise}, \end{cases}$$

where

$$g_{1,j}(\boldsymbol{x};\boldsymbol{\Theta}_k^{(t-1)}) = \frac{\sum_{i=1}^{n} L_{\boldsymbol{\Theta}^{(t-1)},k}(\boldsymbol{x}_i)\big(\sum_{l=1}^{p} x_{il}\omega_{klj}^{(t-1)}\big)}{\omega_{kjj}^{(t-1)} n_k} - \frac{\sum_{l=1}^{p}\mu_{kl}^{(t-1)}\omega_{klj}^{(t-1)}}{\omega_{kjj}^{(t-1)}} + \mu_{kj}^{(t-1)},$$

$$g_{2,j}(\boldsymbol{x}_i;\boldsymbol{\Theta}_k^{(t-1)}) = L_{\boldsymbol{\Theta}^{(t-1)},k}(\boldsymbol{x}_i)\Big(\sum_{l=1,l\neq j}^{p}(x_{il}-\mu_{kl}^{(t-1)})\omega_{klj}^{(t-1)} + x_{ij}\omega_{kjj}^{(t-1)}\Big).$$

Note that if the lasso penalty is replaced with other penalty functions, then the update formula of $\boldsymbol{\mu}_k^{(t)}$ in Lemma 2.2 can be modified accordingly. Given the pseudo sample covariance matrix $\widetilde{S}_k$, we are able to develop an update formula for $\boldsymbol{\Omega}_k$ by establishing its connection with joint estimation of heterogeneous graphical models (2.3).

**Lemma 2.3.** The solution of maximizing (2.7) with respect to $(\boldsymbol{\Omega}_1,\ldots,\boldsymbol{\Omega}_K)$ is equivalent to

$$(\boldsymbol{\Omega}_1^{(t)},\ldots,\boldsymbol{\Omega}_K^{(t)}) := \arg\max_{\boldsymbol{\Omega}_1,\ldots,\boldsymbol{\Omega}_K\succ 0}\sum_{k=1}^{K} n_k\Big[\log\det(\boldsymbol{\Omega}_k) - \text{trace}(\widetilde{S}_k\boldsymbol{\Omega}_k)\Big] - \mathcal{R}(\boldsymbol{\Theta}), \tag{2.11}$$

where $\widetilde{S}_k$ is a pseudo sample covariance matrix defined as

$$\widetilde{S}_k := \frac{\sum_{i=1}^{n} L_{\boldsymbol{\Theta}^{(t-1)},k}(\boldsymbol{x}_i)(\boldsymbol{x}_i-\boldsymbol{\mu}_k^{(t-1)})^\top(\boldsymbol{x}_i-\boldsymbol{\mu}_k^{(t-1)})}{\sum_{i=1}^{n} L_{\boldsymbol{\Theta}^{(t-1)},k}(\boldsymbol{x}_i)}.$$

The solution for (2.11) can be solved efficiently via the ADMM algorithm by slightly modifying the joint graphical lasso algorithm in Danaher et al. (2014). Since Danaher et al. (2014) do not impose the symmetry condition for precision matrix update, $\{\mathbf{\Omega}_k^{(T)}\}_{k=1}^K$ in general is not necessarily symmetric. Following the symmetrization strategy in Cai et al. (2011) and Cai et al. (2016a), we symmetrize $\mathbf{\Omega}_k^{(t)}$ by

$$\omega_{kij}^{(t)} = \omega_{kij}^{(t)} I(|\omega_{kij}^{(t)} \le \omega_{kij}^{(t)}|) + \omega_{kji}^{(t)} I(|\omega_{kij}^{(t)} > \omega_{kij}^{(t)}|), \tag{2.12}$$

where $\omega_{kij}^{(t)}$ is the $ij$-th entry of $\mathbf{\Omega}_k^{(t)}$ and $I(\cdot)$ is the indicator function. This step will not affect the convergence rate of the final estimator, which is illustrated in Cai et al. (2011) and Cai et al. (2016a). We summarize the high-dimensional ECM algorithm for solving the SCAN method in Table 1. Our algorithm is computationally efficient due to fast sparse learning routines shown in Lemmas 2.2 and 2.3.

Table 1: The SCAN Algorithm

| |
|---|
| **Input**: $\boldsymbol{x}_1, \ldots, \boldsymbol{x}_n$, number of clusters $K$, tuning parameters $\lambda_1, \lambda_2, \lambda_3$. |
| **Output**: Cluster label $\boldsymbol{L}$, cluster mean $\boldsymbol{\mu}_k$ and precision matrix $\mathbf{\Omega}_k$. |
| **Step 1**: Initialize cluster mean $\boldsymbol{\mu}_k^{(0)}$, positive definite precision matrix $\mathbf{\Omega}_k^{(0)}$, and set $\pi_k^{(0)} = 1/K$, for each $k \in [K]$. |
| **Step 2**: Until some termination conditions are met, for iteration $t = 1, 2, \ldots$ |
| (a) E-step. Find the cluster assignment $L_{\mathbf{\Theta}^{(t-1)},k}(\boldsymbol{x}_i)$ as in (2.9). |
| (b) CM-step. Given $L_{\mathbf{\Theta}^{(t-1)},k}(\boldsymbol{x}_i)$, update $\pi_k^{(t)}$, $\boldsymbol{\mu}_k^{(t)}$, and $\mathbf{\Omega}_k^{(t)}$ in (2.10), Lemma 2.2, Lemma 2.3, respectively. Symmetrize $\mathbf{\Omega}_k^{(t)}$ by (2.12). |

In all of our experiments, we obtain $(\boldsymbol{\mu}_k^{(0)}, \mathbf{\Omega}_k^{(0)})$ by random initialization, which is computationally efficient and practically reliable. In the theoretical study, we require the initialization to be of a constant distance to the truth. See Remark 3.10 for more discussions. Moreover, in the implementation, ECM step in Step 2 is terminated when the updated parameters are close to their previous values:

$$\sum_{k=1}^K \left\{ \frac{\|\boldsymbol{\mu}_k^{(t)} - \boldsymbol{\mu}_k^{(t-1)}\|_2}{\|\boldsymbol{\mu}_k^{(t)}\|_2} + \frac{\|\mathbf{\Omega}_k^{(t)} - \mathbf{\Omega}_k^{(t-1)}\|_F}{\|\mathbf{\Omega}_k^{(t)}\|_F} \right\} \le 0.01.$$

**Remark 2.4.** In the existing high-dimensional EM algorithms where the covariance matrix is assumed to be an identity matrix (Wang et al., 2015b; Yi and Caramanis, 2015), sample-splitting procedures have been routinely used in the M-step in order to facilitate the theoretical analysis. Although it simplifies theoretical developments, such a sample-splitting procedure does not take advantage of full samples in the M-step and is hard to implement in practice. Our Algorithm 1 is able to avoid this sample-splitting step but still enjoys nice theoretical properties. See Corollary 3.14 for more discussions on its statistical guarantee.

# 3 Statistical Guarantee

In this section, we establish statistical guarantee for the SCAN estimator based on sample-based analysis of (2.8) and population-based analysis of (3.3). Here, we consider the high-dimensional setting where $p \gg n$ and $K$ is allowed to diverge with $n$.

We start by introducing some useful notation. Denote the index set of diagonal components of $K$ precision matrices by

$$\mathcal{G} = \bigcup_{k=1}^{K} \mathcal{G}_k, \text{with } \mathcal{G}_k = \left(k(p+1), k(2p+2), \ldots, k(p^2+p)\right), \tag{3.1}$$

that is, $\mathbf{\Theta}_{\mathcal{G}} = (\omega_{111}, \ldots, \omega_{1pp}, \ldots, \omega_{K11}, \ldots, \omega_{Kpp}) \in \mathbb{R}^{Kp}$. Let $\mathcal{O}$ be the complete index set of $\mathbf{\Theta}$ and $\mathcal{G}^c = \mathcal{O} \setminus \mathcal{G}$ be the complement set of $\mathcal{G}$. Denote $\mathcal{U}_k := \{i : \mu^*_{ki} \neq 0\}$ where $\boldsymbol{\mu}^*_k$ is the true mean parameter, $\mathcal{V}_k := \{(i,j) : i \neq j, \omega^*_{kij} \neq 0\}$ where $\mathbf{\Omega}^*_k$ is the true precision matrix and $\mathcal{S}_1 = \bigcup_{k=1}^{K} \mathcal{U}_k$, $\mathcal{S}_2 = \bigcup_{k=1}^{K} \mathcal{V}_k$. Define $\Xi \subseteq \mathbb{R}^{K(p^2+p)}$ as some non-empty convex set of parameters. Denote the support space $\mathcal{M}$ as

$$\begin{aligned} \mathcal{M} \ := \ \Big\{ \boldsymbol{V} \in \Xi \ \big| \ & \mu_{ki} = 0 \text{ for all } i \notin \mathcal{S}_1, \\ & \omega_{kij} = 0 \text{ for all pairs } (i,j) \notin \mathcal{S}_2, k = 1 \ldots, K \Big\}, \end{aligned} \tag{3.2}$$

where $\boldsymbol{V}$ follows the same definition style used for $\mathbf{\Theta}$ in (2.2). Denote the sparsity parameters:

$$\begin{aligned} & s := \#\{(i,j) : \omega^*_{kij} \neq 0, i,j = 1 \ldots p, i \neq j, k = 1, \ldots, K\}, \\ & d := \#\{i : \mu^*_{ik} \neq 0, i = 1, \ldots, p, k = 1, \ldots, K\}. \end{aligned}$$

## 3.1 Population-Based Analysis

We define a corresponding population version of $Q_n$ in (2.8) as

$$Q(\mathbf{\Theta}^{'}|\mathbf{\Theta}) := \mathbb{E}\left[\sum_{k=1}^{K} L_{\mathbf{\Theta},k}(\boldsymbol{X})[\log \pi^{'}_k + \log f_k(\boldsymbol{X}; \mathbf{\Theta}^{'}_k)]\right]. \tag{3.3}$$

Without loss of generality, we assume the true prior probability $\pi^*_k = 1/K$ for each $k = 1, \ldots, K$. Recall that the update of weights in (2.10) is independent of the updates of other parameters. Consequently, according to (2.1), maximizing $Q(\mathbf{\Theta}^{'}|\mathbf{\Theta})$ over $(\boldsymbol{\mu}^{'}_k, \mathbf{\Omega}^{'}_k)$ is equivalent to maximizing

$$\sum_{k=1}^{K} \mathbb{E}\left[L_{\mathbf{\Theta},k}(\boldsymbol{X}) \left\{\frac{1}{2}\log\det(\mathbf{\Omega}^{'}_k) - \frac{1}{2}(\boldsymbol{X} - \boldsymbol{\mu}^{'}_k)^\top \mathbf{\Omega}^{'}_k (\boldsymbol{X} - \boldsymbol{\mu}^{'}_k)\right\}\right]. \tag{3.4}$$

Clearly, the update of $(\boldsymbol{\mu}^{'}_l, \mathbf{\Omega}^{'}_l)$ is independent of the update of $(\boldsymbol{\mu}^{'}_t, \mathbf{\Omega}^{'}_t)$ for any $t \neq l$. This enables us to characterize the update of each pair of parameters separately. For any $k = 1, \ldots, K$, define

$$M_{\boldsymbol{\mu}^{'}_k}(\mathbf{\Omega}^{'}_k) := \arg\max_{\boldsymbol{\mu}^{'}_k} Q(\mathbf{\Theta}^{'}|\mathbf{\Theta}) \text{ and } M_{\mathbf{\Omega}^{'}_k}(\boldsymbol{\mu}^{'}_k) := \arg\max_{\mathbf{\Omega}^{'}_k} Q(\mathbf{\Theta}^{'}|\mathbf{\Theta}).$$

We show in Lemma 3.1 that the population update of $\boldsymbol{\mu}_k^{'}$ is independent of $\boldsymbol{\Omega}_k^{'}$, while the population update of $\boldsymbol{\Omega}_k^{'}$ is a function of $\boldsymbol{\mu}_k^{'}$.

**Lemma 3.1.** For any $k = 1, \ldots, K$, we have

$$M_{\boldsymbol{\mu}_k^{'}}(\boldsymbol{\Omega}_k^{'}) = \left[\mathbb{E}[L_{\boldsymbol{\Theta},k}(\boldsymbol{X})]\right]^{-1}\mathbb{E}[L_{\boldsymbol{\Theta},k}(\boldsymbol{X})\boldsymbol{X}], \tag{3.5}$$

$$M_{\boldsymbol{\Omega}_k^{'}}(\boldsymbol{\mu}_k^{'}) = \mathbb{E}[L_{\boldsymbol{\Theta},k}(\boldsymbol{X})]\left[\mathbb{E}[L_{\boldsymbol{\Theta},k}(\boldsymbol{X})(\boldsymbol{X}-\boldsymbol{\mu}_k^{'})(\boldsymbol{X}-\boldsymbol{\mu}_k^{'})^{\top}]\right]^{-1}. \tag{3.6}$$

The difficulty of simultaneous clustering and estimation can be characterized by the following *sufficiently separable condition*. Define $\mathcal{B}_\alpha(\boldsymbol{\Theta}^*) := \left\{\boldsymbol{\Theta} \in \Xi : \left\|\boldsymbol{\Theta} - \boldsymbol{\Theta}^*\right\|_2 \leq \alpha\right\}$.

**Condition 3.2** (*Sufficiently Separable Condition*)**.** Denote $W = \max_j W_j$, $W^{'} = \max_j W_j^{'}$, $W^{''} = \max_j W_j^{''}$ with $W_j, W_j^{'}, W_j^{''}$ defined in (S.4), (S.7) and (S.8), respectively. We assume $K$ clusters are sufficiently separable such that given an appropriately small parameter $\gamma > 0$, it holds a.s.

$$L_{\boldsymbol{\Theta},k}(\boldsymbol{X}) \cdot L_{\boldsymbol{\Theta},j}(\boldsymbol{X}) \leq \frac{\gamma}{24(K-1)\sqrt{\max\{W, W^{'}, W^{''}\}}}, \tag{3.7}$$

for each pair $\{(j,k), j,k \in [K], j \neq k\}$ and any $\boldsymbol{\Theta} \in \mathcal{B}_\alpha(\boldsymbol{\Theta}^*)$.

Condition 3.2 requires that $K$ clusters are sufficiently separable in the sense that $\boldsymbol{X}$ belongs to the $k$-th cluster with probability either close to zero or close to one such that $L_{\boldsymbol{\Theta},k}(\boldsymbol{X}) \cdot L_{\boldsymbol{\Theta},j}(\boldsymbol{X})$ is close to zero. In the special case that $K = 2$ and $\boldsymbol{\Omega}_1^* = \boldsymbol{\Omega}_2^* = \mathbb{1}_p$, Balakrishnan et al. (2016) requires $\|\boldsymbol{\mu}_1^* - \boldsymbol{\mu}_2^*\|_2$ is sufficiently large. Our Condition 3.2 extends it to general $K$ and general precision matrices. Note that the condition (3.7) is related with the number of clusters $K$. As $K$ grows, the clustering problem gets harder and hence a stronger sufficiently separable condition is needed.

The next lemma guarantees that the curvature of $Q(\cdot|\boldsymbol{\Theta})$ is similar to that of $Q(\cdot|\boldsymbol{\Theta}^*)$ when $\boldsymbol{\Theta}$ is close to $\boldsymbol{\Theta}^*$, which is a key ingredient in our population-based analysis.

**Lemma 3.3** (*Gradient Stability*)**.** Under Condition 3.2, the function $\{Q(\cdot|\boldsymbol{\Theta}), \boldsymbol{\Theta} \in \Xi\}$ satisfies,

$$\left\|\nabla Q(\boldsymbol{\Theta}^*|\boldsymbol{\Theta}) - \nabla Q(\boldsymbol{\Theta}^*|\boldsymbol{\Theta}^*)\right\|_2 \leq \tau \cdot \left\|\boldsymbol{\Theta} - \boldsymbol{\Theta}^*\right\|_2, \tag{3.8}$$

with parameter $\tau \leq \gamma/12$ for any $\boldsymbol{\Theta} \in \mathcal{B}_\alpha(\boldsymbol{\Theta}^*)$. The gradient $\nabla Q(\boldsymbol{\Theta}^*|\boldsymbol{\Theta})$ is taken with respect to the first variable of $Q(\cdot|\cdot)$.

## 3.2 Sample-Based Analysis

In this section, we analyze the sample-base function $Q_n$, defined as the objective function in (2.8). The statistical error comes from the approximation by using sample-base function $Q_n$ to population-base function $Q$. We need one regularity condition to ensure that $Q_n$ is strongly concave in a specific Euclidean ball.

**Condition 3.4.** There exist some positive constants $\beta_1$, $\beta_2$ such that $0 < \beta_1 < \min_{k\in[K]} \sigma_{\min}(\boldsymbol{\Omega}_k^*) < \max_{k\in[K]} \sigma_{\max}(\boldsymbol{\Omega}_k^*) < \beta_2$.

Lemma 3.5 verifies the restricted strong concavity condition of $Q_n$. Note that (3.9) corresponds to the restricted eigenvalue condition in sparse linear regression (Negahban et al., 2012).

**Lemma 3.5** (*Restricted Strong Concavity*)**.** Suppose that Condition 3.4 holds. Then for any $\boldsymbol{\Theta} \in \mathcal{B}_\alpha(\boldsymbol{\Theta}^*)$, with probability at least $1-\delta$, each $\boldsymbol{\Theta}' \in \mathbb{C} := \{\boldsymbol{\Theta}' \mid \|\boldsymbol{\Theta}' - \boldsymbol{\Theta}^*\|_2 \le 2\alpha\}$ satisfies

$$Q_n(\boldsymbol{\Theta}'|\boldsymbol{\Theta}) - Q_n(\boldsymbol{\Theta}^*|\boldsymbol{\Theta}) - \Big\langle \nabla Q_n(\boldsymbol{\Theta}^*|\boldsymbol{\Theta}), \boldsymbol{\Theta}' - \boldsymbol{\Theta}^* \Big\rangle \le -\frac{\gamma}{2}\Big\|\boldsymbol{\Theta}' - \boldsymbol{\Theta}^*\Big\|_2^2, \tag{3.9}$$

with sufficiently large $n$, where $\gamma = c \cdot \min\{\beta_1, 0.5(\beta_2 + 2\alpha)^{-2}\}$ is the strong concavity parameter for some constant $c$.

Define $\mathcal{P}(\boldsymbol{\Theta}) = M_1\mathcal{P}_1(\boldsymbol{\Theta}) + M_2\mathcal{P}_2(\boldsymbol{\Theta}) + M_3\mathcal{P}_3(\boldsymbol{\Theta})$ for some positive constants $M_1, M_2, M_3$. Let $\mathcal{P}^*$ be the dual norm of $\mathcal{P}$, which is defined as $\mathcal{P}^*(\boldsymbol{\Theta}) = \sup_{\mathcal{P}(\boldsymbol{\Theta}')\le 1}\langle \boldsymbol{\Theta}', \boldsymbol{\Theta}\rangle$. For simplicity, write $\|\cdot\|_{\mathcal{P}^*} = \mathcal{P}^*(\cdot)$.

**Condition 3.6.** For any fixed $\boldsymbol{\Theta} \in \mathcal{B}_\alpha(\boldsymbol{\Theta}^*)$, with probability at least $1-\delta_1$,

$$\Big\|\nabla Q_n(\boldsymbol{\Theta}^*|\boldsymbol{\Theta}) - \nabla Q(\boldsymbol{\Theta}^*|\boldsymbol{\Theta})\Big\|_{\mathcal{P}^*} \le \varepsilon_1, \tag{3.10}$$

and with probability at least $1-\delta_2$, we have

$$\Big\|\Big[\nabla Q_n(\boldsymbol{\Theta}^*|\boldsymbol{\Theta}) - \nabla Q(\boldsymbol{\Theta}^*|\boldsymbol{\Theta})\Big]_{\mathcal{G}}\Big\|_2 \le \varepsilon_2, \tag{3.11}$$

where $\mathcal{G}$ is the diagonal index set defined in (3.1). Here $\varepsilon_1$ and $\varepsilon_2$ are functions of $n, p, K, \delta_1, \delta_2$.

Intuitively, $\varepsilon_1$ and $\varepsilon_2$ quantify the difference between the population-based and sample-based conditional maximization step. Note that $\mathcal{P}$ does not penalize diagonal elements of each precision matrix, thus

$$\Big\|\nabla Q_n(\boldsymbol{\Theta}^*|\boldsymbol{\Theta}) - \nabla Q(\boldsymbol{\Theta}^*|\boldsymbol{\Theta})\Big\|_{\mathcal{P}^*} = \Big\|\Big[\nabla Q_n(\boldsymbol{\Theta}^*|\boldsymbol{\Theta}) - \nabla Q(\boldsymbol{\Theta}^*|\boldsymbol{\Theta})\Big]_{\mathcal{G}^c}\Big\|_{\mathcal{P}^*}.$$

Our analysis makes use of the property of dual norm to bridge the SCAN penalty term and the targeted error term in $L_2$ norm. Note that our SCAN penalty does not penalize diagonal terms of precision matrices, and hence it can be treated as a norm only if it is applied to the parameter $\boldsymbol{\Theta}$ without diagonal terms of precision matrices. Otherwise, it is a semi-norm. For this purpose, we separate all the diagonal terms from $\boldsymbol{\Theta}$. Therefore, our statistical error is split by two parts: one from the sparse estimate of cluster means and non-diagonal terms in precision matrices, and another from the estimate of diagonal terms of precision matrices. In Lemma S.1, $\varepsilon_1$ and $\varepsilon_2$ will be specifically calculated for our proposed SCAN penalty. In the high dimensional ECM algorithm, there is no explicit form for the CM-step update due to the existence of the penalty term. This is a crucial difference from the low-dimensional EM algorithm in Balakrishnan et al. (2016). Fortunately, the decomposability of SCAN penalty enables us to quantify statistical errors by evaluating the gradient of $Q$-function.

### 3.3 Statistical Error versus Optimization Error

In this section, we provide the final theoretical guarantee for the high-dimensional ECM algorithm by combining the population and sample-based analysis.

**Definition 3.7** (*Support Space Compatibility Constant*)**.** For the support subspace $\mathcal{M} \subseteq \mathbb{R}^{K(p^2+p)}$ defined in (3.2), we define

$$\nu(\mathcal{M}) = \sup_{\boldsymbol{\Theta} \in \mathcal{M} \setminus \{0\}} \frac{\mathcal{P}(\boldsymbol{\Theta})}{\|\boldsymbol{\Theta}\|_2}. \tag{3.12}$$

**Remark 3.8.** The support space compatibility constant $\nu(M)$ is a variant of subspace compatibility constant originally proposed by Negahban et al. (2012) and Wainwright (2014). Actually, $\nu(M)$ can be interpreted as a notion of intrinsic dimensionality of $M$. In order to bound the statistical error, we need some measures for the complexity of parameter $\boldsymbol{\Theta}$ reflected by the penalty term. One possible way is to specify a model subspace $\mathcal{M}$ and require $\boldsymbol{\Theta}$ lie in the space. By choosing the support space $\mathcal{M}$ of parameter of interest $\boldsymbol{\Theta}$, the support space compatibility constant $\nu(\mathcal{M})$ can measure the complexity of $\boldsymbol{\Theta}$ relative to the penalty term $\mathcal{P}$ and square norm. The larger $\nu(\mathcal{M})$ is, the more samples are needed to guarantee statistical consistency. For examples, if the penalty $\mathcal{P}$ is $L_1$ penalty with $s$-sparse coordinate support space $\mathcal{M}'$, then we have $\nu(\mathcal{M}') = \sqrt{s}$. In the context of group lasso penalty, we have $\nu(\mathcal{M}') = \sqrt{|S|}$, where $S$ is the index set of active groups. For our SCAN penalty, $\nu(\mathcal{M})$ is specifically calculated by $M_1\sqrt{Kd} + (M_2\sqrt{K} + M_3)\sqrt{s}$, where $d, s$ are the common sparsity parameters for single cluster means and precision matrices accordingly and $M_1, M_2, M_3$ are some absolute constants.

We first provide a general theory that applies to any decomposable penalty, such as the group lasso penalty in Sun et al. (2012) and fused graphical lasso penalty in Danaher et al. (2014). The theoretical result of our SCAN penalty will be discussed in Corollary 3.14.

**Theorem 3.9.** Suppose Conditions 3.2, 3.4, 3.6 hold and $\boldsymbol{\Theta}^*$ lies in the interior of $\Xi$. Let $\kappa = 6\tau/\gamma$, where $\tau, \gamma$ are calculated in Lemma 3.3 and Lemma 3.5. Consider our SCAN algorithm in Table 1 with initialization $\boldsymbol{\Theta}^{(0)}$ falling into a ball $\mathcal{B}_\alpha(\boldsymbol{\Theta}^*)$ for some constant radius $\alpha > 0$ and assume the tuning parameters satisfy $\lambda_1 = M_1\lambda_n^{(t)}$, $\lambda_2 = M_2\lambda_n^{(t)}$, $\lambda_3 = M_3\lambda_n^{(t)}$, and

$$\lambda_n^{(t)} = \varepsilon + \kappa^t \frac{\gamma}{\nu(\mathcal{M})} \left\| \mathbf{\Theta}^{(t-1)} - \mathbf{\Theta}^* \right\|_2 . \tag{3.13}$$

If the sample size $n$ is large enough such that $\varepsilon \le (1-\kappa)\gamma\alpha/(6\nu(\mathcal{M}))$, then $\mathbf{\Theta}^{(t)}$ satisfies, with probability at least $1 - t\delta'$,

$$\left\| \mathbf{\Theta}^{(t)} - \mathbf{\Theta}^* \right\|_2 \le \underbrace{\frac{6\nu(\mathcal{M})}{(1-\kappa)\gamma}\varepsilon}_{\text{Statistical Error(SE)}} + \underbrace{\kappa^t \left\| \mathbf{\Theta}^{(0)} - \mathbf{\Theta}^* \right\|_2}_{\text{Optimiation Error(OE)}} , \tag{3.14}$$

where $\delta' = \delta + \delta_1 + \delta_2$ with $\delta$, $\delta_1$, $\delta_2$ defined in Lemma 3.5 and Condition 3.6 and $\varepsilon = \varepsilon_1 + \varepsilon_2/\nu(\mathcal{M})$.

The above theoretical result suggests that the estimation error in each iteration consists *statistical error* and *optimization error*. From the definition of $\tau$ in Lemma 3.3, $\kappa$ is less than 0.5 so that it is a contractive parameter. With a relatively good initialization, even though ECM algorithm may be trapped into a local optima after enough iterations, it can be guaranteed to be within a small neighborhood of the truth, in the sense of statistical accuracy. In addition, with a proper choice of $\delta'$, the final probability $1 - t\delta'$ will converge to 1; see Corollary 3.14 for details.

**Remark 3.10.** To our limited knowledge, there is no existing literature to guarantee the global convergence of ECM algorithm in a general case. Compromisingly, we have to require some constraints on the initial value. In our framework, the only requirement for the initial value is to fall into a ball with constant radius to the truth. Such a condition has also been imposed in EM algorithms (Balakrishnan et al., 2016; Wang et al., 2015b; Yi and Caramanis, 2015) and can be fulfilled by some spectral-based initializations (Zhang et al., 2014).

**Remark 3.11.** In Theorem 3.9, we introduce an iterative turning procedure (3.13) which appeared in high dimensional regularized $M$-estimation (Negahban et al., 2012), and was also applied in Yi and Caramanis (2015) to facilitate their theoretical analysis.

The error bound in (3.14) measures the estimation error in each iteration. Here, optimization error decays geometrically with the iteration number $t$, while the statistical error remains the same when $t$ grows. Therefore, this enables us to provide a meaningful choice of the maximal number of iterations $T$ beyond which the optimization error is dominated by the statistical error such that the whole error bound is in the same order of the statistical error.

In the following corollary, taking the SCAN penalty as an example, we provide a closed form of the maximal number of iterations $T$ and also an explicit form of the estimation error.

**Condition 3.12.** The largest element of cluster means and precision matrices are both bounded, that is, for some positive constants $c_1$ and $c_2$,

$$\|\boldsymbol{\mu}^*\|_\infty := \max_{k\in[K]} \|\boldsymbol{\mu}_k^*\|_\infty < c_1 \text{ and } \|\boldsymbol{\Omega}^*\|_{\max} := \max_{k\in[K]} \|\boldsymbol{\Omega}_k^*\|_{\max} < c_2.$$

**Condition 3.13.** Suppose that the number of clusters $K$ satisfies $K^2 = o(p(\log n)^{-1})$.

**Corollary 3.14.** Suppose Conditions 3.2, 3.4, 3.12 and 3.13 hold. If sample size $n$ is sufficiently large such that

$$n \geq \left(\frac{6(CK\|\boldsymbol{\Omega}^*\|_\infty + C'K^{1.5})(\sqrt{Kd} + \sqrt{Ks} + \sqrt{K}) + C''K^{1.5}\sqrt{p}}{(1-\kappa)\gamma\alpha}\right)^2 \log p,$$

and the iteration step $t$ is large enough such that

$$t \geq T = \log_{1/\kappa} \frac{\left\|\boldsymbol{\Theta}^{(0)} - \boldsymbol{\Theta}^*\right\|_2}{\varphi(n,p,K)},$$

where $\varphi(n,p,K) = 6\widetilde{C}((1-\kappa)\gamma)^{-1}\|\boldsymbol{\Omega}^*\|_\infty(\sqrt{Kd} + \sqrt{Ks+p})\sqrt{K^3 \log p/n}$ for some positive constant $\widetilde{C}$, the optimization error in (3.14) is dominated by the statistical error, and

$$\sum_{k=1}^K \left( \left\|\boldsymbol{\mu}_k^{(T)} - \boldsymbol{\mu}_k^*\right\|_2 + \left\|\boldsymbol{\Omega}_k^{(T)} - \boldsymbol{\Omega}_k^*\right\|_F \right) \leq \frac{12\widetilde{C}}{(1-\kappa)\gamma} \left( \underbrace{\|\boldsymbol{\Omega}^*\|_\infty \sqrt{\frac{K^5 d \log p}{n}}}_{\text{Cluster means error}} + \underbrace{\|\boldsymbol{\Omega}^*\|_\infty \sqrt{\frac{K^3(Ks+p)\log p}{n}}}_{\text{Precision matrices error}} \right),$$

with probability converging to 1.

**Remark 3.15.** If $K$ is fixed, the above upper bound reduces to

$$\sum_{k=1}^K \left( \left\|\boldsymbol{\mu}_k^{(T)} - \boldsymbol{\mu}_k^*\right\|_2 + \left\|\boldsymbol{\Omega}_k^{(T)} - \boldsymbol{\Omega}_k^*\right\|_F \right) \lesssim \left( \underbrace{\|\boldsymbol{\Omega}^*\|_\infty \sqrt{\frac{d \log p}{n}}}_{\text{Cluster means error}} + \underbrace{\|\boldsymbol{\Omega}^*\|_\infty \sqrt{\frac{(s+p)\log p}{n}}}_{\text{Precision matrices error}} \right). \quad (3.15)$$

Consider the class of precision matrix $\mathcal{Q} := \{\boldsymbol{\Omega} : \boldsymbol{\Omega} \succ 0, \|\boldsymbol{\Omega}\|_\infty \leq C_{\mathcal{Q}}\}$ as in Cai et al. (2016b). When $C_{\mathcal{Q}}$ does not depend on $n, p$, our rate $\sqrt{(s+p)\log p/n}$ in (3.15) is minimax optimal for estimating $s$-sparse precision matrix under Frobenius norm (see Theorem 7 in Cai et al. (2016b)). The same rate has also been obtained in Saegusa and Shojaie (2016) for multiple precision matrix estimation when the true cluster structure is assumed to be given in advance. Moreover, our cluster mean error rate $\sqrt{d \log p/n}$ is minimax optimal for estimating $d$-sparse cluster means; see Wang et al. (2015b). In short, Corollary 3.14 indicates that our procedure is able to achieve optimal statistical rates for both cluster means and multiple precision matrices even when the true cluster structure is unknown.

**Remark 3.16.** As a by-product, we establish the variable selection consistency of $\mathbf{\Omega}_k^{(T)}$, which ensures that our precision matrix estimator can asymptotically identify true connected links. Assume $\|\mathbf{\Omega}_k^*\|_\infty$ is bounded and the minimal signal in the true precision matrix satisfies $\omega_{\min} := \min_{(i,j)\in\mathcal{V}_k, k=1,\ldots,K} w_{kij}^* > 2r_n$, where $r_n = (\sqrt{K^5 d} + \sqrt{K^3(Ks+p)})\sqrt{\log p/n}$. The latter condition is weaker than that assumed in Guo et al. (2011), where they require a constant lower bound of $\omega_{\min}$. To ensure the model selection consistency, we threshold the precision matrix estimator $\mathbf{\Omega}_k^{(T)}$ such that $\widetilde{\omega}_{kij} = \omega_{kij}^{(T)} \mathbb{1}\{|\omega_{kij}^{(T)}| > r_n\}$ as in Bickel and Levina (2008) and Lee and Liu (2015). See Theorem S.2 in the online supplementary for some results on the selection consistency result.

# 4 Numerical Study

In this section, we discuss an efficient tuning parameter selection procedure and demonstrate the superior numerical performance of our method. We compare our algorithm with three clustering and graphical model estimation methods:

- Standard $K$-means clustering (MacQueen, 1967).
- Algorithm in Zhou et al. (2009) which applies graphical lasso for each precision matrix estimation.
- A two-stage approach which first uses $K$-means clustering to obtain the clusters and then applies joint graphical lasso (Danaher et al., 2014) to estimate precision matrices.

For a fair comparison, we assume the number of clusters $K$ is given in all methods.

## 4.1 Selection of Tuning Parameters

In our simultaneous clustering and graph estimation formulation, three tuning parameters $\Lambda := \{\lambda_1, \lambda_2, \lambda_3\}$ need to be appropriately determined so that both the clustering and network estimation performance can be optimized. In our framework, the tuning parameters are selected through the following adaptive BIC-type selection criterion. For a set of tuning parameters $\Lambda := \{\lambda_1, \lambda_2, \lambda_3\}$, the adaptive BIC criterion is defined as

$$\mathrm{BIC}(\Lambda) = -2\log \widehat{L}(\Lambda) + \log(n)\mathrm{df}_\Lambda(\boldsymbol{\mu}) + 2\mathrm{df}_\Lambda(\mathbf{\Omega}), \tag{4.1}$$

where $\widehat{L}(\Lambda)$ is the sample likelihood function and $\{\mathrm{df}_\Lambda(\boldsymbol{\mu}), \mathrm{df}_\Lambda(\mathbf{\Omega})\}$ is the degrees of freedom of the model. Here, $\{\mathrm{df}_\Lambda(\boldsymbol{\mu}), \mathrm{df}_\Lambda(\mathbf{\Omega})\}$ can be approximated by the size of selected variables in the final estimator. Therefore, according to the Gaussian mixture model assumption, the adaptive BIC criterion in (4.1) can be computed as

$$-2\sum_{i=1}^{n}\log\left(\sum_{k=1}^{K}\widehat{\pi}_k f_k\left(\boldsymbol{x}_i;\widehat{\boldsymbol{\mu}}_k,(\widehat{\boldsymbol{\Omega}}_k)^{-1}\right)\right)+\sum_{k=1}^{K}\{\log n\cdot s_{1k}+2s_{2k}\},$$

where $s_{1k}=\text{Card}\{i:\widehat{\mu}_{ki}\neq 0\}$, $s_{2k}=\text{Card}\{(i,j):\widehat{\Omega}_{kij}\neq 0, 1\leq i<j\leq p\}$ and $\widehat{\pi}_k,\widehat{\boldsymbol{\mu}}_k,\widehat{\boldsymbol{\Omega}}_k$ are final updates from Algorithm 1. We choose a smaller weight for the degrees of freedom of precision matrices as suggested in Danaher et al. (2014). The mixing weight $\pi$ is not counted into the degrees of freedom since it only contributes a constant factor.

In our experiment, we choose the optimal set of parameters minimizing the BIC value in (4.1). In the high-dimensional scenario where $p$ is very large, calculation of BIC over a grid search for all $\lambda_1,\lambda_2,\lambda_3$ may be computationally expensive. Following Danaher et al. (2014), we suggest a line search over $\lambda_1$, $\lambda_2$ and $\lambda_3$. In detail, we fix $\lambda_2$ and $\lambda_3$ at their median value of the given range and conduct a grid search over $\lambda_1$. Then with tuned $\lambda_1$ and median value of $\lambda_3$, we conduct a grid search over $\lambda_2$. The line search for $\lambda_3$ is the same. In our simulations, we choose the tuning range $10^{-2+2t/15}$ with $t=0,1,\ldots,15$ for all $\lambda_1,\lambda_2,\lambda_3$.

## 4.2 Illustration

In this subsection, we demonstrate the importance of simultaneous clustering and estimation in improving both the clustering performance and the estimation accuracy of multiple precision matrices.

The simulated data consists of $n=1000$ observations from 2 clusters, and among them 500 observations are from $\mathcal{N}(\boldsymbol{\mu}_1,\boldsymbol{\Sigma})$ and the rest 500 observations are from $\mathcal{N}(\boldsymbol{\mu}_2,\boldsymbol{\Sigma})$ with $\boldsymbol{\mu}_1=(0,1)^\top$, $\boldsymbol{\mu}_2=(0,-1)^\top$, and

$$\boldsymbol{\Sigma}=\begin{pmatrix}1 & 0.8\\ 0.8 & 1\end{pmatrix}.$$

The standard $K$-means algorithm treats the data space as isotropic (distances unchanged by translations and rotations) (Raykov et al., 2016). This means that data points in each cluster are modeled as lying within a sphere around the cluster centroid. A sphere has the same radius in each dimension. However, the non-diagonal covariance matrix in the mixture model makes the cluster structure highly non-spherical. Thus, the $K$-means algorithm is expected to produce an unsatisfactory clustering result. This is illustrated in Figure 2 where $K$-means clustering clearly obtains wrong clusters. On the other hand, by incorporating the precision matrix estimation into clustering, our method is able to identify two correct clusters.

Figure 3 illustrates the estimation performance of precision matrices based on the clusters estimated from the $K$-means clustering and our method. Clearly, our SCAN method delivers an estimator with improved accuracy when compared to the two stage method which applies joint graphical lasso (JGL) to the clusters obtained from the $K$-means clustering. This suggests that an accurate clustering is critical for the estimation performance of heterogeneous graphical models.

Figure 2: The first plot represents the true clusters shown in red and black in the example of Section 4.2. The middle and right plots show the clusters obtained from the standard $K$-means clustering (Kmeans) and our SCAN method.

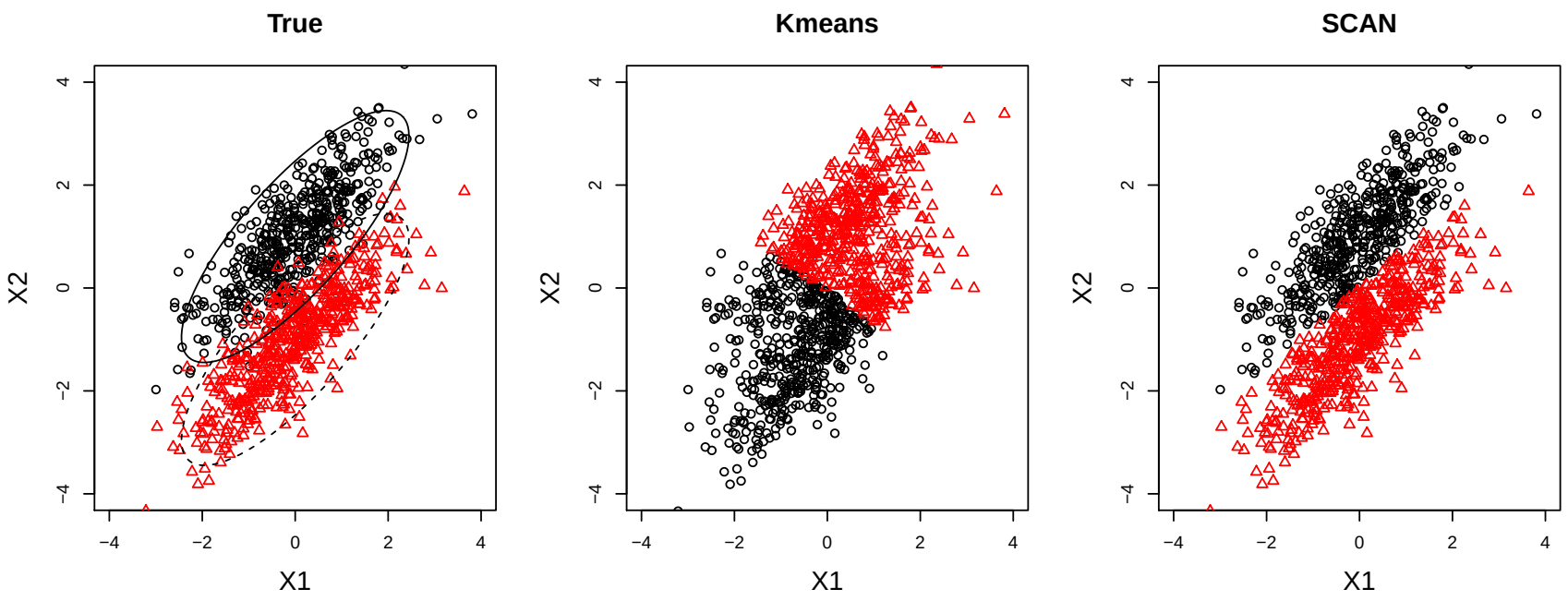

Figure 3: The true precision matrix and the estimated precision matrices from the two stage method (Kmeans + JGL) and our SCAN method in the example of Section 4.2.

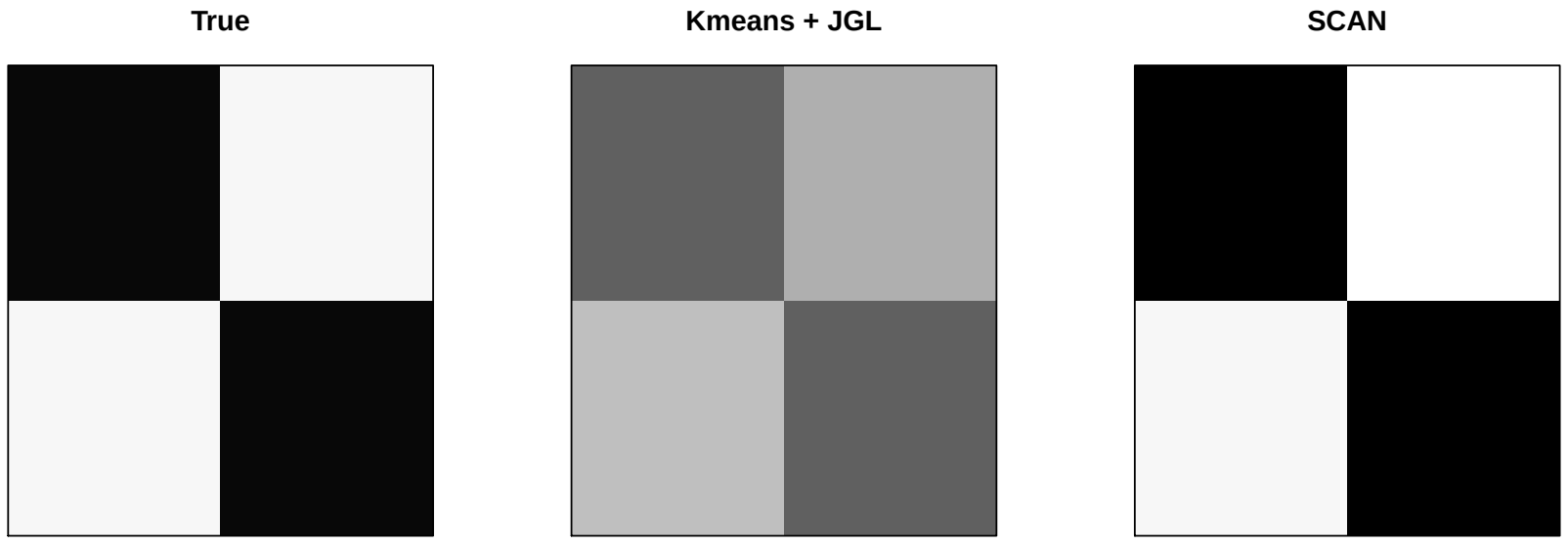

### 4.3 Effect of Sample Size and Dimension

We investigate the effect of sample size and dimension in terms of the estimation error and computational time. First, we empirically demonstrate the derived upper bound (3.15) for the estimation error by drawing the error pattern of our precision matrix estimator against sample size and dimension. The setting is the same as Section 4.2 except that we consider a tri-diagonal convariance structure. The results are summarized in Figure 4. In the first plot, we fix the dimension to be 10 and vary the sample size from 400 to 2000. In the second plot, we fix the sample size to be 5000 and vary the dimension from 5 to 50. The box plot refers to the the actual numerical values of precision matrix estimation errors, and the red dot is the theoretical error rate in each scenario. These results demonstrate that the empirical errors match very well with the theoretical error bound.

Figure 4: Comparison of the numerical error and the theoretical error rates of our SCAN method. The left panel displays the precision matrix estimation error with varying sample sizes. The right panel displays the precision matrix estimation error with varying dimensions.

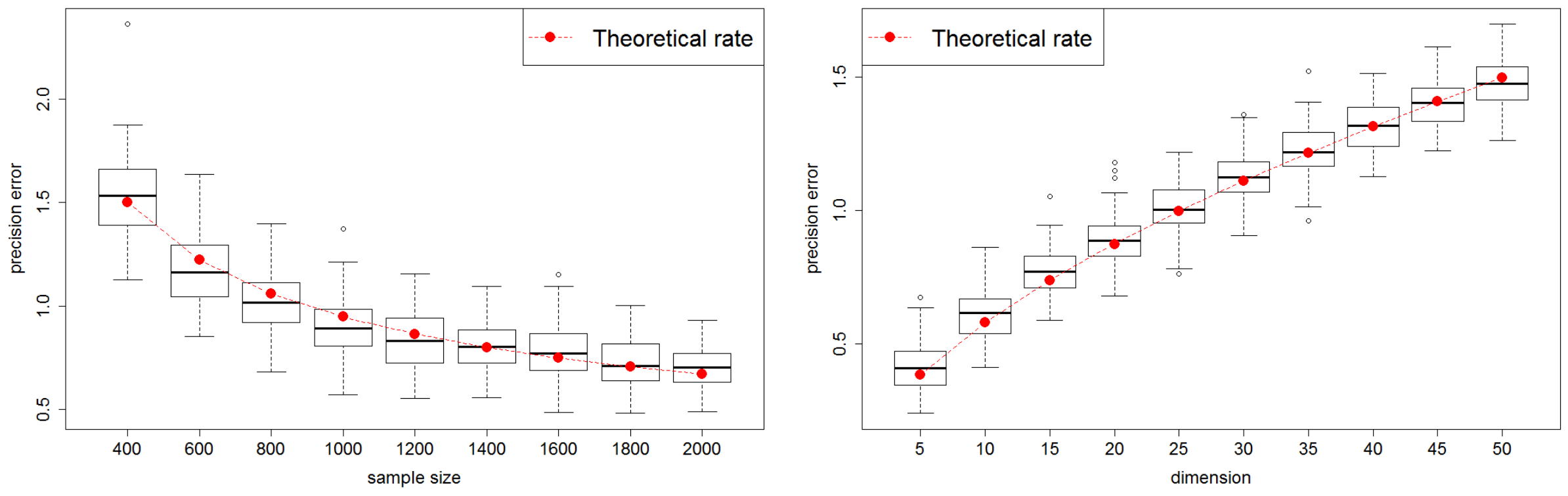

Second, we compare the average running time of our SCAN algorithm with varying sample sizes and dimensions. Figure 5 shows that our algorithm scales linearly with the sample size and roughly linearly with the dimension. This illustrates the efficiency and scalability of our proposed algorithm.

Figure 5: Running time of our algorithm. The left panel is the running time with varying sample sizes and fixed dimension $p = 10$. The right panel is the running time with varying dimensions and fixed sample size $n = 5000$.

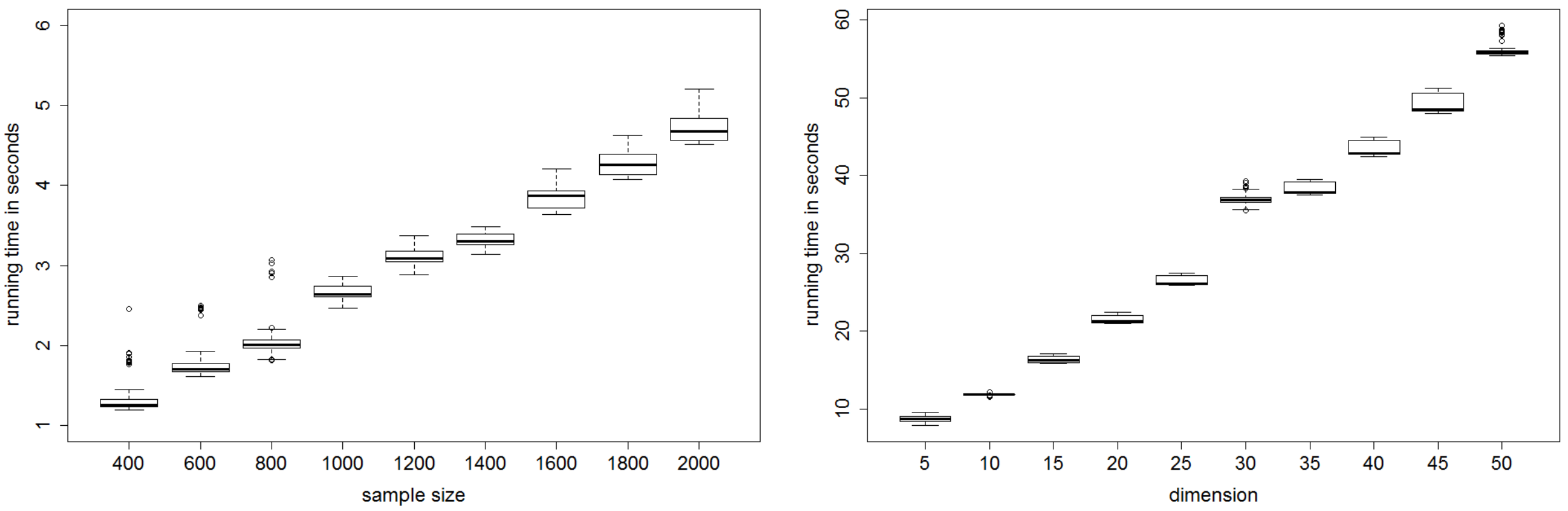

### 4.4 Simulations

In this subsection, we conduct extensive simulation studies to evaluate the performance of our algorithm. To assess the clustering performance of various methods, we compute the following clustering error (CE) which calculates the distance between an estimated clustering assignment $\widehat{\psi}$ and the true assignment $\psi$ of the sample data $\boldsymbol{X}_1, \ldots, \boldsymbol{X}_n$ (Wang, 2010; Sun et al., 2012),

$$\mathrm{CE}(\widehat{\psi}, \psi) := \binom{n}{2}^{-1} \left| \{(i,j) : \mathbb{1}(\widehat{\psi}(\boldsymbol{X}_i) = \widehat{\psi}(\boldsymbol{X}_j)) \neq \mathbb{1}(\psi(\boldsymbol{X}_i) = \psi(\boldsymbol{X}_j)); i < j\} \right|,$$

where $|\mathcal{A}|$ is the cardinality of set $\mathcal{A}$. To measure the estimation quality, we calculate the precision matrix error (PME) and cluster mean error (CME)

$$\mathrm{PME} := \frac{1}{K} \sum_{k=1}^{K} \left\| \widehat{\boldsymbol{\Omega}}^{(k)} - \boldsymbol{\Omega}^{(k)} \right\|_F; \;\; \mathrm{CME} := \frac{1}{K} \sum_{k=1}^{K} \left\| \widehat{\boldsymbol{\mu}}^{(k)} - \boldsymbol{\mu}^{(k)} \right\|_2.$$

Finally, to compare the variable selection performance, we compute the true positive rate (TPR, percentage of true edges selected) and the false positive rate (FPR, percentage of false edges selected)

$$\mathrm{TPR} := \frac{1}{K} \sum_{k=1}^{K} \frac{\sum_{i<j} \mathbb{1}(\omega_{kij} \neq 0, \widehat{\omega}_{kij} \neq 0)}{\sum_{i<j} \mathbb{1}(\omega_{kij} \neq 0)},$$

$$\mathrm{FPR} := \frac{1}{K} \sum_{k=1}^{K} \frac{\sum_{i<j} \mathbb{1}(\omega_{kij} = 0, \widehat{\omega}_{kij} \neq 0)}{\sum_{i<j} \mathbb{1}(\omega_{kij} = 0)}.$$

In the simulation, a three-class problem is considered. We illustrate three different types of network structures. In the first scenario, the network is assumed to have some regular structures. We generate a 5-block tridiagonal precision matrix with $p$ features for the precision matrix. To allow the similarity of precision matrices across clusters, we set the off-diagonal entry of $\Omega_1, \Omega_2, \Omega_3$ as $\eta$, $0.99\eta$, and $1.01\eta$, respectively. The diagonal entries of $\Omega_1, \Omega_2$, and $\Omega_3$ are all 1.

In the second and third scenarios, followed by Danaher et al. (2014), we simulate each network consisting of disjointed modules since many large networks in the real life exhibit a modular structure comprised of many disjointed or loosely connected components of relatively small size (Peng et al., 2009). Thus, each of three networks is generated with $p$ features, which has ten equally sized unconnected subnetworks. Among the ten subnetworks, eight have the same structure and edge values across all the three classes, one remains the same only for the first two classes and the last one appears only for the first class. For the cluster structure of subnetwork, we consider two scenarios: power-law network and chain network, which are generated using the algorithm in Peng et al. (2009) and Fan et al. (2009). The detail construction is described as below.

**Power-law network.** Given an undirected network structure above, the initial ten-block precision matrix $(w^1_{ij})_{p\times p}$ is generated by

$$w^1_{ij} = \begin{cases} 1 & i \neq j; \\ 0 & i \neq j, \text{ no edge;} \\ \text{Unif}([-0.4, -0.1] \cup [0.1, 0.4]) & i \neq j, \text{ edge exits;} \end{cases}$$

To ensure positive definiteness and symmetry, we divide each off-diagonal entry by 0.9 times the sum of the absolute values of off-diagonal entries in its row and average this rescaled matrix with its transpose. Denote the final transformed matrix by $\boldsymbol{A}$. The covariance matrix corresponding to the first class is created by

$$\boldsymbol{\Sigma}_{1ij} = d_{ij} \frac{\boldsymbol{A}^{-1}_{ij}}{\sqrt{\boldsymbol{A}^{-1}_{ii}\boldsymbol{A}^{-1}_{jj}}} \tag{4.2}$$

where $d_{ij} = 0.9$ for non-diagonal entry and $d_{ij} = 1$ for diagonal entry. For the covariance matrix corresponding to the second class, we create $\boldsymbol{\Sigma}_2$ be identical to $\boldsymbol{\Sigma}_1$ but reset one of ten block matrix to the identity matrix. Similarly, we reset one additional block matrix for $\boldsymbol{\Sigma}_3$.

**Chain network.** In the scenario, each of ten blocks of the first covariance matrix $\boldsymbol{\Sigma}_1$ is constructed in the following way. The $ij$-th element of each block has the form $\sigma_{ij} = \exp(-a|s_i - s_j|)$, where $s_1 < s_2 < \cdots < s_{p/10}$ for some $a > 0$. This is related to the autoregressive process of order one. In our case, we choose $a = 1$ and $s_i - s_{i-1} \sim \text{Unif}(0.5, 1)$ for $i = 2, \ldots, p/10$. Similarly, we create $\boldsymbol{\Sigma}_2$ be identical to $\boldsymbol{\Sigma}_1$ but reset one of ten block matrix to the identity matrix and reset one additional block matrix for $\boldsymbol{\Sigma}_3$.

After the networks are constructed, the samples are generated as follows. First, the cluster membership $Y_i$'s are uniformly sampled from $\{1, 2, 3\}$. Given the cluster label, we generate each sample $\boldsymbol{X}_i \sim \mathcal{N}(\boldsymbol{\mu}(Y_i), \boldsymbol{\Sigma}(Y_i))$. Here, the cluster mean $\boldsymbol{\mu}(Y_i)$ is sparse, where its first 10 variables are of the form

$$(\mu\mathbf{1}_5^\top, -\mu\mathbf{1}_5^\top)^\top \mathbb{1}(Y_i = 1) + \mu\mathbf{1}_{10}\mathbb{1}(Y_i = 2) + (-\mu\mathbf{1}_5^\top, -\mu\mathbf{1}_5^\top)^\top \mathbb{1}(Y_i = 3),$$

with $\mathbf{1}_5$ being a 5-dimensional vector of all ones, and its last $p - 10$ variables are zeros. For the first scenario, we consider 3 simulation models with varying choices of $\mu$ and $\eta$:

- Model 1: $\mu = 0.8$ and $\eta = 0.3$,
- Model 2: $\mu = 1$ and $\eta = 0.3$,
- Model 3: $\mu = 1$ and $\eta = 0.4$.

Here $\mu$ controls the separability of the three clusters with larger $\mu$ corresponding to an easier clustering problem, and $\eta$ represents the similarity level of precision matrices across clusters. For the second and third scenarios, we considered three simulation models with sequential choices of $\mu$:

- Models 4,7: $\mu = 0.7$,
- Models 5,8: $\mu = 0.8$,
- Models 6,9: $\mu = 0.9$.

The number of features $p$ is equal to 100 and sample size is equal to 300. The results are averaged over 50 experiments. The code is written in R and implemented on an Intel Xeon-E5 processor with 64 GB of RAM. The average computation time for SCAN of a single run took one and half minute.

In the experiment, our method selected the tuning parameters via the BIC criterion in Section 4.1. For a fair comparison, we also used the same tuning parameters $\lambda_1, \lambda_2$ in Zhou et al. (2009), and the same $\lambda_2, \lambda_3$ in the joint graphical lasso penalty of the two-stage approach. We repeated the procedure 50 times and reported the averaged clustering errors, estimation errors, and variable selection errors for each method as well as their standard errors. Table 2 is for regular network, Table 3 is for power-law networks and Table 4 is for chain networks. As shown in Table 3 and Table 4, the standard $K$-means clustering method has the largest clustering error due to a violation of its diagonal covariance matrix assumption. This will result in poor estimation for multiple precision matrices. The method of Zhou et al. (2009) improves the clustering performance of the standard $K$-means by using a graphical lasso in the precision matrix estimation. However, it obtains a relatively large precision matrix estimation error and very bad false positive rate since it ignores the similarity across different precision matrices. In contrast, our SCAN algorithm achieves the best clustering accuracy and best precision matrix estimation accuracy for both scenarios. This is due to our simultaneous clustering and estimation strategy as well as the consideration of similarity of precision matrices across clusters. This experiment shows that a satisfactory clustering algorithm is critical to achieve accurate estimations of heterogeneous graphical models, and alternatively good estimation of the graphical model can also improve the clustering performance. This explains the success of our simultaneous method in terms of both clustering and graphical model estimation.

Table 2: Simulation results of regular network. The clustering errors (CE), cluster mean errors (CME), precision matrix errors (PME), true positive rates (TPR) and false positive rates (FPR) of precision matrix estimation of four methods. The minimal clustering error and minimal estimation error in each simulation are shown in bold.

| Models | Methods | CE | CME | PME | TPR /FPR |
|---|---|---|---|---|---|
| | $K$-means | $0.166_{0.011}$ | $2.256_{0.108}$ | NA | NA /NA |
| Model 1 | $K$-means + JGL | $0.166_{0.011}$ | $2.256_{0.108}$ | $8.206_{0.090}$ | $0.985_{0.001}$ /$0.023_{0.001}$ |
| $\mu = 0.8$ | Zhou et al. (2009) | $0.104_{0.007}$ | $1.190_{0.052}$ | $10.458_{0.0509}$ | $0.960_{0.002}$ /$0.107_{0.001}$ |
| $\eta = 0.3$ | SCAN | $\mathbf{0.071_{0.007}}$ | $\mathbf{1.120_{0.063}}$ | $\mathbf{7.620_{0.072}}$ | $\mathbf{0.993_{0.001}}$ /$\mathbf{0.022_{0.001}}$ |
| | $K$-means | $0.210_{0.009}$ | $3.428_{0.114}$ | NA | NA/NA |
| Model 2 | $K$-means + JGL | $0.210_{0.009}$ | $3.428_{0.114}$ | $12.099_{0.317}$ | $0.989_{0.001}$ /$0.039_{0.003}$ |
| $\mu = 1$ | Zhou et al. (2009) | $0.125_{0.012}$ | $1.860_{0.118}$ | $12.833_{0.253}$ | $0.993_{0.001}$ /$0.119_{0.006}$ |
| $\eta = 0.3$ | SCAN | $\mathbf{0.058_{0.012}}$ | $\mathbf{1.476_{0.145}}$ | $\mathbf{10.301_{0.332}}$ | $\mathbf{0.997_{0.001}}$ /$\mathbf{0.036_{0.002}}$ |
| | $K$-means | $0.021_{0.002}$ | $1.289_{0.013}$ | NA | NA /NA |
| Model 3 | $K$-means + JGL | $0.021_{0.002}$ | $1.289_{0.013}$ | $7.639_{0.061}$ | $0.993_{0.001}$ /$0.029_{0.002}$ |
| $\mu = 1$ | Zhou et al. (2009) | $0.021_{0.002}$ | $0.968_{0.018}$ | $10.115_{0.047}$ | $0.968_{0.001}$ /$0.106_{0.001}$ |
| $\eta = 0.4$ | SCAN | $\mathbf{0.014_{0.001}}$ | $\mathbf{0.956_{0.018}}$ | $\mathbf{7.614_{0.061}}$ | $\mathbf{0.993_{0.001}}$ /$\mathbf{0.029_{0.002}}$ |

### 4.5 Glioblastoma Cancer Data Analysis

In this section, we apply our simultaneous clustering and graphical model estimation method to a Glioblastoma cancer dataset. We aim to cluster the glioblastoma multiforme (GBM) patients and construct the gene regulatory network of each subtype in order to improve our understanding of the GBM disease.

The raw gene expression dataset measures 17814 levels of mRNA expression of 482 GBM patients. Each patient belongs to one of four subgroups of GBM: Classical, Mesenchymal, Neural, and Proneural (Verhaak et al., 2010). Although they are biologically different, these four subtypes share many similarities since they are all GBM diseases. For our analysis, we considered the 840 signature genes established by Verhaak et al. (2010). Following the preprocess procedures in Lee and Liu (2015), we excluded the genes with no subtype information or the genes with missing values. We then applied the sure independence screening analysis (Fan and Lv, 2008) to finally include 50 genes in our analysis. These 50 signature genes are highly distinctive for these four subtypes. In the analysis, we pretended that the subtype information of each patient was unknown and evaluated the clustering accuracy of various clustering methods by comparing the estimated groups with the true subtypes. In all methods, we fixed $K = 4$. Moreover, we set the tuning parameters $\lambda_1 = 0.065, \lambda_2 = 0.238$, and $\lambda_3 = 0.138$ in our SCAN algorithm. For a fair comparison, we also used the same $\lambda_1, \lambda_2$ in Zhou et al. (2009), and the same $\lambda_2, \lambda_3$ in the joint graphical lasso of the two-stage method.

Table 3: Simulation results of power-law network. The clustering errors (CE), cluster mean errors (CME), precision matrix errors (PME), true positive rates (TPR) and false positive rates (FPR) of precision matrix estimation of four methods. The minimal clustering error and minimal estimation error in each simulation are shown in bold.

| Models | Methods | CE | CME | PME | TPR /FPR |
|---|---|---|---|---|---|
| | $K$-means | $0.331_{0.007}$ | $3.282_{0.047}$ | NA | NA /NA |
| Model 4 | $K$-means + JGL | $0.331_{0.007}$ | $3.282_{0.047}$ | $49.516_{0.159}$ | $0.575_{0.002}$ /$0.034_{0.002}$ |
| $\mu = 0.7$ | Zhou et al. (2009) | $0.311_{0.006}$ | $2.494_{0.055}$ | $50.945_{0.164}$ | $0.578_{0.002}$ /$0.134_{0.002}$ |
| | SCAN | $\mathbf{0.283_{0.008}}$ | $\mathbf{2.385_{0.065}}$ | $\mathbf{48.845_{0.146}}$ | $0.577_{0.003}$ /$0.032_{0.002}$ |
| | $K$-means | $0.228_{0.010}$ | $2.777_{0.111}$ | NA | NA/NA |
| Model 5 | $K$-means + JGL | $0.228_{0.010}$ | $2.777_{0.111}$ | $48.601_{0.132}$ | $0.582_{0.002}$ /$0.044_{0.003}$ |
| $\mu = 0.8$ | Zhou et al. (2009) | $0.186_{0.011}$ | $1.837_{0.113}$ | $49.289_{0.122}$ | $0.584_{0.001}$ /$0.131_{0.001}$ |
| | SCAN | $\mathbf{0.156_{0.012}}$ | $\mathbf{1.789_{0.119}}$ | $\mathbf{47.729_{0.118}}$ | $0.583_{0.002}$ /$0.041_{0.002}$ |
| | $K$-means | $0.083_{0.010}$ | $1.624_{0.120}$ | NA | NA /NA |
| Model 6 | $K$-means + JGL | $0.083_{0.010}$ | $1.624_{0.120}$ | $46.879_{0.093}$ | $0.589_{0.002}$ /$0.070_{0.003}$ |
| $\mu = 0.9$ | Zhou et al. (2009) | $0.050_{0.002}$ | $1.003_{0.018}$ | $47.503_{0.003}$ | $0.591_{0.001}$ /$0.128_{0.001}$ |
| | SCAN | $\mathbf{0.045_{0.002}}$ | $\mathbf{1.003_{0.018}}$ | $\mathbf{46.356_{0.086}}$ | $0.589_{0.001}$ /$0.068_{0.003}$ |

Table 5 reported the clustering errors of all methods as well as the number of informative variables in the corresponding estimated means and precision matrices. The standard $K$-means clustering has the large clustering error due to its ignorance of the network structure in the precision matrices. Therefore, the consequent joint graphical lasso method of the network reconstruction is less reliable. The method in Zhou et al. (2009) performed even worse. This is because their method estimates each precision matrix individually without borrowing information from each other. In this gene network example, all of the four graphical models share many edges due to the commonality in the GBM diseases. Zhou et al. (2009)'s method may suffer from the small sample size. Our method is able to achieve the best clustering performance due to the procedure of simultaneous clustering and heterogeneous graphical model estimation.

To evaluate the ability of reconstructing gene regulatory network of each subtype, we report the four gene networks estimated from our SCAN method in Figure 1. The black lines are links shared in all subtypes, and the color lines are uniquely presented in some subtypes. Clearly, most edges are black lines, which indicates the common structure of all subtypes. For instance, the link between ZNF45 and ZNF134 is significant across all the four subtypes. Those two genes belong to ZNF gene family. They are known to play roles in making zinc finger proteins, which are regulatory proteins that are functional important to many cellulars. As they play roles in the same biological process, it is reasonable to expect this link is shared by all GBM subtypes. There are two links that shared by three subtypes except neural subtype: TNFRSF1B↔TRPM2, PTPRC↔ TRPM2. One link uniquely appears in Proneural subtype: ACTR1A ↔DWED and one link FBXO3↔HMG20B is uniquely shown in neural subtype. These findings agree with the existing results in Verhaak et al. (2010). It has been shown that the PTPRC is a well-described microglia marker and is highly exposed in the set of murine astrocytic samples which are strongly associated with the Mesenchymal group. In addition, TRPM2 and TNFRSF1B are shown frequently in the GOTERM category of Mesenchymal group but less likely to appear in Neural group. And FBXO3 is only significant in the cell part of neural subtype. Furthermore, ACTR1A is only found in the intracellular non-membrane-bound organelle and protein binding of Proneural subtype in the supplemental material of Verhaak et al. (2010). It would also be of interest to investigate unique gene links that were not discovered in existing literatures for better understanding of GBM diseases.

Table 4: Simulation results of chain network. The clustering errors (CE), cluster mean errors (CME), precision matrix errors (PME), true positive rates (TPR) and false positive rates (FPR) of precision matrix estimation of four methods. The minimal clustering error and minimal estimation error in each simulation are shown in bold.

| Models | Methods | CE | CME | PME | TPR /FPR |
|---|---|---|---|---|---|
| | $K$-means | $0.277_{0.005}$ | $2.705_{0.070}$ | NA | NA /NA |
| Model 7 | $K$-means + JGL | $0.277_{0.005}$ | $2.705_{0.070}$ | $25.608_{0.183}$ | $0.995_{0.000}$ /$0.033_{0.001}$ |
| $\mu = 0.7$ | Zhou et al. (2009) | $0.267_{0.006}$ | $1.815_{0.075}$ | $29.341_{0.109}$ | $0.991_{0.001}$ /$0.131_{0.002}$ |
| | SCAN | $\mathbf{0.231_{0.007}}$ | $\mathbf{1.652_{0.087}}$ | $\mathbf{25.110_{0.106}}$ | $0.991_{0.001}$ /$0.031_{0.001}$ |
| | $K$-means | $0.200_{0.008}$ | $2.124_{0.098}$ | NA | NA/NA |
| Model 8 | $K$-means + JGL | $0.200_{0.008}$ | $2.124_{0.098}$ | $24.499_{0.127}$ | $0.996_{0.000}$ /$0.042_{0.001}$ |
| $\mu = 0.8$ | Zhou et al. (2009) | $0.168_{0.004}$ | $1.055_{0.076}$ | $27.494_{0.121}$ | $0.995_{0.001}$ /$0.131_{0.001}$ |
| | SCAN | $\mathbf{0.140_{0.004}}$ | $\mathbf{1.046_{0.038}}$ | $\mathbf{23.804_{0.085}}$ | $0.996_{0.000}$ /$0.039_{0.001}$ |
| | $K$-means | $0.123_{0.005}$ | $1.465_{0.040}$ | NA | NA /NA |
| Model 9 | $K$-means + JGL | $0.123_{0.005}$ | $1.465_{0.040}$ | $23.663_{0.097}$ | $0.997_{0.000}$ /$0.044_{0.001}$ |
| $\mu = 0.9$ | Zhou et al. (2009) | $0.116_{0.003}$ | $1.031_{0.022}$ | $26.476_{0.090}$ | $0.996_{0.001}$ /$0.131_{0.001}$ |
| | SCAN | $\mathbf{0.098_{0.003}}$ | $\mathbf{1.025_{0.022}}$ | $\mathbf{23.425_{0.083}}$ | $0.998_{0.000}$ /$0.043_{0.002}$ |

Table 5: The clustering errors and the number of selected features in cluster mean and precision matrix of various methods in the Glioblastoma Cancer Data.

| Methods | Clustering Error | $\sum_k \|\widehat{\boldsymbol{\mu}}^{(k)}\|_0$ | $\sum_k \|\widehat{\boldsymbol{\Omega}}^{(k)}\|_0$ |
|---|---|---|---|
| $K$-means | 0.262 | 200 | NA |
| Zhou et al. (2009) | 0.336 | 106 | 1820 |
| $K$-means + JGL | 0.262 | 200 | 1360 |
| SCAN | **0.222** | 128 | 1452 |

# 5 Discussion

In this paper, we propose a new SCAN method for simultaneous clustering and estimation of heterogeneous graphical models with common structures. We describe the theoretical properties of SCAN and we show that the estimation error bound of our SCAN algorithm consists of statistical error and optimization error, which explicitly addresses the trade-off between statistical accuracy and computational complexity. In our experiments, the tuning parameters can be chosen via an efficient BIC-type criterion. For future work, it is of interest to investigate the model selection consistency of these tuning parameters and study the distributed implementation of ECM algorithm based on the work in (Wolfe et al., 2008).

# APPENDIX

In this section, we provide detailed proofs of key results: Theorem 3.9 and Corollary 3.14. The proofs of other lemmas and theorems are deferred to the online supplementary.

# A Proof of Theorem 3.9

First we introduce some notation. Recall the definition of support space $\mathcal{M}$ in (3.2). The orthogonal complement of support space $\mathcal{M}$, namely, is defined as the set

$$\mathcal{M}^{\perp} := \left\{\boldsymbol{\Theta}' \in \Xi \mid \langle \boldsymbol{V}, \boldsymbol{\Theta}' \rangle = 0 \text{ for all } \boldsymbol{V} \in \mathcal{M}\right\}.$$

The projection operator $\Pi_{\mathcal{M}}(\boldsymbol{\Theta}) : \Xi \to \Xi$ is defined as

$$\Pi_{\mathcal{M}}(\boldsymbol{\Theta}) := \arg\min_{\boldsymbol{V} \in \mathcal{M}} \|\boldsymbol{V} - \boldsymbol{\Theta}\|_2 .$$

To simplify the notation, we frequently use the shorthand $\boldsymbol{\Theta}_{\mathcal{M}} = \Pi_{\mathcal{M}}(\boldsymbol{\Theta})$ and $\boldsymbol{\Theta}_{\mathcal{M}^{\perp}} = \Pi_{\mathcal{M}^{\perp}}(\boldsymbol{\Theta})$.

In order to efficiently solve the high-dimensional regularized problem, we explore some good properties enjoyed by SCAN penalty in Lemma A.1 and Lemma A.2. Similar properties can be derived by any decomposable penalty.

**Lemma A.1.** The SCAN penalty $\mathcal{P}$ is convex and decomposable with respect to $(\mathcal{M}, \mathcal{M}^{\perp})$. In detail,

$$\mathcal{P}(\boldsymbol{\Theta}_1 + \boldsymbol{\Theta}_2) = \mathcal{P}(\boldsymbol{\Theta}_1) + \mathcal{P}(\boldsymbol{\Theta}_2), \text{ for any } \boldsymbol{\Theta}_1 \in \mathcal{M}, \boldsymbol{\Theta}_2 \in \mathcal{M}^{\perp}.$$

The dual norm of SCAN penalty $\mathcal{P}$ is given by

$$\mathcal{P}^*(\boldsymbol{\Theta}) := \max_{i,j,k,i\neq j} \left( M_1\sqrt{\mu_{kj}^2}, M_2\sqrt{\omega_{kij}^2}, M_3 \left( \sum_{k=1}^{K} \omega_{kij}^2 \right)^{1/2} \right). \tag{A.1}$$

**Proof of Lemma A.1:** The convexity of SCAN comes from the convexity of lasso penalty for cluster means and the convexity of group graphical lasso penalty for precision matrices. The decomposability and derivation of dual norm is obvious from the definition. Also see Wainwright (2014). ∎

**Lemma A.2.** For all vectors $\boldsymbol{\Theta}$ belonging to support space $\mathcal{M}$, $\mathcal{P}(\boldsymbol{\Theta}_{\mathcal{M}})$ satisfies the following inequality:

$$\mathcal{P}(\boldsymbol{\Theta}_{\mathcal{M}}) \leq \nu(\mathcal{M}) \left\| \boldsymbol{\Theta}_{\mathcal{M}} \right\|_2, \tag{A.2}$$

where $\nu(\mathcal{M}) = M_1\sqrt{Kd} + (M_2\sqrt{K} + M_3)\sqrt{s}$ is the *support space compatibility constant* defined in (3.12) .

**Proof of Lemma A.2:** The detailed proof of Lemma A.2 is discussed in S.V. ∎

Next lemma is a key step to establish our main theorem. It quantifies the estimation error in one iteration step. According to this lemma, one can precisely understand how the statistical error and optimization error accumulate with more and more iterations.

**Lemma A.3.** Suppose $\boldsymbol{\Theta}^*$ lies in the interior of $\Xi$. If $\boldsymbol{\Theta}^{(t-1)} \in \mathcal{B}_\alpha(\boldsymbol{\Theta}^*)$, with choice of $\lambda_n^{(t)} = \varepsilon + \tau\|\boldsymbol{\Theta}^{(t-1)} - \boldsymbol{\Theta}^*\|_2/\nu(\mathcal{M})$, final estimation error satisfies $\|\boldsymbol{\Theta}^{(t)} - \boldsymbol{\Theta}^*\|_2 \leq 6\nu(\mathcal{M})\lambda_n^{(t)}/\gamma$ with probability at least $1 - \delta'$ for all $t = 1, 2, \ldots$. Here $\tau$, $\lambda$ and $\nu(\mathcal{M})$ are defined in Lemma 3.3, Lemma 3.5 and Lemma A.2 accordingly.

**Proof of Lemma A.3:** Proof is postponed to section B.1. ∎

Equipped with Lemmas A.3, we are able to precisely quantify the final estimation error after $t$ iteration steps. This can be achieved by mathematical induction. For simplicity, define $\kappa := 6\tau/\gamma$. When $t=1$, we have $\boldsymbol{\Theta}^{(0)} \in \mathcal{B}_\alpha(\boldsymbol{\Theta}^*)$. Applying Lemma A.3 yields that

$$\begin{aligned}\left\|\boldsymbol{\Theta}^{(1)} - \boldsymbol{\Theta}^*\right\|_2 &\leq \frac{6\lambda_n^{(1)}\nu(\mathcal{M})}{\gamma} \\ &= \frac{6\nu(\mathcal{M})}{\gamma}\varepsilon + \kappa\left\|\boldsymbol{\Theta}^{(0)} - \boldsymbol{\Theta}^*\right\|_2.\end{aligned}$$

Suppose the following inequality is true for some $t \geq 1$,

$$\left\|\boldsymbol{\Theta}^{(t)} - \boldsymbol{\Theta}^*\right\|_2 \leq \frac{1-\kappa^t}{1-\kappa}\frac{6\nu(\mathcal{M})}{\gamma}\varepsilon + \kappa^t\left\|\boldsymbol{\Theta}^{(0)} - \boldsymbol{\Theta}^*\right\|_2,$$

with probability at least $1 - t\delta'$. We need to verify when $t = t+1$, the above inequality still holds. First, we show that $\boldsymbol{\Theta}^{(t)}$ is within a ball of $\boldsymbol{\Theta}^*$ with radius $\alpha$. Under the assumption that $\varepsilon \leq (1-\kappa)\alpha\gamma/(6\nu(\mathcal{M}))$ for sufficient large $n$, we have

$$\begin{aligned}\left\|\boldsymbol{\Theta}^{(t)} - \boldsymbol{\Theta}^*\right\|_2 &\leq \frac{1-\kappa^t}{1-\kappa}\frac{6\nu(\mathcal{M})}{\gamma}\frac{(1-\kappa)\alpha\gamma}{6\nu(\mathcal{M})} + \kappa^t\left\|\boldsymbol{\Theta}^{(0)} - \boldsymbol{\Theta}^*\right\|_2 \\ &\leq (1-\kappa^t)\alpha + \kappa^t\alpha = \alpha.\end{aligned}$$

Consequently, we have $\boldsymbol{\Theta}^{(t)} \in \mathcal{B}_\alpha(\boldsymbol{\Theta}^*)$. Applying Lemma A.3 with $t+1$ implies that

$$\begin{aligned}\left\|\boldsymbol{\Theta}^{(t+1)} - \boldsymbol{\Theta}^*\right\|_2 &\leq \frac{6\nu(\mathcal{M})\varepsilon}{\gamma} + \kappa\left\|\boldsymbol{\Theta}^{(t)} - \boldsymbol{\Theta}^*\right\|_2 \\ &\leq \frac{6\nu(\mathcal{M})\varepsilon}{\gamma} + \kappa\left(\frac{1-\kappa^t}{1-\kappa}\frac{6\nu(\mathcal{M})\varepsilon}{\gamma} + \kappa^t\left\|\boldsymbol{\Theta}^{(0)} - \boldsymbol{\Theta}^*\right\|_2\right) \\ &= \frac{1-\kappa^{t+1}}{1-\kappa}\frac{6\nu(\mathcal{M})\varepsilon}{\gamma} + \kappa^{t+1}\left\|\boldsymbol{\Theta}^{(0)} - \boldsymbol{\Theta}^*\right\|_2,\end{aligned}$$

with probability at least $1-(t+1)\delta'$. Therefore, we reach the conclusion that

$$\begin{aligned}\left\|\boldsymbol{\Theta}^{(t)} - \boldsymbol{\Theta}^*\right\|_2 &\leq \frac{1-\kappa^t}{1-\kappa}\frac{6\nu(\mathcal{M})}{\gamma}\varepsilon + \kappa^t\left\|\boldsymbol{\Theta}^{(0)} - \boldsymbol{\Theta}^*\right\|_2 \\ &\leq \frac{6\nu(\mathcal{M})\varepsilon}{(1-\kappa)\gamma} + \kappa^t\left\|\boldsymbol{\Theta}^{(0)} - \boldsymbol{\Theta}^*\right\|_2,\end{aligned}$$

with probability at least $1 - t\delta'$. This concludes the proof of Theorem 3.9. ■

## A.1 Proof of Corollary 3.14

It is worth to notice that sufficiently large iterations ensure that the optimization error will be dominated by statistical error finally as $\kappa < 1/2$. First we provide a stopping rule $T$. Plugging $\varepsilon_1, \varepsilon_2$ from (S.14) & (S.15) into statistical error part and letting $\delta = 1/p$, we have:

$$
\begin{aligned}
SE = & \frac{1}{1-\kappa}\frac{6}{\gamma}\left[\left(\sqrt{Kd} + (\sqrt{K}+1)\sqrt{s}\right)\left(CK\|\mathbf{\Omega}^*\|_\infty + C'K^{1.5}\right)\sqrt{\frac{\log p}{n}}\right] \\
& + \frac{1}{1-\kappa}\frac{6}{\gamma}\left[C''\sqrt{p}\sqrt{\frac{K^3\log p}{n}}\right].
\end{aligned}
$$

Note that under Condition 3.13, $K = o(p)$. Then $SE$ is simplified by

$$
SE \le \frac{6\widetilde{C}}{(1-\kappa)\gamma}\|\mathbf{\Omega}^*\|_\infty\left(\sqrt{Kd} + \sqrt{Ks+p}\right)\sqrt{\frac{K^3\log p}{n}},
$$

for some constant $\widetilde{C}$. For simplicity, let's denote

$$
\varphi(n,p,K) = \frac{6\widetilde{C}}{(1-\kappa)\gamma}\|\mathbf{\Omega}^*\|_\infty\left(\sqrt{Kd} + \sqrt{Ks+p}\right)\sqrt{\frac{K^3\log p}{n}}.
$$

Therefore, the bound (3.14) suggests a reasonable choice of the number of iterations. In particular, when

$$
t \ge T = \log_{1/\kappa}\left(\frac{\left\|\mathbf{\Theta}^{(0)} - \mathbf{\Theta}^*\right\|_2}{\varphi(n,p,K)}\right), \tag{A.3}
$$

the optimization error is dominated by statistical error. Final estimation error will be upper bounded by

$$
\left\|\mathbf{\Theta}^{(T)} - \mathbf{\Theta}^*\right\|_2 \le \frac{12\widetilde{C}}{(1-\kappa)\gamma}\left(\|\mathbf{\Omega}^*\|_\infty\sqrt{\frac{K^5 d\log p}{n}} + \|\mathbf{\Omega}^*\|_\infty\sqrt{\frac{K^3(Ks+p)\log p}{n}}\right),
$$

with probability at least $1 - T(26K^2 + 8K + 1)/p$. Plugging in the expression of $T$ in (A.3), the probability term is bounded by:

$$
\begin{aligned}
\frac{T(26K^2+8K+1)}{p} & \lesssim \frac{\log_{1/\kappa}\left(n/\left(\left(\sqrt{Kd} + \sqrt{Ks+p}\right)\sqrt{K^3\log p}\right)\right)K^2}{p} \\
& \lesssim \frac{K^2\log_{1/\kappa} n}{p}.
\end{aligned}
$$

Under Condition 3.13, $T(26K^2 + 8K + 1)/p$ goes to zero as $K$ and $p$ diverging. Putting pieces together, we have

$$
\left\|\mathbf{\Theta}^{(T)} - \mathbf{\Theta}^*\right\|_2 \le \frac{12\widetilde{C}}{(1-\kappa)\gamma}\left(\|\mathbf{\Omega}^*\|_\infty\sqrt{\frac{K^5 d\log p}{n}} + \|\mathbf{\Omega}^*\|_\infty\sqrt{\frac{K^3(Ks+p)\log p}{n}}\right),
$$

which implies

$$
\sum_{k=1}^K\left(\left\|\boldsymbol{\mu}_k^{(T)} - \boldsymbol{\mu}_k^*\right\|_2 + \left\|\mathbf{\Omega}_k^{(T)} - \mathbf{\Omega}_k^*\right\|_F\right) \le \frac{12\widetilde{C}}{(1-\kappa)\gamma}\left(\|\mathbf{\Omega}^*\|_\infty\sqrt{\frac{K^5 d\log p}{n}} + \|\mathbf{\Omega}^*\|_\infty\sqrt{\frac{K^3(Ks+p)\log p}{n}}\right),
$$

with probability converging to 1. It ends the proof of Corollary 3.14. ■

# B Proof of Key Lemmas

## B.1 Proof of Lemma A.3

We first consider an unsymmetrized version of $\mathbf{\Theta}^{(t)}$. Our proof makes use of the function $f : \Xi \to \mathbb{R}$ given by:

$$f(\Delta) := Q_n(\mathbf{\Theta}^* + \Delta|\mathbf{\Theta}^{(t-1)}) - Q_n(\mathbf{\Theta}^*|\mathbf{\Theta}^{(t-1)}) - \lambda_n^{(t)} \left(\mathcal{P}(\mathbf{\Theta}^* + \Delta) - \mathcal{P}(\mathbf{\Theta}^*)\right).$$

This function helps us evaluate the error between the iterative estimator $\mathbf{\Theta}^{(t)}$ and the true parameter $\mathbf{\Theta}^*$. In addition, we exploit the following fact:

$$\begin{cases} f(0) = 0 \\ f(\widehat{\Delta}) \geq 0 \text{ when } \widehat{\Delta} = \mathbf{\Theta}^{(t)} - \mathbf{\Theta}^*. \end{cases} \tag{B.1}$$

The second property is from the optimality of $\mathbf{\Theta}^{(t)}$ in terms of the sample version objective function. In detail,

$$\mathbf{\Theta}^{(t)} = \arg\max_{\mathbf{\Theta}'} Q_n(\mathbf{\Theta}'|\mathbf{\Theta}^{(t-1)}) - \lambda_n^{(t)} \mathcal{P}(\mathbf{\Theta}'). \tag{B.2}$$

Correspondingly, there is a classical result named self-consistency property for population version objective function in McLachlan and Krishnan (2007), which in detail is

$$\mathbf{\Theta}^* = \arg\max_{\mathbf{\Theta}'} Q(\mathbf{\Theta}'|\mathbf{\Theta}^*). \tag{B.3}$$

The whole proof follows two steps. In Step I, we show that $f(\Delta) < 0$ if $\|\Delta\|_2 = \xi$. Next in Step II, we show that the error term $\widehat{\Delta}$ must satisfy $\|\widehat{\Delta}\|_2 < \xi$ under the result in Step I.

**Step I**: we begin to establish an upper bound on $f(\Delta)$ over the set $\mathbb{C}(\xi) := \{\Delta : \|\Delta\|_2 = \xi\}$ for the chosen radius $\xi = 6\lambda_n^{(t)}\nu(\mathcal{M})/\gamma$. From the assumption on $n$, when $n$ is large enough,

$$\begin{aligned} \varepsilon &\leq \frac{(1-\kappa)\alpha\gamma}{6\nu(\mathcal{M})} \leq \frac{(2-\kappa)\alpha\gamma}{6\nu(\mathcal{M})}, \\ \frac{6\nu(\mathcal{M})\varepsilon}{\gamma} &\leq (2-\kappa)\alpha. \end{aligned}$$

On the other hand, as $\|\mathbf{\Theta}^{(t-1)} - \mathbf{\Theta}^*\|_2 \leq \alpha$, $\xi$ satisfies,

$$\xi = \frac{6\nu(\mathcal{M})\varepsilon}{\gamma} + \kappa \left\|\mathbf{\Theta}^{(t-1)} - \mathbf{\Theta}^*\right\|_2 \leq 2\alpha.$$

It is sufficient to show that $\mathbb{C}(\xi) \subseteq \mathbb{C} = \{\Delta | \|\Delta\|_2 \leq 2\alpha\}$. According to Lemma 3.5, replacing $\mathbf{\Theta}' - \mathbf{\Theta}^*$ by $\Delta$, then any $\Delta \in \mathbb{C}(\xi)$ enjoys restricted strong concavity property, which implies:

$$Q_n(\mathbf{\Theta}^* + \Delta|\mathbf{\Theta}^{(t-1)}) - Q_n(\mathbf{\Theta}^*|\mathbf{\Theta}^{(t-1)}) \leq \left\langle \nabla Q_n(\mathbf{\Theta}^*|\mathbf{\Theta}^{(t-1)}), \Delta \right\rangle - \frac{\gamma}{2}\|\Delta\|_2^2,$$

with probability at least $1-\delta$. Subtracting $\lambda_n^{(t)}(\mathcal{P}(\boldsymbol{\Theta}^*+\Delta)-\mathcal{P}(\boldsymbol{\Theta}^*))$ from both sides, we construct an upper bound of $f(\Delta)$ in the right side,

$$f(\Delta) \leq \underbrace{\left\langle \nabla Q_n(\boldsymbol{\Theta}^*|\boldsymbol{\Theta}^{(t-1)}), \Delta \right\rangle}_{(i)} - \lambda_n^{(t)} \underbrace{\left(\mathcal{P}(\boldsymbol{\Theta}^*+\Delta)-\mathcal{P}(\boldsymbol{\Theta}^*)\right)}_{(ii)} - \frac{\gamma}{2}\|\Delta\|_2^2.$$

**Bounding** $(i)$**:** Note that $Q_n$ is a sample version $Q$-function but $\boldsymbol{\Theta}^*$ comes from population version $Q$-function (B.3). So we use $\nabla Q(\boldsymbol{\Theta}^*|\boldsymbol{\Theta}^{(t-1)})$ as a bridge to connect the sample-based analysis and population-based analysis together.

$$\begin{aligned} (i) &\leq \left|\left\langle \nabla Q_n(\boldsymbol{\Theta}^*|\boldsymbol{\Theta}^{(t-1)}) - \nabla Q(\boldsymbol{\Theta}^*|\boldsymbol{\Theta}^{(t-1)}) \right.\right. \\ &\qquad\qquad \left.\left. + \nabla Q(\boldsymbol{\Theta}^*|\boldsymbol{\Theta}^{(t-1)}) - \nabla Q(\boldsymbol{\Theta}^*|\boldsymbol{\Theta}^*), \Delta \right\rangle\right| \\ &\leq \underbrace{\left|\left\langle \nabla Q_n(\boldsymbol{\Theta}^*|\boldsymbol{\Theta}^{(t-1)}) - \nabla Q(\boldsymbol{\Theta}^*|\boldsymbol{\Theta}^{(t-1)}), \Delta \right\rangle\right|}_{\text{Statistical Error(SE)}} \\ &\qquad\qquad + \underbrace{\left|\left\langle \nabla Q(\boldsymbol{\Theta}^*|\boldsymbol{\Theta}^{(t-1)}) - \nabla Q(\boldsymbol{\Theta}^*|\boldsymbol{\Theta}^*), \Delta \right\rangle\right|}_{\text{Optimization Error(OE)}}. \end{aligned}$$

Note that $\boldsymbol{\Theta}^*$ lies in the interior of $\Xi$. According to the self-consistency property (B.3), $\nabla Q(\boldsymbol{\Theta}^*|\boldsymbol{\Theta}^*) = 0$ which implies the first inequality holds. This decomposition for $(i)$ leads to the optimization error part and statistical error part.

For simplicity, we write $h(\boldsymbol{\Theta}^*|\boldsymbol{\Theta}^{(t-1)}) = \nabla Q_n(\boldsymbol{\Theta}^*|\boldsymbol{\Theta}^{(t-1)}) - \nabla Q(\boldsymbol{\Theta}^*|\boldsymbol{\Theta}^{(t-1)})$. Since the group graphical lasso penalty does not penalize the diagonal element, it is a semi-norm. Recall that both $\Delta$ and $h(\boldsymbol{\Theta}^*|\boldsymbol{\Theta}^{(t-1)})$ are $K(p^2+p)$ dimensional vectors. Then by the definition of $\mathcal{G}$ and $\mathcal{G}^c$ in (3.1), statistical error can be decomposed further by:

$$\begin{aligned} \text{SE} &\leq \left| \langle h(\boldsymbol{\Theta}^*|\boldsymbol{\Theta}^{(t-1)})_{\mathcal{G}^c}, \Delta_{\mathcal{G}^c}\rangle| + |\langle h(\boldsymbol{\Theta}^*|\boldsymbol{\Theta}^{(t-1)})_{\mathcal{G}}, \Delta_{\mathcal{G}}\rangle \right| \\ &\leq \left\| h(\boldsymbol{\Theta}^*|\boldsymbol{\Theta}^{(t-1)})_{\mathcal{G}^c} \right\|_{\mathcal{P}^*} \cdot \mathcal{P}(\Delta_{\mathcal{G}^c}) + \|h(\boldsymbol{\Theta}^*|\boldsymbol{\Theta}^{(t-1)})_{\mathcal{G}}\|_2 \cdot \|\Delta_{\mathcal{G}}\|_2 \\ &\leq \|h(\boldsymbol{\Theta}^*|\boldsymbol{\Theta}^{(t-1)})\|_{\mathcal{P}^*} \cdot \mathcal{P}(\Delta) + \|h(\boldsymbol{\Theta}^*|\boldsymbol{\Theta}^{(t-1)})_{\mathcal{G}}\|_2 \cdot \|\Delta\|_2. \end{aligned}$$

The second inequality comes from the generalized Cauchy-Schwarz inequality. After excluding the diagonal terms from precision matrices, $\mathcal{P}(\Delta_{\mathcal{G}^c})$ can be treated as a norm. The last inequality is because both the penalties $\mathcal{P}$ and $\mathcal{P}^*$ do not penalize the diagonal term of precision matrices. Under statistical error Condition 3.6,

$$\text{SE} \leq \varepsilon_1 \mathcal{P}(\Delta) + \varepsilon_2 \|\Delta\|_2, \tag{B.4}$$

with probability at least $1-(\delta_1+\delta_2)$.

On the other hand, from the assumption that $\mathbf{\Theta}^{(t-1)}$ is in the $\mathcal{B}_\alpha(\mathbf{\Theta}^*)$, we are able to apply the Gradient Stability condition in Lemma 3.3 to bound OE.

$$\begin{aligned} \text{OE} &\le \|\nabla Q(\mathbf{\Theta}^*|\mathbf{\Theta}^{(t-1)}) - \nabla Q(\mathbf{\Theta}^*|\mathbf{\Theta}^*)\|_2 \cdot \|\Delta\|_2 \\ &\le \tau\|\mathbf{\Theta}^{(t-1)} - \mathbf{\Theta}^*\|_2 \cdot \|\Delta\|_2. \end{aligned} \tag{B.5}$$

Therefore, putting (B.4) and (B.5) together, $(i)$ is upper bounded by

$$(i) \le \varepsilon_1 \mathcal{P}(\Delta) + \varepsilon_2\|\Delta\|_2 + \tau\|\mathbf{\Theta}^{(t-1)} - \mathbf{\Theta}^*\|_2 \cdot \|\Delta\|_2, \tag{B.6}$$

with probability at least $1 - (\delta_1 + \delta_2)$.

**Bounding** $(ii)$**:** The decomposability of SCAN penalty in Lemma A.1 implies $\mathcal{P}(\mathbf{\Theta}^* + \Delta) = \mathcal{P}(\mathbf{\Theta}^* + \Delta_{\mathcal{M}}) + \mathcal{P}(\Delta_{\mathcal{M}^\perp})$. By triangle inequality, it is sufficient to bound $(ii)$,

$$\begin{aligned} (ii) &= \mathcal{P}(\mathbf{\Theta}^* + \Delta_{\mathcal{M}}) + \mathcal{P}(\Delta_{\mathcal{M}^\perp}) - \mathcal{P}(\mathbf{\Theta}^*) \\ &\ge \mathcal{P}(\mathbf{\Theta}^*) - \mathcal{P}(\Delta_{\mathcal{M}}) + \mathcal{P}(\Delta_{\mathcal{M}^\perp}) - \mathcal{P}(\mathbf{\Theta}^*) \\ &= \mathcal{P}(\Delta_{\mathcal{M}^\perp}) - \mathcal{P}(\Delta_{\mathcal{M}}). \end{aligned} \tag{B.7}$$

Combining (B.6) and (B.7), $f(\Delta)$ is upper bounded by:

$$\begin{aligned} f(\Delta) \le{}& \varepsilon_1 \mathcal{P}(\Delta) + \varepsilon_2\|\Delta\|_2 + \tau\|\mathbf{\Theta}^{(t-1)} - \mathbf{\Theta}^*\|_2 \cdot \|\Delta\|_2 \\ &- \lambda_n^{(t)}\left(\mathcal{P}(\Delta_{\mathcal{M}^\perp}) - \mathcal{P}(\Delta_{\mathcal{M}})\right) - \frac{\gamma}{2}\|\Delta\|_2^2. \end{aligned}$$

Triangle inequality implies $\mathcal{P}(\Delta) \le \mathcal{P}(\Delta_{\mathcal{M}}) + \mathcal{P}(\Delta_{\mathcal{M}^\perp})$. After combining some terms, the right hand side above could be further bounded by:

$$\begin{aligned} f(\Delta) \le{}& -\frac{\gamma}{2}\|\Delta\|_2^2 + (\lambda_n^{(t)} + \varepsilon_1)\mathcal{P}(\Delta_{\mathcal{M}}) + (\varepsilon_1 - \lambda_n^{(t)})\mathcal{P}(\Delta_{\mathcal{M}^\perp}) \\ &+ \varepsilon_2\|\Delta\|_2 + \tau\|\mathbf{\Theta}^{(t-1)} - \mathbf{\Theta}^*\|_2 \cdot \|\Delta\|_2, \end{aligned} \tag{B.8}$$

with probability at least $1 - (\delta + \delta_1 + \delta_2)$. Let $\delta' = \delta + \delta_1 + \delta_2$. According to Lemma A.2, we have the inequality $\mathcal{P}(\Delta_{\mathcal{M}}) \le \nu(\mathcal{M})\|\Delta_{\mathcal{M}}\|_2$. By the definition of $\Pi_{\mathcal{M}}(\Delta)$, we have

$$\|\Delta_{\mathcal{M}}\|_2 = \|\Pi_{\mathcal{M}}(\Delta) - \Pi_{\mathcal{M}}(0)\|_2 \le \|\Delta - 0\|_2 = \|\Delta\|_2.$$

Then $\mathcal{P}(\Delta_{\mathcal{M}})$ is further bounded by

$$\mathcal{P}(\Delta_{\mathcal{M}}) \le \nu(\mathcal{M})\|\Delta\|_2. \tag{B.9}$$

Substituting (B.9) into (B.8), we obtain:

$$\begin{aligned} f(\Delta) \le{}& \left(\varepsilon_1 + \frac{\varepsilon_2 + \tau\|\mathbf{\Theta}^{(t-1)} - \mathbf{\Theta}^*\|_2}{\nu(\mathcal{M})}\right)\nu(\mathcal{M})\|\Delta\|_2 - \frac{\gamma}{2}\|\Delta\|_2^2 \\ &+\lambda_n^{(t)}\nu(\mathcal{M})\|\Delta\|_2 + (\varepsilon_1 - \lambda_n^{(t)})\mathcal{P}(\Delta_{\mathcal{M}^\perp}), \end{aligned}$$

with at least probability $1-\delta'$. Recall that we choose

$$\lambda_n^{(t)} = \varepsilon + \frac{\tau\|\boldsymbol{\Theta}^{(t-1)} - \boldsymbol{\Theta}^*\|_2}{\nu(\mathcal{M})}, \epsilon = \epsilon_1 + \frac{\epsilon_2}{\nu(\mathcal{M})}.$$

From the construction of $\lambda_n^{(t)}$, the inequality $\varepsilon_1 - \lambda_n^{(t)} < 0$ always holds. Therefore, the upper bound for $f(\Delta)$ can be simplified by

$$\begin{aligned} f(\Delta) &\leq -\frac{\gamma}{2}\|\Delta\|_2^2 + 2\lambda_n^{(t)}\nu(\mathcal{M})\|\Delta\|_2 \\ &= -\frac{6(\lambda_n^{(t)}\nu(\mathcal{M}))^2}{\gamma} < 0. \end{aligned}$$

where the above equality is due to $\Delta \in \mathbb{C}(\xi)$. Now we reach the conclusion that $f(\Delta) < 0$ for all vectors $\Delta \in \mathbb{C}(\xi)$.

**Step II**: Now we start to prove the following statement: if for some optimal solution $\boldsymbol{\Theta}^{(t)}$ in (B.2), the corresponding error term $\widehat{\Delta} = \boldsymbol{\Theta}^{(t)} - \boldsymbol{\Theta}^*$ satisfies the inequality $\|\widehat{\Delta}\|_2 > \xi$, there must exist some vectors $\widetilde{\Delta}$ which belong to $\mathbb{C}(\xi)$ such that $f(\widetilde{\Delta}) \geq 0$. Before our forward proofs, let's state a lemma which describe the curvature of function $Q_n(\cdot|\boldsymbol{\Theta}^{(t-1)})$.

**Lemma B.1.** $Q_n(\cdot|\boldsymbol{\Theta}^{(t-1)})$ satisfies the following inequality a.s.:

$$Q_n(\boldsymbol{\Theta}^{(1)}|\boldsymbol{\Theta}^{(t-1)}) - Q_n\left(\boldsymbol{\Theta}^{(2)}|\boldsymbol{\Theta}^{(t-1)}\right) \leq \left\langle \nabla Q_n\left(\boldsymbol{\Theta}^{(2)}|\boldsymbol{\Theta}^{(t-1)}\right), \boldsymbol{\Theta}^{(1)} - \boldsymbol{\Theta}^{(2)}\right\rangle.$$

when $(\boldsymbol{\Theta}^{(1)}, \boldsymbol{\Theta}^{(2)}) = (\boldsymbol{\Theta}^{(t)}, t^*\boldsymbol{\Theta}^{(t)} + (1-t^*)\boldsymbol{\Theta}^*)$ or $(\boldsymbol{\Theta}^*, t^*\boldsymbol{\Theta}^{(t)} + (1-t^*)\boldsymbol{\Theta}^*)$.

**Proof of Lemma B.1:** The detailed proof of Lemma B.1 is discussed in S.VI. ■

The lemma tells us that we only require sample-based $Q$-function to be point-wise concave rather than global concave. If $\|\widehat{\Delta}\|_2 > \xi$, then the line joining $\widehat{\Delta}$ to 0 must intersect the set $\mathbb{C}(\xi)$ at some intermediate points $t^*\widehat{\Delta}$, for some $t^* \in (0,1)$. According to Lemma B.1,

$$\begin{aligned} &Q_n(\boldsymbol{\Theta}^{(t)}|\boldsymbol{\Theta}^{(t-1)}) - Q_n(t^*\boldsymbol{\Theta}^{(t)} + (1-t^*)\boldsymbol{\Theta}^*|\boldsymbol{\Theta}^{(t-1)}) \\ &\qquad \leq \left\langle \nabla Q_n(t^*\boldsymbol{\Theta}^{(t)} + (1-t^*)\boldsymbol{\Theta}^*|\boldsymbol{\Theta}^{(t-1)}), (1-t^*)(\boldsymbol{\Theta}^{(t)} - \boldsymbol{\Theta}^*)\right\rangle \\ &Q_n(\boldsymbol{\Theta}^*|\boldsymbol{\Theta}^{(t-1)}) - Q_n\left(t^*\boldsymbol{\Theta}^{(t)} + (1-t^*)\boldsymbol{\Theta}^*|\boldsymbol{\Theta}^{(t-1)}\right) \\ &\qquad \leq \left\langle \nabla Q_n\left(t^*\boldsymbol{\Theta}^{(t)} + (1-t^*)\boldsymbol{\Theta}^*|\boldsymbol{\Theta}^{(t-1)}\right), -t^*(\boldsymbol{\Theta}^{(t)} - \boldsymbol{\Theta}^*)\right\rangle. \end{aligned}$$

Adding the above two inequalities together with proper scaling, we can get

$$t^*Q_n(\boldsymbol{\Theta}^{(t)}|\boldsymbol{\Theta}^{(t-1)}) + (1-t^*)Q_n(\boldsymbol{\Theta}^*|\boldsymbol{\Theta}^{(t-1)}) \leq Q_n(t^*\boldsymbol{\Theta}^{(t)} + (1-t^*)\boldsymbol{\Theta}^*|\boldsymbol{\Theta}^{(t-1)}).$$

According to the convexity of $\mathcal{P}(\boldsymbol{\Theta})$,

$$\mathcal{P}\left(\boldsymbol{\Theta}^* + t^*\widehat{\Delta}\right) - \mathcal{P}\left(\boldsymbol{\Theta}^*\right) = \mathcal{P}\left(t^*\boldsymbol{\Theta}^{(t)} + (1-t^*)\boldsymbol{\Theta}^*\right) - \mathcal{P}(\boldsymbol{\Theta}^*)$$
$$\leq t^*\mathcal{P}(\boldsymbol{\Theta}^{(t)}) + (1-t^*)\mathcal{P}(\boldsymbol{\Theta}^*) - \mathcal{P}(\boldsymbol{\Theta}^*) = t^*\left(\mathcal{P}(\boldsymbol{\Theta}^{(t)}) - \mathcal{P}(\boldsymbol{\Theta}^*)\right).$$

Putting the above pieces together, it is shown that

$$\begin{aligned} f(t^*\widehat{\Delta}) \quad &= Q_n\left(t^*\boldsymbol{\Theta}^{(t)} + (1-t^*)\boldsymbol{\Theta}^*|\boldsymbol{\Theta}^{(t-1)}\right) - Q_n\left(\boldsymbol{\Theta}^*|\boldsymbol{\Theta}^{(t-1)}\right) \\ &\qquad - \lambda_n^{(t)}\left(\mathcal{P}(\boldsymbol{\Theta}^* + \Delta) - \mathcal{P}(\boldsymbol{\Theta}^*)\right) \\ &\geq t^*\left(Q_n(\boldsymbol{\Theta}^{(t)}|\boldsymbol{\Theta}^{(t-1)}) - Q_n(\boldsymbol{\Theta}^*|\boldsymbol{\Theta}^{(t-1)}) - \lambda_n^{(t)}\left(\mathcal{P}(\boldsymbol{\Theta}^{(t)}) - \mathcal{P}(\boldsymbol{\Theta}^*)\right)\right) \\ &= t^* f(\widehat{\Delta}). \end{aligned}$$

On the other hand, the optimality property (B.1) guarantees $f(\widehat{\Delta}) \geq 0$, and hence $f(t^*\widehat{\Delta}) \geq 0$ as well. Thus, we have constructed a vector $\widetilde{\Delta} = t^*\widehat{\Delta}$ with the claimed properties. This proves the statement in the beginning of Step II. Therefore, combining with the result in Step I, the contradiction of the statement in Step II implies that

$$\left\|\boldsymbol{\Theta}^{(t)} - \boldsymbol{\Theta}^*\right\|_2 \leq \xi = \frac{6\lambda_n^{(t)}\nu(\mathcal{M})}{\gamma}, \tag{B.10}$$

with probability at least $1 - \delta'$. This concludes the proof of Lemma A.3. ■

## References

Balakrishnan, S., Wainwright, M. J. and Yu, B. (2016). Statistical guarantees for the em algorithm: From population to sample-based analysis. *Annals of Statistics* To appear.

Bickel, P. and Levina, E. (2008). Covariance regularization by thresholding. *Annals of Statistics* **36** 2577–2604.

Boyd, S., Parikh, N., Chu, E., Peleato, B. and Eckstein, J. (2011). Distributed optimization and statistical learning via the alternating direction method of multipliers. *Foundations and Trends in Machine Learning* **3** 1–122.

Cai, T., Liu, W. and Luo, X. (2011). A constrained 1 minimization approach to sparse precision matrix estimation. *Journal of the American Statistical Association* **106** 594–607.

Cai, T. T., Li, H., Liu, W. and Xie, J. (2016a). Joint estimation of multiple high-dimensional precision matrices. *Statistica Sinica* **26** 445–464.

Cai, T. T., Liu, W. and Zhou, H. H. (2016b). Estimating sparse precision matrix: Optimal rates of convergence and adaptive estimation. *Ann. Statist.* **44** 455–488.

Chen, J. (1995). Optimal rate of convergence for finite mixture models. *The Annals of Statistics* 221–233.

Chen, Y., Pavlov, D. and Canny, J. (2009). Large-scale behavioral targeting. In *ACM SIGKDD*.

Danaher, P., Wang, P. and Witten, D. M. (2014). The joint graphical lasso for inverse covariance estimation across multiple classes. *Journal of the Royal Statistical Society: Series B* **76** 373–397.

Fan, J., Feng, Y. and Wu, Y. (2009). Network exploration via the adaptive lasso and scad penalties. *The Annals of Applied Statistics* **3** 521.

Fan, J. and Lv, J. (2008). Sure independence screening for ultrahigh dimensional feature space. *Journal of the Royal Statistical Society: Series B* **70** 849–911.

Friedman, J., Hastie, H. and Tibshirani, R. (2008). Sparse inverse covariance estimation with the graphical lasso. *Biostatistics* **9** 432–441.

Gao, C., Zhu, Y., Shen, X. and Pan, W. (2016). Estimation of multiple networks in gaussian mixture models. *Electronic Journal of Statistics* **10** 1133–1154.

Guo, J., Levina, E., Michailidis, G. and Zhu, J. (2011). Joint estimation of multiple graphical models. *Biometrika* **98** 1–15.

Ho, N. and Nguyen, X. (2015). Identifiability and optimal rates of convergence for parameters of multiple types in finite mixtures. *arXiv preprint arXiv:1501.02497* .

Horn, R. A. and Johnson, C. R. (1988). *Matrix Analysis*. New York: Cambridge Univ. Press.

Jeziorski, P. and Segal, I. (2015). What makes them click: Empirical analysis of consumer demand for search advertising. *American Economic Journal* **7** 24–53.

Lauritzen, S. (1996). *Graphical Models*. Oxford Science Publications.

Lee, W. and Liu, Y. (2015). Joint estimation of multiple precision matrices with common structures. *Journal of Machine Learning Research* **16** 1035–1062.

MacQueen, J. (1967). Some methods for clasification and analysis of multivariate observations. *In Proceedings of the Fifth Berkeley Symposium on Mathematical Statistics and Probability* 281–297.

McLachlan, G. and Krishnan, T. (2007). *The EM Algorithm and Extensions*. Wiley Series in Probability and Statistics.

Meng, X.-L. and Rubin, D. B. (1993). Maximum likelihood estimation via the ecm algorithm: A general framework. *Biometrika* **80** 267–278.

Negahban, S. N., Ravikumar, P., Wainwright, M. J. and Yu, B. (2012). A unified framework for high-dimensional analysis of $m$-estimators with decomposable regularizers. *Statist. Sci.* **27** 538–557.

Nguyen, X. (2013). Convergence of latent mixing measures in finite and infinite mixture models. *The Annals of Statistics* **41** 370–400.

Pan, W. and Shen, X. (2007). Penalized model-based clustering with application to variable selection. *Journal of Machine Learning Research* **8** 1145–1164.

Peng, J., Wang, P., Zhou, N. and Zhu, J. (2009). Partial correlation estimation by joint sparse regression models. *Journal of the American Statistical Association* **104** 735–746.

Peterson, C., Stingo, F. C. and Vannucci, M. (2015). Bayesian inference of multiple gaussian graphical models. *Journal of the American Statistical Association* **110** 159–174.

Qiu, H., Han, F., Liu, H. and Caffo, B. (2016). Joint estimation of multiple graphical models from high dimensional time series. *Journal of the Royal Statistical Society: Series B* **78** 487–504.

Raykov, Y. P., Boukouvalas, A., Baig, F. and Little, M. A. (2016). What to do when k-means clustering fails: a simple yet principled alternative algorithm. *PloS one* **11** e0162259.

Rothman, A. J. and Forzani, L. (2014). On the existence of the weighted bridge penalized gaussian likelihood precision matrix estimator. *Electronic Journal of Statistics* **8** 2693–2700.

Saegusa, T. and Shojaie, A. (2016). Joint estimation of precision matrices in heterogeneous populations. *Electronic Journal of Statistics* To appear.

Shen, X., Pan, W. and Zhu, Y. (2012). Likelihood-based selection and sharp parameter estimation. *Journal of the American Statistical Association* **107** 223–232.

Shojaie, A. and Michailidis, G. (2010a). Penalized likelihood methods for estimation of sparse high-dimensional directed acyclic graphs. *Biometrika* **97** 519–538.

Shojaie, A. and Michailidis, G. (2010b). Penalized principal component regression on graphs for analysis of subnetworks. *Advances in Neural Information Processing Systems* 2155–2163.

Sun, W., Wang, J. and Fang, Y. (2012). Regularized k-means clustering of high-dimensional data and its asymptotic consistency. *Electron. J. Statist.* **6** 148–167.

TCGA (2008). Comprehensive genomic characterization defines human glioblastoma genes and core pathways. *Nature* **455** 1061–1068.

Verhaak, R. G., Hoadley, K. A., Purdom, E., Wang, V., Qi, Y., Wilkerson, M. D., Miller, C. R., Ding, L., Golub, T., Mesirov, J. P., Alexe, G., Lawrence, M., OKelly, M., Tamayo, P., Weir, B. A., Gabriel, S., Winckler, W., Gupta, S., Jakkula, L., Feiler, H. S., Hodgson, J. G., James, C. D., Sarkaria, J. N., Brennan, C., Kahn, A., Spellman, P. T., Wilson, R. K., Speed, T. P., Gray, J. W., Meyerson, M., Getz, G., Perou, C. M., Hayes, D. N. and TCGA (2010). Integrated genomic analysis identifies clinically relevant subtypes of glioblastoma characterized by abnormalities in pdgfra, idh1, egfr, and nf1. *Cancer Cell* **17** 98–110.

Vershynin, R. (2012). *Compressed sensing*, chap. Introduction to the non-asymptotic analysis of random matrices. Cambridge Univ. Press, 210–268.

Wainwright, M. J. (2014). Structured regularizers for high-dimensional problems: Statistical and computational issues. *Annual Review of Statistics and Its Application* **1** 233–253.

Wang, J. (2010). Consistent selection of the number of clusters via crossvalidation. *Biometrika* **97** 893–904.

Wang, J. (2015). Joint estimation of sparse multivariate regression and conditional graphical models. *Statistica Sinica* **25** 831–851.

Wang, P., Sun, W., Yin, D., Yang, J. and Chang, Y. (2015a). Robust tree-based causal inference for complex ad effectiveness analysis. In *Proceedings of 8th ACM Conference on Web Search and Data Mining.*

Wang, Z., Gu, Q., Ning, Y. and Liu, H. (2015b). High dimensional em algorithm: Statistical optimization and asymptotic normality. *Advances in Neural Information Processing Systems* 2512–2520.

Wolfe, J., Haghighi, A. and Klein, D. (2008). Fully distributed em for very large datasets. *The International Conference on Machine Learning* 1184–1191.

Yan, J., Liu, N., Wang, G., Zhang, W., Jiang, Y. and Chen, Z. (2009). How much can behavioral targeting help online advertising? In *International ACM WWW Conference.*

Yi, X. and Caramanis, C. (2015). Regularized em algorithms: A unified framework and statistical guarantees. *Advances in Neural Information Processing Systems* 1567–1575.

Yuan, M. and Lin, Y. (2007). Model selection and estimation in the gaussian graphical model. *Biometrika* **94** 19–35.

Zhang, Y., Chen, X., Zhou, D. and Jordan, M. I. (2014). Spectral methods meet em: A provably optimal algorithm for crowdsourcing. *Advances in Neural Information Processing Systems* 1260–1268.

Zhou, H., Pan, W. and Shen, X. (2009). Penalized model-based clustering with unconstrained covariance matrices. *Electron. J. Statist.* **3** 1473–1496.

Zhu, Y., Shen, X. and Pan, W. (2014). Structural pursuit over multiple undirected graphs. *Journal of the American Statistical Association* **109** 1683–1696.

*Online Supplementary*

Simultaneous Clustering and Estimation of Heterogeneous Graphical Models

This supplementary contains supporting lemmas and their proofs for the theoretical developments in the main paper.

# A Proof of Several Lemmas and Selection Consistency

## S.I Proof of Lemma 3.1

The result follows by setting the derivative of $Q(\mathbf{\Theta}^{'}|\mathbf{\Theta})$ with respect to $\boldsymbol{\mu}_k^{'}$ or $\mathbf{\Omega}_k^{'}$ as zero. In particular, solving

$$\frac{\partial Q(\mathbf{\Theta}^{'}|\mathbf{\Theta})}{\partial \boldsymbol{\mu}_k^{'}} = \mathbb{E}[L_{\mathbf{\Theta},k}(\boldsymbol{X})\mathbf{\Omega}_k^{'}(\boldsymbol{X}-\boldsymbol{\mu}_k^{'})] = 0,$$

implies that

$$\arg\max_{\boldsymbol{\mu}_k^{'}} Q(\mathbf{\Theta}^{'}|\mathbf{\Theta}) = \frac{[\mathbf{\Omega}_k^{'}]^{-1}\mathbb{E}[L_{\mathbf{\Theta},k}(\boldsymbol{X})\mathbf{\Omega}_k^{'}\boldsymbol{X}]}{\mathbb{E}[L_{\mathbf{\Theta},k}(\boldsymbol{X})]} = \frac{\mathbb{E}[L_{\mathbf{\Theta},k}(\boldsymbol{X})\boldsymbol{X}]}{\mathbb{E}[L_{\mathbf{\Theta},k}(\boldsymbol{X})]}.$$

Similarly, solving

$$\frac{\partial Q(\mathbf{\Theta}^{'}|\mathbf{\Theta})}{\partial \mathbf{\Omega}_k^{'}} = \frac{1}{2}\mathbb{E}[L_{\mathbf{\Theta},k}(\boldsymbol{X})][\mathbf{\Omega}_k^{'}]^{-1} - \frac{1}{2}\mathbb{E}[L_{\mathbf{\Theta},k}(\boldsymbol{X})(\boldsymbol{X}-\boldsymbol{\mu}_k^{'})(\boldsymbol{X}-\boldsymbol{\mu}_k^{'})^\top] = 0,$$

implies (3.6). This ends the proof of Lemma 3.1. ■

## S.II Proof of Lemma 3.3

We consider $k$-th group first

$$\left\|\nabla_{\mathbf{\Theta}_k^{'}} Q(\boldsymbol{\mu}_k^*, \mathbf{\Omega}_k^*|\mathbf{\Theta}^*) - \nabla_{\mathbf{\Theta}_k^{'}} Q(\boldsymbol{\mu}_k^*, \mathbf{\Omega}_k^*|\mathbf{\Theta})\right\|_2 \le \tau \left\|\mathbf{\Theta}-\mathbf{\Theta}^*\right\|_2, \tag{S.1}$$

for any $\mathbf{\Theta} \in \mathbb{B}_\alpha(\mathbf{\Theta}^*)$. Remind that $\mathbf{\Theta}_k^{'} = \text{vec}(\boldsymbol{\mu}_k, \mathbf{\Omega}_k) \in \mathbb{R}^{p^2+p}$. According to the derivation in the proof of Lemma 3.1, we have

$$\nabla_{\mathbf{\Theta}_k^{'}} Q(\mathbf{\Theta}_k^{'}|\mathbf{\Theta}) = \begin{pmatrix} \mathbb{E}\left[L_{\mathbf{\Theta},k}(\boldsymbol{X})\mathbf{\Omega}_k^{'}(\boldsymbol{X}-\boldsymbol{\mu}_k^{'})\right] \\ \text{vec}\left\{\frac{1}{2}\mathbb{E}[L_{\mathbf{\Theta},k}(\boldsymbol{X})]\mathbf{\Omega}_k^{'-1} - \frac{1}{2}\mathbb{E}[L_{\mathbf{\Theta},k}(\boldsymbol{X})(\boldsymbol{X}-\boldsymbol{\mu}_k^{'})(\boldsymbol{X}-\boldsymbol{\mu}_k^{'})^\top]\right\}^\top \end{pmatrix}.$$

Define $D_L(\mathbf{\Theta}^*, \mathbf{\Theta}) = L_{\mathbf{\Theta}^*,k}(\boldsymbol{X}) - L_{\mathbf{\Theta},k}(\boldsymbol{X})$. Therefore, the square of the left hand side of ($S.1$) can be simplified to

$$\begin{aligned} & \left\|\nabla_{\mathbf{\Theta}_k'} Q(\boldsymbol{\mu}_k^*, \mathbf{\Omega}_k^* | \mathbf{\Theta}^*) - \nabla_{\mathbf{\Theta}_k'} Q(\boldsymbol{\mu}_k^*, \mathbf{\Omega}_k^* | \mathbf{\Theta})\right\|_2^2 \\ = & \underbrace{\|\mathbb{E}\left[D_L(\mathbf{\Theta}^*, \mathbf{\Theta}) \mathbf{\Omega}_k^* (\boldsymbol{X} - \boldsymbol{\mu}_k^*)\right]\|_2^2}_{I} \\ & + \underbrace{\left\| \frac{1}{2} \mathbb{E}[D_L(\mathbf{\Theta}^*, \mathbf{\Theta}) \mathbf{\Omega}_k^{*-1} - \frac{1}{2} \mathbb{E}[D_L(\mathbf{\Theta}^*, \mathbf{\Theta})(\boldsymbol{X} - \boldsymbol{\mu}_k^*)(\boldsymbol{X} - \boldsymbol{\mu}_k^*)^\top] \right\|_F^2}_{II}. \end{aligned}$$

If we can show $I \le \tau_1 \|\mathbf{\Theta} - \mathbf{\Theta}^*\|_2^2$ and $II \le \tau_2 \|\mathbf{\Theta} - \mathbf{\Theta}^*\|_2^2$, then we have $\tau = \sqrt{\tau_1 + \tau_2}$ since

$$\left\|\nabla_{\mathbf{\Theta}_k'} Q(\boldsymbol{\mu}_k^*, \mathbf{\Omega}_k^* | \mathbf{\Theta}^*) - \nabla_{\mathbf{\Theta}_k'} Q(\boldsymbol{\mu}_k^*, \mathbf{\Omega}_k^* | \mathbf{\Theta})\right\|_2 \le \sqrt{\tau_1 + \tau_2} \|\mathbf{\Theta} - \mathbf{\Theta}^*\|_2.$$

**Bounding I:** We apply Taylor expansion to simplify $D_L(\mathbf{\Theta}^*, \mathbf{\Theta})$. Remind that, by assumption, $\pi_k = 1/K$, and hence we have

$$L_{\mathbf{\Theta},k}(\boldsymbol{X}) = \frac{\pi_k f_k(\boldsymbol{X}; \mathbf{\Theta}_k)}{\sum_{k=1}^K \pi_k f_k(\boldsymbol{X}; \mathbf{\Theta}_k)} = \frac{|\mathbf{\Omega}_k|^{1/2} \exp\left\{-\frac{1}{2}(\boldsymbol{X} - \boldsymbol{\mu}_k)^\top \mathbf{\Omega}_k (\boldsymbol{X} - \boldsymbol{\mu}_k)\right\}}{\sum_{k=1}^K |\mathbf{\Omega}_k|^{1/2} \exp\left\{-\frac{1}{2}(\boldsymbol{X} - \boldsymbol{\mu}_k)^\top \mathbf{\Omega}_k (\boldsymbol{X} - \boldsymbol{\mu}_k)\right\}}.$$

Then, Taylor expansion of $L_{\mathbf{\Theta},k}(\boldsymbol{X})$ around $\mathbf{\Theta}_k^*$ leads to

$$L_{\mathbf{\Theta},k}(\boldsymbol{X}) = L_{\mathbf{\Theta}^*,k}(\boldsymbol{X}) + [\nabla_{\mathbf{\Theta}} L_{\mathbf{\Theta}_t,k}(\boldsymbol{X})]^\top (\mathbf{\Theta} - \mathbf{\Theta}^*), \tag{S.2}$$

where $\mathbf{\Theta}_t = \mathbf{\Theta}^* + t\Delta$ with $t \in [0,1]$ and $\Delta = \mathbf{\Theta} - \mathbf{\Theta}^*$. Here the derivative of $L_{\mathbf{\Theta},k}(\boldsymbol{X})$ with respect to $\mathbf{\Theta} = (\mathbf{\Theta}_1, \ldots, \mathbf{\Theta}_K)$ can be written as

$$\nabla_{\mathbf{\Theta}} L_{\mathbf{\Theta},k}(\boldsymbol{X}) = \left([\nabla_{\mathbf{\Theta}_1} L_{\mathbf{\Theta},k}(\boldsymbol{X})]^\top, \ldots, [\nabla_{\mathbf{\Theta}_K} L_{\mathbf{\Theta},k}(\boldsymbol{X})]^\top\right)^\top, \tag{S.3}$$

where

$$\nabla_{\mathbf{\Theta}_j} L_{\mathbf{\Theta},k}(\boldsymbol{X}) = \begin{cases} -L_{\mathbf{\Theta},k}(\boldsymbol{X}) \cdot L_{\mathbf{\Theta},j}(\boldsymbol{X}) \cdot \delta_{\mathbf{\Theta}_j}(\boldsymbol{X}) \text{ when } j \neq k; \\ L_{\mathbf{\Theta},k}(\boldsymbol{X})[1 - L_{\mathbf{\Theta},k}(\boldsymbol{X})] \cdot \delta_{\mathbf{\Theta}_k}(\boldsymbol{X}) \text{ when } j = k, \end{cases}$$

and, for $j = 1 \ldots, K$, and $\mathbf{\Theta}_j = \text{vec}(\mu_j, \mathbf{\Omega}_j)$,

$$\delta_{\mathbf{\Theta}_j}(\boldsymbol{X}) = \begin{pmatrix} \mathbf{\Omega}_j (\boldsymbol{X} - \boldsymbol{\mu}_j) \\ \frac{1}{2} \text{vec}\left\{\mathbf{\Omega}_j^{-1} - (\boldsymbol{X} - \boldsymbol{\mu}_j)(\boldsymbol{X} - \boldsymbol{\mu}_j)^\top\right\} \end{pmatrix}.$$

Next we apply this Taylor expansion to bound $I$. According to ($S.2$), we have

$$\begin{aligned} I &= \left\|\mathbb{E}\left[\mathbf{\Omega}_k^*(\boldsymbol{X} - \boldsymbol{\mu}_k^*)[\nabla_{\mathbf{\Theta}} L_{\mathbf{\Theta}_t,k}(\boldsymbol{X})]^\top (\mathbf{\Theta} - \mathbf{\Theta}^*)\right]\right\|_2^2 \\ &= \left\|\mathbb{E}\left[\mathbf{\Omega}_k^*(\boldsymbol{X} - \boldsymbol{\mu}_k^*)[\nabla_{\mathbf{\Theta}} L_{\mathbf{\Theta}_t,k}(\boldsymbol{X})]^\top\right]\right\|_2^2 \cdot \|\mathbf{\Theta} - \mathbf{\Theta}^*\|_2^2 \\ &\le \underbrace{\sup_{t \in [0,1]} \mathbb{E}\left[\|\mathbf{\Omega}_k^*(\boldsymbol{X} - \boldsymbol{\mu}_k^*)\|_2^2 \cdot \|\nabla_{\mathbf{\Theta}} L_{\mathbf{\Theta}_t,k}(\boldsymbol{X})\|_2^2\right]}_{\tau_1} \cdot \|\mathbf{\Theta} - \mathbf{\Theta}^*\|_2^2. \end{aligned}$$

By the definition of $\nabla_{\mathbf{\Theta}} L_{\mathbf{\Theta}_t,k}(\boldsymbol{X})$, which equals to (S.3) with $\mathbf{\Theta} = \mathbf{\Theta}_t$, we have

$$\begin{aligned}\|\nabla_{\mathbf{\Theta}} L_{\mathbf{\Theta}_t,k}(\boldsymbol{X})\|_2^2 &= \underbrace{\sum_{j\neq k}[L_{\mathbf{\Theta}_t,k}(\boldsymbol{X})L_{\mathbf{\Theta}_t,j}(\boldsymbol{X})]^2 \cdot [\delta_{\mathbf{\Theta}_{tj}}(\boldsymbol{X})]^\top \delta_{\mathbf{\Theta}_{tj}}(\boldsymbol{X})}_{A_1} \\ &\quad + \underbrace{\left[L_{\mathbf{\Theta}_t,k}(\boldsymbol{X})\big(1 - L_{\mathbf{\Theta}_t,k}(\boldsymbol{X})\big)\right]^2 \cdot [\delta_{\mathbf{\Theta}_{tk}}(\boldsymbol{X})]^\top \delta_{\mathbf{\Theta}_{tk}}(\boldsymbol{X})}_{A_2}.\end{aligned}$$

For each $j = 1, \ldots, K$, we define

$$W_j := \sup_{t\in[0,1]} \mathbb{E}\left\{[\delta_{\mathbf{\Theta}_{tj}}(\boldsymbol{X})]^\top \delta_{\mathbf{\Theta}_{tj}}(\boldsymbol{X}) \cdot \|\mathbf{\Omega}_k^*(\boldsymbol{X} - \boldsymbol{\mu}_k^*)\|_2^2\right\}, \tag{S.4}$$

Then

$$\tau_1 \leq \sup_{t\in[0,1]} \mathbb{E}\left[\|\mathbf{\Omega}_k^*(\boldsymbol{X} - \boldsymbol{\mu}_k^*)\|_2^2 (A_1 + A_2)\right]. \tag{S.5}$$

Under Condition 3.2, it is sufficient to get an upper bound for $\tau_1$,

$$\begin{aligned}\tau_1 &\leq \sup_{t\in[0,1]} \mathbb{E}\left[\|\mathbf{\Omega}_k^*(\boldsymbol{X} - \boldsymbol{\mu}_k^*)\|_2^2 A_1\right] + \sup_{t\in[0,1]} \mathbb{E}\left[\|\mathbf{\Omega}_k^*(\boldsymbol{X} - \boldsymbol{\mu}_k^*)\|_2^2 A_2\right] \\ &\leq \sum_{j\neq k} \frac{\gamma^2}{24^2(K-1)^2 M_j} \cdot W_j + \left(\frac{\gamma}{24(K-1)\sqrt{M_k}}(K-1)\right)^2 \cdot W_k.\end{aligned}$$

It implies that

$$\tau_1 \leq \frac{\gamma^2}{288}. \tag{S.6}$$

**Bounding II:** We can apply similar trick above to bound II. By triangle inequality, we have

$$\begin{aligned}II &\leq \underbrace{\left\|\frac{1}{2}\mathbb{E}\big[D_L(\mathbf{\Theta}^*, \mathbf{\Theta})\mathbf{\Omega}_k^{*-1}\big]\right\|_F^2}_{II_1} \\ &\quad + \underbrace{\left\|\frac{1}{2}\mathbb{E}\big[D_L(\mathbf{\Theta}^*, \mathbf{\Theta})(\boldsymbol{X} - \boldsymbol{\mu}_k^*)(\boldsymbol{X} - \boldsymbol{\mu}_k^*)^\top\big]\right\|_F^2}_{II_2}.\end{aligned}$$

Apply Taylor expansion in (*S*.2), we obtain

$$\begin{aligned}II_1 &\leq \underbrace{\frac{1}{2}\mathbb{E}\left[\|\nabla_{\mathbf{\Theta}} L_{\mathbf{\Theta}_t,k}(\boldsymbol{X})\|_2^2 \|\mathbf{\Omega}_k^{*-1}\|_F^2\right]}_{\gamma_{21}} \cdot \|\mathbf{\Theta} - \mathbf{\Theta}^*\|_2^2 \\ II_2 &\leq \underbrace{\frac{1}{2}\mathbb{E}\left[\|\nabla_{\mathbf{\Theta}} L_{\mathbf{\Theta}_t,k}(\boldsymbol{X})\|_2^2 \left\|(\boldsymbol{X} - \boldsymbol{\mu}_k^*)(\boldsymbol{X} - \boldsymbol{\mu}_k^*)^\top\right\|_F^2\right]}_{\gamma_{22}} \cdot \|\mathbf{\Theta} - \mathbf{\Theta}^*\|_2^2.\end{aligned}$$

Analogously to (S.4), we define

$$W_j^{'} := \sup_{t\in[0,1]} \mathbb{E}\left\{[\delta_{\mathbf{\Theta}_{tj}}(\boldsymbol{X})]^\top \delta_{\mathbf{\Theta}_{tj}}(\boldsymbol{X}) \left\|\mathbf{\Omega}_k^{*-1}\right\|_F^2\right\}, \tag{S.7}$$

$$W_j^{''} := \sup_{t\in[0,1]} \mathbb{E}\left\{[\delta_{\mathbf{\Theta}_{tj}}(\boldsymbol{X})]^\top \delta_{\mathbf{\Theta}_{tj}}(\boldsymbol{X}) \left\|(\boldsymbol{X}-\boldsymbol{\mu}_k^*)(\boldsymbol{X}-\boldsymbol{\mu}_k^*)^\top\right\|_F^2\right\}. \tag{S.8}$$

for each $j = 1, \ldots, K$. Under Condition 3.2, we have that,

$$\tau_{21} < \frac{\gamma^2}{576}, \quad \tau_{22} < \frac{\gamma^2}{576}, \text{ and hence } \tau_2 < \frac{\gamma^2}{288}.$$

This together with (S.6) implies that $\tau = \sqrt{\tau_1 + \tau_2} < \gamma/12$, namely

$$\left\|\nabla_{\mathbf{\Theta}_k^{'}} Q(\boldsymbol{\mu}_k^*, \mathbf{\Omega}_k^*|\mathbf{\Theta}^*) - \nabla_{\mathbf{\Theta}_k^{'}} Q(\boldsymbol{\mu}_k^*, \mathbf{\Omega}_k^*|\mathbf{\Theta})\right\|_2 \leq \frac{\gamma}{12}.$$

Now we take the summation

$$\sum_{k=1}^{K} \left\|\nabla_{\mathbf{\Theta}_k^{'}} Q(\boldsymbol{\mu}_k^*, \mathbf{\Omega}_k^*|\mathbf{\Theta}^*) - \nabla_{\mathbf{\Theta}_k^{'}} Q(\boldsymbol{\mu}_k^*, \mathbf{\Omega}_k^*|\mathbf{\Theta})\right\|_2^2 \leq \frac{\gamma}{12}\|\mathbf{\Theta} - \mathbf{\Theta}^*\|_2, \tag{S.9}$$

for any $\mathbf{\Theta} \in \mathbb{B}_\alpha(\mathbf{\Theta}^*)$. This ends the proof of Lemma 3.3. ■

## S.III Proof of Lemma 3.5

In order to compute $\gamma$, we consider each $\mathbf{\Theta}_k = \{\mu_k, \mathbf{\Omega}_k\}$ individually. That means we prove the following part first:

$$Q_n(\mathbf{\Theta}_k'|\mathbf{\Theta}) - Q_n(\mathbf{\Theta}_k^*|\mathbf{\Theta}) - \left\langle \nabla Q_n(\mathbf{\Theta}_k^*|\mathbf{\Theta}), \mathbf{\Theta}_k' - \mathbf{\Theta}_k^* \right\rangle \leq -\frac{\gamma}{2}\left\|\mathbf{\Theta}_k' - \mathbf{\Theta}_k^*\right\|_2^2,$$

where $Q_n(\mathbf{\Theta}_k|\mathbf{\Theta})$ means we set $\mathbf{\Theta}_i$ $i \neq k$ to zero.

It is sufficient to compute $\gamma_k$ in (3.9). Remind that $\mathbf{\Theta}_k^{'} = \text{vec}(\boldsymbol{\mu}_k, \mathbf{\Omega}_k) \in \mathbb{R}^{p^2+p}$. Therefore,

$$\nabla_{\mathbf{\Theta}_k^{'}} Q_n(\mathbf{\Theta}_k^{'}|\mathbf{\Theta}) = ([\nabla_{\boldsymbol{\mu}_k^{'}} Q_n(\mathbf{\Theta}_k^{'}|\mathbf{\Theta})]^\top, [\text{vec}(\nabla_{\mathbf{\Omega}_k^{'}} Q_n(\mathbf{\Theta}_k^{'}|\mathbf{\Theta}))]^\top)^\top, \tag{S.10}$$

with

$$\begin{aligned}
\nabla_{\boldsymbol{\mu}_k^{'}} Q_n(\mathbf{\Theta}_k^{'}|\mathbf{\Theta}) &= \frac{1}{n}\sum_{i=1}^{n}\left[L_{\mathbf{\Theta},k}(\boldsymbol{x}_i)\mathbf{\Omega}_k^{'}(\boldsymbol{x}_i - \boldsymbol{\mu}_k^{'})\right] \\
\nabla_{\mathbf{\Omega}_k^{'}} Q_n(\mathbf{\Theta}_k^{'}|\mathbf{\Theta}) &= \frac{1}{2n}\sum_{i=1}^{n}[L_{\mathbf{\Theta},k}(\boldsymbol{x}_i)]\mathbf{\Omega}_k^{'-1} \\
&- \frac{1}{2n}\sum_{i=1}^{n}[L_{\mathbf{\Theta},k}(\boldsymbol{x}_i)(\boldsymbol{x}_i - \boldsymbol{\mu}_k^{'})(\boldsymbol{x}_i - \boldsymbol{\mu}_k^{'})^\top].
\end{aligned}$$

Denote $h(\boldsymbol{\mu}, \boldsymbol{\Omega}) := \frac{1}{2}(\boldsymbol{x}_i - \boldsymbol{\mu})^\top \boldsymbol{\Omega}(\boldsymbol{x}_i - \boldsymbol{\mu})$. According to the definition in (2.8), we have

$$\begin{aligned} Q_n(\boldsymbol{\Theta}'_k|\boldsymbol{\Theta}) - Q_n(\boldsymbol{\Theta}^*_k|\boldsymbol{\Theta}) \quad = \quad & \frac{1}{n}\sum_{i=1}^n \Big[L_{\boldsymbol{\Theta},k}(\boldsymbol{x}_i)\{\frac{1}{2}\log\det(\boldsymbol{\Omega}'_k) \\ & -\frac{1}{2}\log\det(\boldsymbol{\Omega}^*_k) + h(\boldsymbol{\mu}^*_k, \boldsymbol{\Omega}^*_k) - h(\boldsymbol{\mu}'_k, \boldsymbol{\Omega}'_k)\}\Big]. \end{aligned}$$

This together with ($S$.10) implies that

$$Q_n(\boldsymbol{\Theta}'_k|\boldsymbol{\Theta}) - Q_n(\boldsymbol{\Theta}^*_k|\boldsymbol{\Theta}) - \left\langle \nabla_{\boldsymbol{\Theta}'_k} Q_n(\boldsymbol{\Theta}^*_k|\boldsymbol{\Theta}), \boldsymbol{\Theta}'_k - \boldsymbol{\Theta}^*_k \right\rangle = I + II,$$

where

$$\begin{aligned} I &= \frac{1}{n}\sum_{i=1}^n \Big[L_{\boldsymbol{\Theta},k}(\boldsymbol{x}_i)\Big\{h(\boldsymbol{\mu}^*_k, \boldsymbol{\Omega}^*_k) - h(\boldsymbol{\mu}'_k, \boldsymbol{\Omega}^*_k)\Big\}\Big] \\ & -(\boldsymbol{\mu}'_k - \boldsymbol{\mu}^*_k)^\top \nabla_{\boldsymbol{\mu}'_k} Q_n(\boldsymbol{\Theta}^*_k|\boldsymbol{\Theta}^{(t)}), \\ II &= \frac{1}{n}\sum_{i=1}^n \Big[L_{\boldsymbol{\Theta},k}(\boldsymbol{x}_i)\Big\{\frac{1}{2}\log\det(\boldsymbol{\Omega}'_k) - \frac{1}{2}\log\det(\boldsymbol{\Omega}^*_k) \\ & + h(\boldsymbol{\mu}'_k, \boldsymbol{\Omega}^*_k) - h(\boldsymbol{\mu}'_k, \boldsymbol{\Omega}'_k)\Big\}\Big] - [\mathrm{vec}(\boldsymbol{\Omega}'_k - \boldsymbol{\Omega}^*_k)]^\top \nabla_{\boldsymbol{\Omega}'_k} Q_n(\boldsymbol{\Theta}^*_k|\boldsymbol{\Theta}^{(t)}). \end{aligned}$$

By a little algebra, we can show that

$$I = -\frac{1}{2n}\sum_{i=1}^n L_{\boldsymbol{\Theta},k}(\boldsymbol{x}_i)(\boldsymbol{\mu}'_k - \boldsymbol{\mu}^*_k)^\top \boldsymbol{\Omega}^*_k(\boldsymbol{\mu}'_k - \boldsymbol{\mu}^*_k).$$

Due to the positive definiteness of $\boldsymbol{\Omega}^*_k$ , it is shown the following inequality

$$(\boldsymbol{\mu}'_k - \boldsymbol{\mu}^*_k)^\top (\boldsymbol{\Omega}^*_k - \sigma_{\min}(\boldsymbol{\Omega}^*_k)I_p)(\boldsymbol{\mu}'_k - \boldsymbol{\mu}^*_k) \geq 0$$

$$(\boldsymbol{\mu}'_k - \boldsymbol{\mu}^*_k)^\top \boldsymbol{\Omega}^*_k(\boldsymbol{\mu}'_k - \boldsymbol{\mu}^*_k) \geq (\boldsymbol{\mu}'_k - \boldsymbol{\mu}^*_k)^\top \sigma_{\min}(\boldsymbol{\Omega}^*_k)I_p(\boldsymbol{\mu}'_k - \boldsymbol{\mu}^*_k) \geq \beta_1 \left\|\boldsymbol{\mu}'_k - \boldsymbol{\mu}^*_k\right\|_2^2.$$

Substituting the above bound, it is shown that

$$I \leq -\frac{\beta_1}{2n}\sum_{i=1}^n L_{\boldsymbol{\Theta},k}(\boldsymbol{x}_i)\|\boldsymbol{\mu}'_k - \boldsymbol{\mu}^*_k\|_2^2. \tag{S.11}$$

Therefore, it remains to show that

$$II \leq -\frac{1}{2n}\sum_{i=1}^n \frac{L_{\boldsymbol{\Theta},k}(\boldsymbol{x}_i)}{2(\beta_2 + 2\alpha)^2}\|\mathrm{vec}(\boldsymbol{\Omega}'_k - \boldsymbol{\Omega}^*_k)\|_2^2. \tag{S.12}$$

Note that, in order to show ($S$.12), it is equivalent to deriving the strong concavity parameter of $g(\boldsymbol{\Omega}_k)$, where

$$g(\boldsymbol{\Omega}_k) := \frac{1}{n}\sum_{i=1}^n \left[L_{\boldsymbol{\Theta},k}(\boldsymbol{x}_i)\left\{\frac{1}{2}\log\det(\boldsymbol{\Omega}_k) - h(\boldsymbol{\mu}'_k, \boldsymbol{\Omega}_k)\right\}\right].$$

To see it, finding the strong concavity parameter of $g(\mathbf{\Omega}_k)$ aims to compute $\rho_k$ such that, for any $\mathbf{\Omega}_k', \mathbf{\Omega}_k^* \in \mathcal{B}_\alpha(\mathbf{\Omega}_k^*)$,

$$g(\mathbf{\Omega}_k') - g(\mathbf{\Omega}_k^*) - \left\langle \mathrm{vec}\left(\nabla g(\mathbf{\Omega}_k^*)\right), \mathrm{vec}(\mathbf{\Omega}_k' - \mathbf{\Omega}_k^*)\right\rangle \leq -\rho_k/2 \cdot \|\mathbf{\Omega}_k' - \mathbf{\Omega}_k^*\|_F^2,$$

where the left hand side is exactly $II$. According to Taylor expansion, we can expand $g(\mathbf{\Omega}_k')$ around $\mathbf{\Omega}_k^*$ and obtain

$$\begin{aligned} g(\mathbf{\Omega}_k') &= g(\mathbf{\Omega}_k^*) + \left\langle \mathrm{vec}(\nabla g(\mathbf{\Omega}_k^*), \mathrm{vec}(\mathbf{\Omega}_k' - \mathbf{\Omega}_k^*)\right\rangle \\ &\quad + \frac{1}{2}\left[\mathrm{vec}(\mathbf{\Omega}_k' - \mathbf{\Omega}_k^*)\right]^\top \nabla^2 g(\mathbf{Z})\left[\mathrm{vec}(\mathbf{\Omega}_k' - \mathbf{\Omega}_k^*)\right], \end{aligned}$$

where $\mathbf{Z} = t\mathbf{\Omega}_k' + (1-t)\mathbf{\Omega}_k^*$ with $t \in [0,1]$. For any two matrices $\mathbf{A}, \mathbf{B}$, we write $\mathbf{A} \succeq \mathbf{B}$ if $\mathbf{A} - \mathbf{B}$ is positive semi-definite. We denote $\mathbb{1}_p$ as the identity matrix with dimension $p \times p$. And $\sigma_i(A)$ is the $i$-th eigenvalue of matrix A. Therefore, if we can show that $-\nabla^2 g(\mathbf{Z}) \succeq m\,\mathbb{1}_p$, i.e., the minimal eigenvalue value $\sigma_{\min}(-\nabla^2 g(\mathbf{Z})) \geq m$, for some positive $m \in \mathbb{R}$, then we have the strongly concavity parameter $\rho_k = m$. By the definition, we have $\nabla^2 g(\mathbf{\Omega}_k^*) = -\frac{1}{2n}\sum_{i=1}^n L_{\mathbf{\Theta},k}(\boldsymbol{x}_i)[\mathbf{\Omega}_k^*]^{-1} \otimes [\mathbf{\Omega}_k^*]^{-1}$. Denote $\widetilde{\Delta} = \mathbf{\Omega}_k' - \mathbf{\Omega}_k^*$. We obtain

$$-\nabla^2 g(\mathbf{Z}) = \frac{1}{2n}\sum_{i=1}^n L_{\mathbf{\Theta},k}(\boldsymbol{x}_i)\left(\mathbf{\Omega}_k^* + t\widetilde{\Delta}\right)^{-1} \otimes \left(\mathbf{\Omega}_k^* + t\widetilde{\Delta}\right)^{-1}.$$

According to Theorem 4.2.1 2 in Horn and Johnson (1988), for any two matrices $\mathbf{A}, \mathbf{B}$, the minimal eigenvalue value of $\mathbf{A} \otimes \mathbf{B}$ equals the products of the minimal eigenvalue values of $\mathbf{A}$ and $\mathbf{B}$. Therefore, we have $\sigma_{\min}\left(\mathbf{A}^{-1} \otimes \mathbf{A}^{-1}\right) = \left[\sigma_{\min}(\mathbf{A}^{-1})\right]^2 = [\sigma_{\max}(\mathbf{A})]^{-2} = \|\mathbf{A}\|_2^{-2}$, where $\|\mathbf{A}\|_2$ refers to the spectral norm of matrix $\mathbf{A}$. Hence,

$$\begin{aligned} \sigma_{\min}(-\nabla^2 g(\mathbf{Z})) &= \frac{1}{2n}\sum_{i=1}^n L_{\mathbf{\Theta},k}(\boldsymbol{x}_i)\|\mathbf{\Omega}_k^* + t\widetilde{\Delta}\|_2^{-2} \\ &\geq \frac{1}{2n}\sum_{i=1}^n L_{\mathbf{\Theta},k}(\boldsymbol{x}_i)\left[\|\mathbf{\Omega}_k^*\|_2 + \|t\widetilde{\Delta}\|_2\right]^{-2}. \end{aligned}$$

As $\|\mathbf{\Theta}' - \mathbf{\Theta}^*\| \leq 2\alpha$, $\|\mathbf{\Omega}_k' - \mathbf{\Omega}_k^*\|_2 \leq \|\mathbf{\Theta}' - \mathbf{\Theta}^*\|_2 \leq 2\alpha$. Therefore,

$$\begin{aligned} \sigma_{\min}(-\nabla^2 g(\mathbf{Z})) &\geq \frac{1}{2n}\sum_{i=1}^n L_{\mathbf{\Theta},k}(\boldsymbol{x}_i)\left[\|\mathbf{\Omega}_k^*\|_2 + 2\alpha\right]^{-2} \\ &\geq \frac{1}{2n}\sum_{i=1}^n L_{\mathbf{\Theta},k}(\boldsymbol{x}_i)\left(\beta_2 + 2\alpha\right)^{-2}, \end{aligned}$$

which implies ($S$.12). Putting the upper bound of $I$ and $II$ together,

$$I + II \leq -\underbrace{\frac{1}{2n}\sum_{i=1}^n L_{\mathbf{\Theta},k}(\boldsymbol{x}_i)}_{(a)} \cdot \min\left\{\beta_1, \frac{1}{2(\beta_2 + 2\alpha)^2}\right\}\|\mathbf{\Theta}_k' - \mathbf{\Theta}_k^*\|_2^2. \tag{S.13}$$

However, $(a)$ is a random term but we require a non-random strong concavity parameter. Thus a concentration bound will be applied on it. $\{L_{\boldsymbol{\Theta},k}(\boldsymbol{x}_i), i = 1, \ldots, n\}$ are independent random variables with $0 \leq L_{\boldsymbol{\Theta},k}(\boldsymbol{x}_i) \leq 1$. After applying a basic Hoeffding's inequality, we have

$$\mathbb{P}\left(\left|\frac{1}{n}\sum_{i=1}^{n} L_{\boldsymbol{\Theta},k}(\boldsymbol{x}_i) - \mathbb{E}[L_{\boldsymbol{\Theta},k}(\boldsymbol{X})]\right| \leq t\right) \geq 1 - 2e^{-2nt^2},$$

which implies

$$\left|\frac{1}{n}\sum_{i=1}^{n} L_{\boldsymbol{\Theta},k}(\boldsymbol{x}_i) - \mathbb{E}[L_{\boldsymbol{\Theta},k}(\boldsymbol{X})]\right| \leq \sqrt{\frac{1}{2}\log\frac{2K}{\delta}}\sqrt{\frac{1}{n}},$$

with probability at least $1 - \delta/K$. As $\sqrt{\log(2K/\delta)/2n} = o(1)$, there exists some constant $c$ such that

$$\sqrt{\frac{\log 2K}{2\delta n}} - \mathbb{E}[L_{\boldsymbol{\Theta},k}(\boldsymbol{X})] \leq -c,$$

when $n$ is large enough. Then plugging it into (S.13),

$$I + II \leq -\frac{1}{2}c \cdot \min\left\{\beta_1, \frac{1}{2(\beta_2 + 2\alpha)^2}\right\} \|\boldsymbol{\Theta}'_k - \boldsymbol{\Theta}^*_k\|_2^2,$$

with probability at least $1 - \delta/K$, where

$$\gamma = c \min\left\{\beta_1, \frac{1}{2(\beta_2 + 2\alpha)^2}\right\}.$$

Once the individual strong concavity parameter is computed, we can simply take the summation from 1 to $K$:

$$\sum_{k=1}^{K} Q_n(\boldsymbol{\Theta}'_k|\boldsymbol{\Theta}) - Q_n(\boldsymbol{\Theta}^*_k|\boldsymbol{\Theta}) - \langle \nabla Q_n(\boldsymbol{\Theta}^*_k|\boldsymbol{\Theta}), \boldsymbol{\Theta}'_k - \boldsymbol{\Theta}^*_k \rangle \leq -\frac{1}{2}\sum_{k=1}^{K} \gamma \left\|\boldsymbol{\Theta}'_k - \boldsymbol{\Theta}^*_k\right\|_2^2$$

which implies

$$Q_n(\boldsymbol{\Theta}'|\boldsymbol{\Theta}) - Q_n(\boldsymbol{\Theta}^*|\boldsymbol{\Theta}) - \langle \nabla Q_n(\boldsymbol{\Theta}^*|\boldsymbol{\Theta}), \boldsymbol{\Theta}' - \boldsymbol{\Theta}^* \rangle \leq -\frac{1}{2}\gamma \left\|\boldsymbol{\Theta}' - \boldsymbol{\Theta}^*\right\|_2^2$$

with probability at least $1 - \delta$. This ends the proof of Lemma 3.5. ■

## S.IV A Key Lemma for Proving Corollary 3.14

The next lemma computes the statistical errors in Condition 3.6 for our SCAN penalty and provides explicit forms of the corresponding $\varepsilon_1, \varepsilon_2$ and $\delta_1, \delta_2$.

**Lemma S.1.** Suppose that Condition 3.12, 3.13 hold, then Condition 3.6 is satisfied for SCAN penalty with

$$\varepsilon_1 = (CK\|\mathbf{\Omega}^*\|_\infty + C'K^{1.5})\sqrt{\frac{\log p + \log(e/\delta)}{n}}, \delta_1 = (18K^2 + 6K)\delta, \tag{S.14}$$

$$\varepsilon_2 = C^{''}\sqrt{p}\sqrt{\frac{K^3\left(\log p + \log(e/\delta)\right)}{n}}, \delta_2 = (8K^2 + 2K)\delta, \tag{S.15}$$

for some absolute constant $C, C', C^{''} > 0$. Here $\|\mathbf{\Omega}^*\|_\infty$ is the overall max induced norm defined as $\|\mathbf{\Omega}^*\|_\infty = \max_{k\in[K]} \|\mathbf{\Omega}_k^*\|_\infty$.

In Lemma S.1, the number of clusters $K$ is allowed to grow with the sample size $n$ and the dimension $p$. The diverging rate of $K$ controls the convergence probability at each iteration and is upper bounded to ensure that the statistical errors hold with a high probability tending to 1 with a proper choice of $\delta$, e.g., $\delta = 1/p$.

**Proof of Lemma S.1**: For the first part of this proof, we focus on the upper bound of $\|\nabla Q_n(\mathbf{\Theta}^*|\mathbf{\Theta}) - \nabla Q(\mathbf{\Theta}^*|\mathbf{\Theta})\|_{\mathcal{P}^*}$. Recall that

$$\begin{aligned}
\nabla Q_n(\mathbf{\Theta}^*|\mathbf{\Theta}) - \nabla Q(\mathbf{\Theta}^*|\mathbf{\Theta}) &= \begin{pmatrix} \nabla_{\mathbf{\Theta}_1^*} Q_n(\mathbf{\Theta}^*|\mathbf{\Theta}) - \nabla_{\mathbf{\Theta}_1^*} Q(\mathbf{\Theta}^*|\mathbf{\Theta}) \\ \vdots \\ \nabla_{\mathbf{\Theta}_K^*} Q_n(\mathbf{\Theta}^*|\mathbf{\Theta}) - \nabla_{\mathbf{\Theta}_K^*} Q(\mathbf{\Theta}^*|\mathbf{\Theta}) \end{pmatrix} \\
&= \begin{pmatrix} \nabla_{\boldsymbol{\mu}_1^*} Q_n(\mathbf{\Theta}^*|\mathbf{\Theta}) - \nabla_{\boldsymbol{\mu}_1^*} Q(\mathbf{\Theta}^*|\mathbf{\Theta}) \\ \text{vec}\left\{\nabla_{\mathbf{\Omega}_1^*} Q_n(\mathbf{\Theta}^*|\mathbf{\Theta}) - \nabla_{\mathbf{\Omega}_1^*} Q(\mathbf{\Theta}^*|\mathbf{\Theta})\right\}^\top \\ \vdots \\ \nabla_{\boldsymbol{\mu}_K^*} Q_n(\mathbf{\Theta}^*|\mathbf{\Theta}) - \nabla_{\boldsymbol{\mu}_K^*} Q(\mathbf{\Theta}^*|\mathbf{\Theta}) \\ \text{vec}\left\{\nabla_{\mathbf{\Omega}_K^*} Q_n(\mathbf{\Theta}^*|\mathbf{\Theta}) - \nabla_{\mathbf{\Omega}_K^*} Q(\mathbf{\Theta}^*|\mathbf{\Theta})\right\}^\top \end{pmatrix}.
\end{aligned} \tag{S.16}$$

For simplicity, we define $h_{\mu_k}(\mathbf{\Theta}^*) = \nabla_{\boldsymbol{\mu}_k^*} Q_n(\mathbf{\Theta}^*|\mathbf{\Theta}) - \nabla_{\boldsymbol{\mu}_k^*} Q(\mathbf{\Theta}^*|\mathbf{\Theta})$ and $h_{\mathbf{\Omega}_k^*}(\mathbf{\Theta}^*) = \nabla_{\mathbf{\Omega}_k^*} Q_n(\mathbf{\Theta}^*|\mathbf{\Theta}) - \nabla_{\mathbf{\Omega}_k^*} Q(\mathbf{\Theta}^*|\mathbf{\Theta})$. Then from the definition of dual norm $\mathcal{P}^*$ (A.1), we can have

$$\begin{aligned}
&\|\nabla Q_n(\mathbf{\Theta}^*|\mathbf{\Theta}) - \nabla Q(\mathbf{\Theta}^*|\mathbf{\Theta})\|_{\mathcal{P}^*} \le M_1 \max_{k\in[K]} \underbrace{\|h_{\mu_k}(\mathbf{\Theta}^*)\|_\infty}_{I} \\
&+ M_2 \max_{k\in[K]} \underbrace{\left\|h_{\mathbf{\Omega}_k^*}(\mathbf{\Theta}^*)\right\|_{\max}}_{II} + M_3 \max_{i,j} \underbrace{\left\|\left[h_{\mathbf{\Omega}_k^*}(\mathbf{\Theta}^*)\right]_{ij}, \ldots, \left[h_{\mathbf{\Omega}_k^*}(\mathbf{\Theta}^*)\right]_{ij}\right\|_2}_{III},
\end{aligned}$$

which are corresponding to the penalty on element-wise cluster means, element-wise precision matrices and group structures of multiple precision matrices, respectively.

**Bounding Statistical Error for $k$-th Cluster Mean:** Referring to the proof in Lemma 3.1,

$$h_{\mu_k^*}(\mathbf{\Theta}^*) \quad = \quad \frac{1}{n}\sum_{i=1}^n L_{\mathbf{\Theta},k}(\boldsymbol{x}_i)\mathbf{\Omega}_k^*(\boldsymbol{x}_i - \boldsymbol{\mu}_k^*) - \mathbb{E}\left[L_{\mathbf{\Theta},k}(\boldsymbol{X})\mathbf{\Omega}_k^*(\boldsymbol{X} - \boldsymbol{\mu}_k^*)\right].$$

Note that $\|\boldsymbol{\Omega}_k^*\|_\infty$ is a scalar. By using triangle inequality, we simplify $I$ by two parts:

$$\begin{aligned} I &\leq \|\boldsymbol{\Omega}_k^*\|_\infty \left\| \frac{1}{n}\sum_{i=1}^n L_{\boldsymbol{\Theta},k}(\boldsymbol{x}_i)(\boldsymbol{x}_i - \boldsymbol{\mu}_k^*) - \mathbb{E}\left[L_{\boldsymbol{\Theta},k}(\boldsymbol{X})(\boldsymbol{X} - \boldsymbol{\mu}_k^*)\right] \right\|_\infty \\ &\leq \|\boldsymbol{\Omega}_k^*\|_\infty \underbrace{\left\| \frac{1}{n}\sum_{i=1}^n L_{\boldsymbol{\Theta},k}(\boldsymbol{x}_i)\boldsymbol{x}_i - \mathbb{E}\left[L_{\boldsymbol{\Theta},k}(\boldsymbol{X})\boldsymbol{X}\right] \right\|_\infty}_{I_1} \\ &\quad + \|\boldsymbol{\Omega}_k^*\|_\infty \underbrace{\left\| \left( \frac{1}{n}\sum_{i=1}^n L_{\boldsymbol{\Theta},k}(\boldsymbol{x}_i) - \mathbb{E}\left[L_{\boldsymbol{\Theta},k}(\boldsymbol{X})\right] \right) \boldsymbol{\mu}_k^* \right\|_\infty}_{I_2}. \end{aligned}$$

*Bounding $I_1$:* Denote

$$\zeta = \frac{1}{n}\sum_{i=1}^n L_{\boldsymbol{\Theta},k}(\boldsymbol{x}_i)\boldsymbol{x}_i - \mathbb{E}\left[L_{\boldsymbol{\Theta},k}(\boldsymbol{X})\boldsymbol{X}\right]$$

For $\zeta \in \mathbb{R}^p$, we consider the $j$-th coordinate $\zeta_j$ of $\zeta$

$$\zeta_j = \frac{1}{n}\sum_{i=1}^n L_{\boldsymbol{\Theta},k}(\boldsymbol{x}_i)x_{ij} - \mathbb{E}\left[L_{\boldsymbol{\Theta},k}(\boldsymbol{X})X_j\right]. \tag{S.17}$$

We introduce a set of missing data $\{c_i, i = 1, \ldots, n\}$, which are independent copies of random variable $c$. The pair $(\boldsymbol{x}_i, c_i)$ are the independent copy of $(\boldsymbol{X}, c)$. Here $c$ takes a value from the set $\{1, \ldots, K\}$, where $c = k'$ indicates that $\boldsymbol{X}$ was generated by the $k'$-th mixture component. In another word, the conditional distribution of $\boldsymbol{X}$ is defined below:

$$\boldsymbol{X}|c = k' \sim \mathcal{N}(\boldsymbol{\mu}_{k'}^*, \boldsymbol{\Sigma}_{k'}^*)$$

$$\mathbb{P}(c = k') = \pi_{k'},\ \sum_{k'}^K \pi_{k'} = 1.$$

This is the usual choice of missing data in EM approaches to mixture modeling. The quantity $(\boldsymbol{x}_i, c_i)$ is referred to as the completed data. Now by the assumption, the $j$-th coordinate $x_{ij}$ of $\boldsymbol{x}_i$ can be rewritten as the form below:

$$x_{ij} = \sum_{k'=1}^K I\{c_i = k'\}(\mu_{k'j}^* + V_{k'j}),\ j \in [p] \tag{S.18}$$

where $\mu_{k'j}^*$ is the $j$-th coordinate of the true cluster mean $\boldsymbol{\mu}_{k'}^*$ and $V_{k'j} \sim \mathcal{N}(0, \boldsymbol{\Sigma}_{k'jj}^*)$. Plugging (S.18) into (S.17), it suffices to bound $\zeta_j$.

$$\begin{aligned} |\zeta_j| &\le \left| \frac{1}{n}\sum_{i=1}^n \sum_{k'=1}^K L_{\boldsymbol{\Theta},k}(\boldsymbol{x}_i) I\{c_i = k'\}\mu^*_{k'j} - \mathbb{E}\left[\sum_{k'=1}^K L_{\boldsymbol{\Theta},k}(\boldsymbol{X}) I\{c = k'\}\mu^*_{k'j}\right]\right| \\ &+ \left| \frac{1}{n}\sum_{i=1}^n \sum_{k'=1}^K L_{\boldsymbol{\Theta},k}(\boldsymbol{x}_i) I\{c_i = k'\}V^*_{k'j} - \mathbb{E}\left[\sum_{k'=1}^K L_{\boldsymbol{\Theta},k}(\boldsymbol{X}) I\{c = k'\}V^*_{k'j}\right]\right| \\ &\le \sum_{k'=1}^K \underbrace{\left| \frac{1}{n}\sum_{i=1}^n L_{\boldsymbol{\Theta},k}(\boldsymbol{x}_i) I\{c_i = k'\}\mu^*_{k'j} - \mathbb{E}\Big[L_{\boldsymbol{\Theta},k}(\boldsymbol{X}) I\{c = k'\}\mu^*_{k'j}\Big]\right|}_{\zeta_{j_1}} \\ &+ \sum_{k'=1}^K \underbrace{\left| \frac{1}{n}\sum_{i=1}^n L_{\boldsymbol{\Theta},k}(\boldsymbol{x}_i) I\{c_i = k'\}V^*_{k'j} - \mathbb{E}\Big[L_{\boldsymbol{\Theta},k}(\boldsymbol{X}) I\{c = k'\}V^*_{k'j}\Big]\right|}_{\zeta_{j_2}}. \end{aligned}$$

We bound $\zeta_{j_1}$ first. Based on the fact that $|L_{\boldsymbol{\Theta},k}(\boldsymbol{x}_i) I\{c_i = k'\}\mu^*_{k'j}| \le |\mu^*_{k'j}| \le \|\boldsymbol{\mu}^*_{k'}\|_\infty$ almost surely it can show that $L_{\boldsymbol{\Theta},k}(\boldsymbol{x}_i) I\{c_i = k'\}\mu^*_{k'j}$ is a sub-gaussian random variable with norm $\|\boldsymbol{\mu}^*_{k'}\|_\infty$. Following the Example 5.8 in Vershynin (2012), $\|L_{\boldsymbol{\Theta},k}(\boldsymbol{x}_i) I\{c_i = k'\}\mu^*_{k'j}\|_{\psi_2} \le \|\boldsymbol{\mu}^*_{k'}\|_\infty$ where $\|\cdot\|_{\psi_2}$ is defined as sub-Gaussian norm. According to supporting Lemma S.3

$$\Big\| L_{\boldsymbol{\Theta},k}(\boldsymbol{x}_i) I\{c_i = k'\}\mu^*_{k'j} - \mathbb{E}\left[L_{\boldsymbol{\Theta},k}(\boldsymbol{X}) I\{c = k'\}\mu^*_{k'j}\right] \Big\|_{\psi_2} \le 2\Big\|\boldsymbol{\mu}^*_{k'}\Big\|_\infty.$$

The standard concentration result in supporting Lemma S.4 yields that for every $t \ge 0$ and some constant $D_1$,

$$\mathbb{P}\left(|\zeta_{j_1}| \ge t\right) \le e\exp\left(-\frac{D_1 n t^2}{4\|\boldsymbol{\mu}^*_{k'}\|^2_\infty}\right),$$

which implies that, with probability at least $1-\delta$,

$$|\zeta_{j_1}| \le \sqrt{\frac{4}{D_1}}\|\boldsymbol{\mu}^*_{k'}\|_\infty \sqrt{\frac{\log(e/\delta)}{n}}. \tag{S.19}$$

Now we start to bound $\zeta_{j_2}$. The fact that $L_{\boldsymbol{\Theta},k}(\boldsymbol{x}_i) I\{c_i = k'\} \le 1$ shows that it is a sub-gaussian random variable with norm $\|L_{\boldsymbol{\Theta},k}(\boldsymbol{x}_i) I\{c_i = k'\}\|_{\psi_2} \le 1$. $V^*_{k'j}$ is a Gaussian random variable so that it is also a sub-gaussian random variable with norm $\|V^*_{k'j}\|_{\psi_2} \le (\|\boldsymbol{\Sigma}^*_{k'}\|_{\max})^{1/2}$. Then using the result in supporting Lemma S.2, $L_{\boldsymbol{\Theta},k}(\boldsymbol{x}_i) I\{c_i = k'\}V^*_{k'j}$ is sub-exponential random variable. Moreover, there exists constant $D_2$ such that

$$\Big\| L_{\boldsymbol{\Theta},k}(\boldsymbol{x}_i) I\{c_i = k'\}V^*_{k'j} \Big\|_{\psi_1} \le D_2\Big(\|\boldsymbol{\Sigma}^*_{k'}\|_{\max}\Big)^{1/2}.$$

Supporting lemma S.3 implies

$$\Big\| L_{\boldsymbol{\Theta},k}(\boldsymbol{x}_i) I\{c_i = k'\}V^*_{k'j} - \mathbb{E}\left[L_{\boldsymbol{\Theta},k}(\boldsymbol{X}) I\{c = k'\}V^*_{k'j}\right] \Big\|_{\psi_1} \le 2D_2\Big(\|\boldsymbol{\Sigma}^*_{k'}\|_{\max}\Big)^{1/2}.$$

Following the concentration inequality of sub-exponential random variables in supporting Lemma S.5, there exists some constant $D_3$ such that the following inequality

$$\mathbb{P}\Big(|\zeta_{j_2}| \geq t\Big) \leq 2\exp\left(-D_3 \min\left\{\frac{t^2}{4D_2^2\|\mathbf{\Sigma}_{k'}^*\|_{\max}}, \frac{t}{2D_2(\|\mathbf{\Sigma}_{k'}^*\|_{\max})^{1/2}}\right\} n\right),$$

holds every $t \geq 0$. For sufficient small $t$, it reduces to

$$\mathbb{P}\Big(|\zeta_{j_2}| \geq t\Big) \leq 2\exp\left(-D_3 \frac{nt^2}{4D_2\|\mathbf{\Sigma}_{k'}^*\|_{\max}}\right),$$

which implies that

$$|\zeta_{j_2}| \leq \sqrt{\frac{4D_2}{D_3}}(\|\mathbf{\Sigma}_{k'}^*\|_{\max})^{1/2}\sqrt{\frac{\log(2/\delta)}{n}}, \tag{S.20}$$

with probability at least $1-\delta$.

Adding (S.19) and (S.20) together, we have

$$\begin{aligned}
|\zeta_{j_1}| + |\zeta_{j_2}| &\leq \sqrt{\frac{4}{D_1}}\|\boldsymbol{\mu}_{k'}^*\|_\infty\sqrt{\frac{\log(e/\delta)}{n}} + \sqrt{\frac{4D_2}{D_3}}(\|\mathbf{\Sigma}_{k'}^*\|_{\max})^{1/2}\sqrt{\frac{\log(2/\delta)}{n}} \\
&\leq \sqrt{\frac{4}{D}}\left(\|\boldsymbol{\mu}_{k'}^*\|_\infty + (\|\mathbf{\Sigma}_{k'}^*\|_{\max})^{1/2}\right)\sqrt{\frac{\log(e/\delta)}{n}},
\end{aligned}$$

by taking $D = \min\{D_1, D_3/D_2\}$, with at least probability $1-2\delta$. Therefore, it's sufficient to bound $|\zeta_j|$ by

$$|\zeta_j| \leq \sqrt{\frac{4}{D}}\sum_{k'=1}^{K}\left(\|\boldsymbol{\mu}_{k'}^*\|_\infty + (\|\mathbf{\Sigma}_{k'}^*\|_{\max})^{1/2}\right)\sqrt{\frac{\log(e/\delta)}{n}},$$

with at least probability $1-2K\delta$. Taking the union bound over $p$ coordinates, we obtain

$$I_1 \leq \sqrt{\frac{4}{D}}\sum_{k'=1}^{K}\left(\|\boldsymbol{\mu}_{k'}^*\|_\infty + (\|\mathbf{\Sigma}_{k'}^*\|_{\max})^{1/2}\right)\sqrt{\frac{\log(e/\delta)+\log p}{n}}, \tag{S.21}$$

with at least probability $1-2K\delta$.

*Bounding $I_2$:* Recall that

$$I_2 = \left\|\left(\frac{1}{n}\sum_{i=1}^{n} L_{\mathbf{\Theta},k}(\boldsymbol{x}_i) - \mathbb{E}\left[L_{\mathbf{\Theta},k}(\boldsymbol{X})\right]\right)\boldsymbol{\mu}_k^*\right\|_\infty \leq \left|\frac{1}{n}\sum_{i=1}^{n} L_{\mathbf{\Theta},k}(\boldsymbol{x}_i) - \mathbb{E}\left[L_{\mathbf{\Theta},k}(\boldsymbol{X})\right]\right| \|\boldsymbol{\mu}_k^*\|_\infty.$$

$\{L_{\mathbf{\Theta},k}(\boldsymbol{x}_i)|i=1,\ldots n\}$ are bounded independent random variables within interval between 0 and 1. Then it follows Hoeffding's inequality in supporting Lemma S.6 that

$$\mathbb{P}\left(\left|\frac{1}{n}\sum_{i=1}^{n} L_{\mathbf{\Theta},k}(\boldsymbol{x}_i) - \mathbb{E}[L_{\mathbf{\Theta},k}(\boldsymbol{X})]\right| \leq t\right) \geq 1 - 2e^{-2nt^2},$$

which implies

$$\left|\frac{1}{n}\sum_{i=1}^{n} L_{\mathbf{\Theta},k}(\boldsymbol{x}_i) - \mathbb{E}[L_{\mathbf{\Theta},k}(\boldsymbol{X})]\right| \le \sqrt{\frac{1}{2}\log\frac{2}{\delta}}\cdot\sqrt{\frac{1}{n}}, \tag{S.22}$$

with probability at least $1-\delta$. Combining with the reminder term $\|\boldsymbol{\mu}_k^*\|$,

$$I_2 \le \sqrt{\frac{1}{2}\log\frac{2}{\delta}}\cdot\sqrt{\frac{1}{n}}\|\boldsymbol{\mu}_k^*\|_\infty. \tag{S.23}$$

Note that the bound in (S.21) is $O_P((\log p/n)^{1/2})$ while the bound in (S.23) is $O_P((1/n)^{1/2})$, there exists some constant $D_4$ such that $I_2 \le D_4 I_1$. Consequently, we conclude that $I$ is upper bounded by

$$I \quad\le\quad (1+D_4)\|\mathbf{\Omega}_k^*\|_\infty\sqrt{\frac{4}{D}}\sum_{k'=1}^{K}\left(\|\boldsymbol{\mu}_{k'}^*\|_\infty + (\|\mathbf{\Sigma}_{k'}^*\|_{\max})^{1/2}\right)\sqrt{\frac{\log(e/\delta)+\log p}{n}},$$

with probability at least $1-(2K+1)\delta$. For simplicity, let

$$\varphi_K = \sum_{k'=1}^{K}\left(\|\boldsymbol{\mu}_{k'}^*\|_\infty + (\|\mathbf{\Sigma}_{k'}^*\|_{\max})^{1/2}\right),\ C_1 = \sqrt{\frac{4(1+D_4)^2}{D}}. \tag{S.24}$$

Applying union bound,

$$\max_{k\in[K]} I \le C_1\,\|\mathbf{\Omega}^*\|_\infty\,\varphi_K\sqrt{\frac{\log p+\log(e/\delta)}{n}}, \tag{S.25}$$

with probability at least $1-K(2K+1)\delta$.

**Bounding Statistical Error for $k$-th Precision Matrix:** Referring to the proof in Lemma 3.1,

$$\begin{aligned} h_{\mathbf{\Omega}_k^*}(\mathbf{\Theta}^*) \quad &= \quad \frac{1}{2n}\sum_{i=1}^{n} L_{\mathbf{\Theta},k}(\boldsymbol{x}_i)\mathbf{\Sigma}_k^* - \frac{1}{2n}\sum_{i=1}^{n} L_{\mathbf{\Theta},k}(\boldsymbol{x}_i)(\boldsymbol{x}_i-\boldsymbol{\mu}_k^*)(\boldsymbol{x}_i-\boldsymbol{\mu}_k^*)^\top \\ &\quad - \frac{1}{2}\mathbb{E}\left[L_{\mathbf{\Theta},k}(\boldsymbol{X})\right]\mathbf{\Sigma}_k^* + \frac{1}{2}\mathbb{E}\left[L_{\mathbf{\Theta},k}(\boldsymbol{X})(\boldsymbol{X}-\boldsymbol{\mu}_k^*)(\boldsymbol{X}-\boldsymbol{\mu}_k^*)^\top\right]. \end{aligned}$$

Now we get an explicit from for $h_{\mathbf{\Omega}_k^*}(\mathbf{\Theta}^*)$. Then $II$ is decomposed as below:

$$\begin{aligned} II \quad &\le \quad \underbrace{\left\|\frac{1}{2}\left(\frac{1}{n}\sum_{i=1}^{n} L_{\mathbf{\Theta},k}(\boldsymbol{x}_i)\mathbf{\Sigma}_k^* - \mathbb{E}\left[L_{\mathbf{\Theta},k}(\boldsymbol{X})\mathbf{\Sigma}_k^*\right]\right)\right\|_{\max}}_{II_1} \\ &\quad + \underbrace{\left\|\frac{1}{2}\left(\frac{1}{n}\sum_{i=1}^{n} L_{\mathbf{\Theta},k}(\boldsymbol{x}_i)(\boldsymbol{x}_i-\boldsymbol{\mu}_k^*)(\boldsymbol{x}_i-\boldsymbol{\mu}_k^*)^\top - \mathbb{E}\left[L_{\mathbf{\Theta},k}(\boldsymbol{X})(\boldsymbol{X}-\boldsymbol{\mu}_k^*)(\boldsymbol{X}-\boldsymbol{\mu}_k^*)^\top\right]\right)\right\|_{\max}}_{II_2}. \end{aligned}$$

The first term is easy to deal with: since $\frac{1}{n}\sum_{i=1}^{n} L_{\mathbf{\Theta},k}(\boldsymbol{x}_i) - \mathbb{E}\left[L_{\mathbf{\Theta},k}(\boldsymbol{X})\right]$ is scalar by the definition of $L_{\mathbf{\Theta},k}(\boldsymbol{X})$ we can pull it out of the norm. Combining with the result in (S.22), the first term is upper bounded by

$$II_1 \le \|\mathbf{\Sigma}_k^*\|_{\max}\sqrt{\frac{1}{2}\log\frac{2}{\delta}}\cdot\sqrt{\frac{1}{n}}, \tag{S.26}$$

with probability at least $1-\delta$.

For the second term $II_2$, it can be decomposed as four following terms:

$$
\begin{aligned}
II_2 \quad \le \quad & \underbrace{\left\| \frac{1}{2}\left( \frac{1}{n}\sum_{i=1}^{n} L_{\boldsymbol{\Theta},k}(\boldsymbol{x}_i)\boldsymbol{x}_i\boldsymbol{x}_i^\top - \mathbb{E}\left[ L_{\boldsymbol{\Theta},k}(\boldsymbol{X})\boldsymbol{X}\boldsymbol{X}^\top \right] \right) \right\|_{\max}}_{II_{21}} \\
+ \; & \underbrace{\left\| \frac{1}{2}\left( \frac{1}{n}\sum_{i=1}^{n} L_{\boldsymbol{\Theta},k}(\boldsymbol{x}_i)\boldsymbol{x}_i\boldsymbol{\mu}_k^{*T} - \mathbb{E}\left[ L_{\boldsymbol{\Theta},k}(\boldsymbol{X})\boldsymbol{X}\boldsymbol{\mu}_k^{*\top} \right] \right) \right\|_{\max}}_{II_{22}} \\
+ \; & \underbrace{\left\| \frac{1}{2}\left( \frac{1}{n}\sum_{i=1}^{n} L_{\boldsymbol{\Theta},k}(\boldsymbol{x}_i)\boldsymbol{\mu}_k^{*}\boldsymbol{x}_i^\top - \mathbb{E}\left[ L_{\boldsymbol{\Theta},k}(\boldsymbol{X})\boldsymbol{\mu}_k^{*}\boldsymbol{X}^\top \right] \right) \right\|_{\max}}_{II_{23}} \\
+ \; & \underbrace{\left\| \frac{1}{2}\left( \frac{1}{n}\sum_{i=1}^{n} L_{\boldsymbol{\Theta},k}(\boldsymbol{x}_i)\boldsymbol{\mu}_k^{*}\boldsymbol{\mu}_k^{*\top} - \mathbb{E}\left[ L_{\boldsymbol{\Theta},k}(\boldsymbol{X})\boldsymbol{\mu}_k^{*}\boldsymbol{\mu}_k^{*\top} \right] \right) \right\|_{\max}}_{II_{24}} .
\end{aligned}
$$

For the bound of $II_{22}$ and $II_{23}$, we can just simply pull the $\boldsymbol{\mu}_k^*$ out, which implies

$$
\begin{aligned}
II_{22} \quad &= \quad \left\| \frac{1}{2}\left( \frac{1}{n}\sum_{i=1}^{n} L_{\boldsymbol{\Theta},k}(\boldsymbol{x}_i)\boldsymbol{x}_i - \mathbb{E}\left[ L_{\boldsymbol{\Theta},k}(\boldsymbol{X})\boldsymbol{X} \right] \right) \boldsymbol{\mu}_k^{*\top} \right\|_{\max} \qquad \text{(S.27)} \\
&\le \quad \left\| \frac{1}{2}\left( \frac{1}{n}\sum_{i=1}^{n} L_{\boldsymbol{\Theta},k}(\boldsymbol{x}_i)\boldsymbol{x}_i - \mathbb{E}\left[ L_{\boldsymbol{\Theta},k}(\boldsymbol{X})\boldsymbol{X} \right] \right) \right\|_{\infty} \|\boldsymbol{\mu}_k^*\|_\infty \\
&\overset{(a)}{\le} \quad \sqrt{\frac{4}{D}}\, \|\boldsymbol{\mu}_k^*\|_\infty \, \varphi_K \sqrt{\frac{\log(e/\delta)+\log p}{n}},
\end{aligned}
$$

with probability at least $1-2K\delta$, where $(a)$ follows (S.21).

Next we turn to bound $II_{21}$. Expand $\boldsymbol{x}_i\boldsymbol{x}_i^\top$ to matrix form for convenient use

$$
\boldsymbol{x}_i\boldsymbol{x}_i^\top = \begin{pmatrix} x_{i1}x_{i1} & \dots & x_{i1}x_{ip} \\ \vdots & \ddots & \vdots \\ x_{ip}x_{i1} & \dots & x_{ip}x_{ip} \end{pmatrix}.
$$

Since we require a matrix max norm here, it suffices to bound $II_{21}$ individually, namely

$$
\zeta_{jj'} = \frac{1}{2}\left( \frac{1}{n}\sum_{i=1}^{n} L_{\boldsymbol{\Theta},k}(\boldsymbol{x}_i)x_{ij}x_{ij'} - \mathbb{E}\left[ L_{\boldsymbol{\Theta},k}(\boldsymbol{X})X_jX_{j'} \right] \right).
$$

Recall in (S.18) the $j$-th coordinate of $\boldsymbol{x}_i$ could be expressed as

$$
x_{ij} = \sum_{k'=1}^{K} I\{c_i = k'\}(\mu_{k'j}^* + V_{k'j}).
$$

By straightforward algebra,

$$\begin{aligned} x_{ij}x_{ij'} &= \sum_{k'=1}^{K} I\{c_i = k'\}(\mu^*_{k'j} + V_{k'j}) \cdot \sum_{k'=1}^{K} I\{c_i = k'\}(\mu^*_{k'j'} + V_{k'j'}) \\ &\overset{(a)}{=} \sum_{k'=1}^{K} I\{c_i = k'\}^2(\mu^*_{k'j} + V_{k'j})(\mu^*_{k'j'} + V_{k'j'}) \\ &= \sum_{k'=1}^{K} I\{c_i = k'\}\left(\mu^*_{k'j}\mu^*_{k'j'} + \mu^*_{k'j}V_{k'j'} + V_{k'j}\mu^*_{k'j'} + V_{k'j}V_{k'j'}\right), \end{aligned}$$

where $(a)$ follows the fact that $I\{c_i = k\}I\{c_i = k'\} = 0$ for any $k \neq k'$. Consequently, we divide $\zeta_{jj'}$ into four parts:

$$\zeta_{jj'} = \frac{1}{2}\sum_{k'=1}^{K}\left(\zeta_{jj'}(\mu^*_{k'j}\mu^*_{k'j'}) + \zeta_{jj'}(\mu^*_{k'j}V_{k'j'}) + \zeta_{jj'}(V_{k'j}\mu^*_{k'j'}) + \zeta_{jj'}(V_{k'j}V_{k'j'})\right),$$

where

$$\begin{aligned} \zeta_{jj'}(\mu^*_{k'j}\mu^*_{k'j'}) &= \frac{1}{n}\sum_{i=1}^{n} L_{\boldsymbol{\Theta},k}(\boldsymbol{x}_i)I\{c_i = k'\}\mu^*_{k'j}\mu^*_{k'j'} \\ &\quad -\mathbb{E}\left[L_{\boldsymbol{\Theta},k}(\boldsymbol{X})I\{c = k'\}\mu^*_{k'j}\mu^*_{k'j'}\right]. \end{aligned}$$

Taking the supreme over set $[p]$ in terms of $p, p'$,

$$\begin{aligned} \sup_{j,j'\in[p]}|\zeta_{jj'}| \leq\ & \sum_{k'=1}^{K}\underbrace{\left(\sup_{j,j'\in[p]}|\zeta_{jj'}(\mu^*_{k'j}\mu^*_{k'j'})|\right)}_{(i)} + \sum_{k'=1}^{K}\underbrace{\left(\sup_{j,j'\in[p]}|\zeta_{jj'}(\mu^*_{k'j}V_{k'j'})|\right)}_{(ii)} \\ & + \sum_{k'=1}^{K}\underbrace{\left(\sup_{j,j'\in[p]}|\zeta_{jj'}(V_{k'j}\mu^*_{k'j'})|\right)}_{(iii)} + \sum_{k'=1}^{K}\underbrace{\left(\sup_{j,j'\in[p]}|\zeta_{jj'}(V_{k'j}V_{k'j'})|\right)}_{(iv)}. \end{aligned}$$

We will bound $(i)$, $(ii)$, $(iii)$ and $(iv)$ sequentially. $L_{\boldsymbol{\Theta},k}(\boldsymbol{x}_i)I\{c_i = k'\}\mu^*_{k'j}\mu^*_{k'j'}$ is a sub-gaussian random variable with

$$\|L_{\boldsymbol{\Theta},k}(\boldsymbol{x}_i)I\{c_i = k'\}\mu^*_{k'j}\mu^*_{k'j'}\|_{\psi_2} \leq \|\boldsymbol{\mu}^*_{k'}\|^2_\infty.$$

According to supporting Lemma S.3,

$$\left\|L_{\boldsymbol{\Theta},k}(\boldsymbol{x}_i)I\{c_i = k'\}\mu^*_{k'j}\mu^*_{k'j'} - \mathbb{E}[L_{\boldsymbol{\Theta},k}(\boldsymbol{X})I\{c = k'\}\mu^*_{k'j}\mu^*_{k'j'}]\right\|_{\psi_2} \leq 2\|\boldsymbol{\mu}^*_{k'}\|^2_\infty.$$

Applying concentration inequality in supporting Lemma S.4 yields that

$$\mathbb{P}\left(|\zeta_{jj'}(\mu^*_{k'j}\mu^*_{k'j'})| \leq t\right) \geq 1 - e\exp\left(-\frac{D_4 n t^2}{4\|\boldsymbol{\mu}^*_{k'}\|^4_\infty}\right), \tag{S.28}$$

for any $t > 0$ and some constant $D_4$. After properly choosing $t$,

$$(i) \leq \sqrt{\frac{4}{D_4}} \|\boldsymbol{\mu}_{k'}^*\|_\infty^2 \sqrt{\frac{\log p + \log(e/\delta)}{n}}, \tag{S.29}$$

with probability at least $1 - \delta$. Note that both $L_{\boldsymbol{\Theta},k}(\boldsymbol{x}_i) I\{c_i = k'\} \mu_{k'j}^* V_{k'j'}$ and $L_{\boldsymbol{\Theta},k}(\boldsymbol{x}_i) I\{c_i = k'\} V_{k'j'} \mu_{k'j}^*$ are sub-exponential random variables with norm $\|\boldsymbol{\mu}_{k'}^*\|_\infty (\|\boldsymbol{\Sigma}_{k'}^*\|_{\max})^{1/2}$. Similar to the step in (S.20),

$$\left|\zeta_{jj'}(\mu_{k'j}^* V_{k'j'})\right| \leq \sqrt{\frac{4}{D_5}} \left(\|\boldsymbol{\mu}_{k'}^*\|_\infty (\|\boldsymbol{\Sigma}_{k'}^*\|_{\max})^{1/2}\right) \sqrt{\frac{\log(2/\delta)}{n}},$$

with at least probability $1 - \delta$. Taking the union bound, it is shown that

$$(ii), (iii) \leq \sqrt{\frac{4}{D_5}} \left(\|\boldsymbol{\mu}_{k'}^*\|_\infty (\|\boldsymbol{\Sigma}_{k'}^*\|_{\max})^{1/2}\right) \sqrt{\frac{\log p + \log(2/\delta)}{n}}, \tag{S.30}$$

with probability at least $1 - \delta$ for sufficient large $n$.

Lastly, the fact that both $L_{\boldsymbol{\Theta},k}(\boldsymbol{x}_i) I\{c_i = k'\} V_{k'j}$ and $V_{k'j'}$ are sub-gaussian random variables implies $L_{\boldsymbol{\Theta},k}(\boldsymbol{x}_i) I\{c_i = k'\} V_{k'j} V_{k'j'}$ is sub-exponential random variable with parameter $\|\boldsymbol{\Sigma}_{k'}^*\|_{\max}$. Applying concentration result, there exists some constant $D_6$ such that the following inequality

$$\mathbb{P}\left(|\zeta_{jj'}(V_{k'j} V_{k'j'})| \geq t\right) \leq 2 \exp\left(-\frac{D_6 n t^2}{4\|\boldsymbol{\Sigma}_{k'}^*\|_{\max}^2}\right),$$

holds for sufficiently small $t > 0$. Therefore,

$$\mathbb{P}\left(\sup_{j,j' \in [p]} |\zeta_{jj'}(V_{k'j} V_{k'j'})| \geq t\right) \leq 2p^2 \exp\left(-\frac{D_6 n t^2}{4\|\boldsymbol{\Sigma}_{k'}^*\|_{\max}^2}\right).$$

When $n$ is sufficiently large, with probability at least $1 - \delta$

$$(iv) \leq \sqrt{\frac{4}{D_6}} \|\boldsymbol{\Sigma}_{k'}^*\|_{\max} \sqrt{\frac{2\log p + \log(2/\delta)}{n}}. \tag{S.31}$$

Putting (S.29), (S.30) and (S.31) together and after some adjustments, $II_{21}$ is upper bounded by

$$II_{21} \leq \sqrt{\frac{1}{D_7}} \sum_{k'=1}^{K} \left(\|\boldsymbol{\mu}_{k'}^*\|_\infty + (\|\boldsymbol{\Sigma}_{k'}^*\|_{\max})^{1/2}\right)^2 \sqrt{\frac{2\log p + \log(e/\delta)}{n}},$$

with probability at least $1 - 4K\delta$. $D_7 = \min(D_4, D_5, D_6)$. For simplicity, we denote

$$\varphi_K' = \sum_{k'=1}^{K} \left(\|\boldsymbol{\mu}_{k'}^*\|_\infty + (\|\boldsymbol{\Sigma}_{k'}^*\|_{\max})^{1/2}\right)^2.$$

Therefore,

$$II_{21} \leq \sqrt{\frac{2}{D_7}} \varphi_K' \sqrt{\frac{\log p + \log(e/\delta)}{n}}, \tag{S.32}$$

with probability at least $1 - 4K\delta$.

For the last, it remains to bound $II_{24}$. Recall that

$$\begin{aligned} II_{24} &= \left\| \frac{1}{2}\left( \frac{1}{n}\sum_{i=1}^{n} L_{\boldsymbol{\Theta},k}(\boldsymbol{x}_i)\boldsymbol{\mu}_k^*\boldsymbol{\mu}_k^{*\top} - \mathbb{E}\left[ L_{\boldsymbol{\Theta},k}(\boldsymbol{X})\boldsymbol{\mu}_k^*\boldsymbol{\mu}_k^{*\top}\right]\right)\right\|_{\max} \\ &\leq \left| \frac{1}{2}\left( \frac{1}{n}\sum_{i=1}^{n} L_{\boldsymbol{\Theta},k}(\boldsymbol{x}_i) - \mathbb{E}\left[L_{\boldsymbol{\Theta},k}(\boldsymbol{X})\right]\right)\right| \|\boldsymbol{\mu}_k^*\boldsymbol{\mu}_k^{*\top}\|_{\max}. \end{aligned}$$

Applying the result in (S.22), we have

$$II_{24} \leq \|\boldsymbol{\mu}_k^*\boldsymbol{\mu}_k^{*\top}\|_{\max}\sqrt{\frac{1}{2}\log\frac{2}{\delta}}\cdot\sqrt{\frac{1}{n}}, \tag{S.33}$$

with probability at least $1 - \delta$.

Putting (S.27), (S.32) and (S.33) together, now we can have a upper bound for $II_2$.

$$II_2 \;\leq\; \sqrt{\frac{1}{D_7}}\left(2\|\boldsymbol{\mu}_k^*\|_\infty\varphi_K + \varphi_K'\right)\sqrt{\frac{\log p + \log(e/\delta)}{n}}, \tag{S.34}$$

for $D_7 < D/2$ with at least probability $1 - (8K+1)\delta$. The upper bound in (S.26) is of order $O_P(n^{-1/2})$ while the upper bound in (S.34) is of order $O_P((\log p/n)^{1/2})$. Thus there exists some constant $D_8$ such that $II_1 \leq D_8 II_2$. Let $C_2 = ((1+D_8)^2/D_7)^{1/2}$. Applying union bound,

$$\max_{k\in[K]} II \;\leq\; C_2\left(2\|\boldsymbol{\mu}^*\|_\infty\varphi_K + \varphi_K'\right)\sqrt{\frac{\log p + \log(e/\delta)}{n}}, \tag{S.35}$$

with at least probability $1 - K(8K+2)\delta$.

**Bound the Group Structure Part of Precision Matrix:**

Recall that

$$\begin{aligned} III &= \max_{i,j}\Big\| \left[\nabla_{\boldsymbol{\Omega}_1^*}Q_n(\boldsymbol{\Theta}^*|\boldsymbol{\Theta}) - \nabla_{\boldsymbol{\Omega}_1^*}Q(\boldsymbol{\Theta}^*|\boldsymbol{\Theta})\right]_{ij}, \\ &\qquad\qquad \ldots, \left[\nabla_{\boldsymbol{\Omega}_K^*}Q_n(\boldsymbol{\Theta}^*|\boldsymbol{\Theta}) - \nabla_{\boldsymbol{\Omega}_K^*}Q(\boldsymbol{\Theta}^*|\boldsymbol{\Theta})\right]_{ij}\Big\|_2 \\ &\leq \max_{i,j}\sqrt{K}\Big\| \left[\nabla_{\boldsymbol{\Omega}_1^*}Q_n(\boldsymbol{\Theta}^*|\boldsymbol{\Theta}) - \nabla_{\boldsymbol{\Omega}_1^*}Q(\boldsymbol{\Theta}^*|\boldsymbol{\Theta})\right]_{ij}, \\ &\qquad\qquad \ldots, \left[\nabla_{\boldsymbol{\Omega}_K^*}Q_n(\boldsymbol{\Theta}^*|\boldsymbol{\Theta}) - \nabla_{\boldsymbol{\Omega}_K^*}Q(\boldsymbol{\Theta}^*|\boldsymbol{\Theta})\right]_{ij}\Big\|_\infty \\ &\leq \sqrt{K}\max_{k\in[K]}\left\|\left[\nabla_{\boldsymbol{\Omega}_k^*}Q_n(\boldsymbol{\Theta}^*|\boldsymbol{\Theta}) - \nabla_{\boldsymbol{\Omega}_k^*}Q(\boldsymbol{\Theta}^*|\boldsymbol{\Theta})\right]\right\|_{\max}. \end{aligned}$$

According to the result in (S.35) and applying union bound over $[K]$,

$$\mathbb{P}\left( III \geq C_2\sqrt{K}\left(2\|\boldsymbol{\mu}_k^*\|_\infty\varphi_K + \varphi_K'\right)\sqrt{\frac{\log p + \log(e/\delta)}{n}}\right) \leq K(8K+2)\delta.$$

Thus, $III$ is upper bounded by

$$III \quad \leq \quad C_2\sqrt{K}\left(2\|\boldsymbol{\mu}^*\|_\infty\varphi_K + \varphi_K'\right)\sqrt{\frac{\log p + \log(e/\delta)}{n}}, \tag{S.36}$$

with at least probability $1 - K(8K+2)\delta$.

Finally, putting the upper bound (S.25), (S.35) and (S.36) together, we have a upper bound for the following statistical error

$$\begin{aligned} & \left\|\nabla Q_n(\boldsymbol{\Theta}^*|\boldsymbol{\Theta}) - \nabla Q(\boldsymbol{\Theta}^*|\boldsymbol{\Theta})\right\|_{\mathcal{P}^*} \\ \leq \quad & C\left((\|\boldsymbol{\Omega}^*\|_\infty + (\sqrt{K}+1)\|\boldsymbol{\mu}^*\|_\infty)\varphi_K + 2(\sqrt{K}+1)\varphi_K'\right)\sqrt{\frac{\log p + \log(e/\delta)}{n}}, \end{aligned}$$

with probability at least $1-(18K+6)\delta$, where $C = \max(M_1C_1, M_2C_2, M_3C_3)$. Under regularity Condition 3.12, $\varphi_K \leq (c_1 + c_2^{1/2})K$, $\varphi_K' \leq (c_1 + c_2^{1/2})^2 K$. Let $C = C(c_1 + c_2^{1/2})$ and $C' = c_1^2 + c_1 c_2^{1/2} + 2(c_1 + c_2^{1/2})^2$. Consequently, the upper bound for statistical error can be written as:

$$\|\nabla Q_n(\boldsymbol{\Theta}^*|\boldsymbol{\Theta}) - \nabla Q(\boldsymbol{\Theta}^*|\boldsymbol{\Theta})\|_{\mathcal{P}^*} \leq \left(CK\|\boldsymbol{\Omega}^*\|_\infty + C'K^{1.5}\right)\sqrt{\frac{\log p + \log(e/\delta)}{n}},$$

with probability at least $1-(18K+6)\delta$. ■

For the second part of Lemma S.1, we are aiming to bound the statistical error arising from the estimation for diagonal term. The definition of $\mathcal{G}$ in (3.1) implies that $[\nabla Q_n(\boldsymbol{\Theta}^*|\boldsymbol{\Theta}) - \nabla Q(\boldsymbol{\Theta}^*|\boldsymbol{\Theta})]_{\mathcal{G}}$ is a $Kp$-dimensional vector. Following the same derivation before, it suffices to have:

$$\begin{aligned} & \left\|[\nabla Q_n(\boldsymbol{\Theta}^*|\boldsymbol{\Theta}) - \nabla Q(\boldsymbol{\Theta}^*|\boldsymbol{\Theta})]_{\mathcal{G}}\right\|_2 \\ \leq \quad & \sqrt{Kp}\left\|[\nabla Q_n(\boldsymbol{\Theta}^*|\boldsymbol{\Theta}) - \nabla Q(\boldsymbol{\Theta}^*|\boldsymbol{\Theta})]_{\mathcal{G}}\right\|_{\max} \\ \overset{(a)}{\leq} \quad & \sqrt{Kp}\cdot C_2\left(2\|\boldsymbol{\mu}^*\|_\infty\varphi_K + \varphi_K'\right)\sqrt{\frac{\log p + \log(e/\delta)}{n}} \\ = \quad & \sqrt{K}\cdot C_2\left(2\|\boldsymbol{\mu}^*\|_\infty\varphi_K + \varphi_K'\right)\sqrt{\frac{p(\log p + \log(e/\delta))}{n}}, \end{aligned}$$

with probability at least $1-(8K^2+2K)\delta$ where $(a)$ comes from (S.36). Now combining two parts together, we end the proof of Lemma S.1. ■

## S.V Proof of Lemma A.2

For any $\boldsymbol{\Theta} \in \mathcal{M}$,

$$\begin{aligned} \frac{\mathcal{P}(\boldsymbol{\Theta})}{\|\boldsymbol{\Theta}\|_2} \quad & = \quad \frac{\mathcal{P}_1(\boldsymbol{\Theta})}{\|\boldsymbol{\Theta}\|_2} + \frac{\mathcal{P}_2(\boldsymbol{\Theta})}{\|\boldsymbol{\Theta}\|_2} + \frac{\mathcal{P}_3(\boldsymbol{\Theta})}{\|\boldsymbol{\Theta}\|_2} \\ & \leq \quad \frac{M_1\sum_{k=1}^K\sum_{j=1}^p|\mu_{kj}|}{\sqrt{\sum_{k=1}^K\|\boldsymbol{\mu}_k\|_2^2}} + \frac{M_2\sum_{k=1}^K\sum_{i\neq j}|\omega_{kij}|}{\sqrt{\sum_{k=1}^K\|\boldsymbol{\Omega}_k\|_F^2}} + \frac{\sum_{i\neq j}M_3(\sum_{k=1}^K\omega_{kij}^2)^{1/2}}{\sqrt{\sum_{k=1}^K\|\boldsymbol{\Omega}_k\|_F^2}}. \end{aligned}$$

By Cauchy's inequality, we can have

$$\frac{\mathcal{P}(\boldsymbol{\Theta}_{\mathcal{M}})}{\|\boldsymbol{\Theta}_{\mathcal{M}}\|_2} \leq M_1\sqrt{Kd} + M_2\sqrt{Ks} + M_3\sqrt{s}.$$

Recall that $d$ and $s$ are the sparse parameter for a single cluster mean and precision matrix, respectively. This ends the proof of Lemma A.2. ■

## S.VI Proof of Lemma B.1

First we consider each $\boldsymbol{\Theta}_k = \{\mu_k, \boldsymbol{\Omega}_k\}$ individually. That means we prove the following part first:

$$Q_n(\boldsymbol{\Theta}_k^{(1)}|\boldsymbol{\Theta}^{(t-1)}) - Q_n(\boldsymbol{\Theta}_k^{(2)}|\boldsymbol{\Theta}^{(t-1)}) - \left\langle \nabla_{\boldsymbol{\Theta}_k} Q_n(\boldsymbol{\Theta}_k^{(2)}|\boldsymbol{\Theta}^{(t-1)}), \boldsymbol{\Theta}_k^{(1)} - \boldsymbol{\Theta}_k^{(2)} \right\rangle \leq 0,$$

where $Q_n(\boldsymbol{\Theta}_k|\boldsymbol{\Theta})$ means we set $\boldsymbol{\Theta}_i$ $i \neq k$ to zero.

Following the same technique we use in the proof of Lemma (3.5), the decomposition can be made as below:

$$Q_n(\boldsymbol{\Theta}_k^{(1)}|\boldsymbol{\Theta}^{(t-1)}) - Q_n(\boldsymbol{\Theta}_k^{(2)}|\boldsymbol{\Theta}^{(t-1)}) - \left\langle \nabla_{\boldsymbol{\Theta}_k} Q_n(\boldsymbol{\Theta}_k^{(2)}|\boldsymbol{\Theta}^{(t-1)}), \boldsymbol{\Theta}_k^{(1)} - \boldsymbol{\Theta}_k^{(2)} \right\rangle = I + II,$$

where

$$\begin{aligned} I &= \frac{1}{n}\sum_{i=1}^{n}\left[L_{\boldsymbol{\Theta},k}(\boldsymbol{x}_i)\left\{h(\boldsymbol{\mu}_k^{(2)}, \boldsymbol{\Omega}_k^{(2)}) - h(\boldsymbol{\mu}_k^{(1)}, \boldsymbol{\Omega}_k^{(2)})\right\}\right] \\ &-(\boldsymbol{\mu}_k^{(1)} - \boldsymbol{\mu}_k^{(2)})^\top \nabla_{\boldsymbol{\mu}_k} Q_n(\boldsymbol{\Theta}_k^{(2)}|\boldsymbol{\Theta}^{(t-1)}), \\ II &= \frac{1}{n}\sum_{i=1}^{n}\left[L_{\boldsymbol{\Theta},k}(\boldsymbol{x}_i)\{\frac{1}{2}\log\det(\boldsymbol{\Omega}_k^{(1)}) - \frac{1}{2}\log\det(\boldsymbol{\Omega}_k^{(2)})\right. \\ &\left.+h(\boldsymbol{\mu}_k^{(1)}, \boldsymbol{\Omega}_k^{(2)}) - h(\boldsymbol{\mu}_k^{(1)}, \boldsymbol{\Omega}_k^{(1)})\}\right] - [\mathrm{vec}(\boldsymbol{\Omega}_k^{(1)} - \boldsymbol{\Omega}_k^{(2)})]^\top \nabla_{\boldsymbol{\Omega}_k} Q_n(\boldsymbol{\Theta}_k^{(2)}|\boldsymbol{\Theta}^{(t-1)}). \end{aligned}$$

**Bounding $I$:** By a little algebra, we can show that

$$I = -\frac{1}{2n}\sum_{i=1}^{n} L_{\boldsymbol{\Theta},k}(\boldsymbol{x}_i)(\boldsymbol{\mu}_k^{(1)} - \boldsymbol{\mu}_k^{(2)})^\top \boldsymbol{\Omega}_k^{(2)}(\boldsymbol{\mu}_k^{(1)} - \boldsymbol{\mu}_k^{(2)}).$$

Plugging in $(\boldsymbol{\Theta}^{(t)}, t^*\boldsymbol{\Theta}^{(t)} + (1-t^*)\boldsymbol{\Theta}^*)$, we have

$$I = -\frac{(1-t^*)^2}{2n}\sum_{i=1}^{n} L_{\boldsymbol{\Theta},k}(\boldsymbol{x}_i)(\boldsymbol{\mu}_k^{(t)} - \boldsymbol{\mu}_k^*)^\top \left(t^*\boldsymbol{\Omega}_k^{(t)} + (1-t^*)\boldsymbol{\Omega}_k^*\right)(\boldsymbol{\mu}_k^{(t)} - \boldsymbol{\mu}_k^*).$$

Recall that $\boldsymbol{\Theta}^{(t)}$ is the solution of the optimization problem (B.2). The algorithm guarantees that $\boldsymbol{\Omega}_k^{(t)}$ is positive definite. Thus, from the positive definiteness of $\boldsymbol{\Omega}_k^{(t)}$ and $\boldsymbol{\Omega}_k^*$, it is sufficient to show that

$$I \leq 0 \quad \text{holds a.s..} \tag{S.37}$$

When plugging in $(\boldsymbol{\Theta}^*, t^*\boldsymbol{\Theta}^{(t)} + (1-t^*)\boldsymbol{\Theta}^*)$, we have the same conclusion.

**Bounding $II$:** Define

$$g(\boldsymbol{\Omega}_k^{(2)}) := \frac{1}{n}\sum_{i=1}^{n}\left[L_{\boldsymbol{\Theta},k}(\boldsymbol{x}_i)\left\{\frac{1}{2}\log\det(\boldsymbol{\Omega}_k^{(2)}) - h\left(\boldsymbol{\mu}_k^{(1)}, \boldsymbol{\Omega}_k^{(2)}\right)\right\}\right].$$

We rewrite $II$ as

$$g(\boldsymbol{\Omega}_k^{(1)}) - g(\boldsymbol{\Omega}_k^{(2)}) - \langle \mathrm{vec}\left(\nabla g(\boldsymbol{\Omega}_k^{(2)})\right), \mathrm{vec}\left(\boldsymbol{\Omega}_k^{(1)} - \boldsymbol{\Omega}_k^{(2)}\right)\rangle.$$

According to Taylor expansion, we can expand $g(\boldsymbol{\Omega}_k^{(1)})$ around $\boldsymbol{\Omega}_k^{(2)}$ and obtain

$$\begin{aligned} g(\boldsymbol{\Omega}_k^{(1)}) &= g(\boldsymbol{\Omega}_k^{(2)}) + \langle \mathrm{vec}(\nabla g(\boldsymbol{\Omega}_k^{(2)}), \mathrm{vec}(\boldsymbol{\Omega}_k^{(1)} - \boldsymbol{\Omega}_k^{(2)})\rangle \\ &\quad + \frac{1}{2}\left[\mathrm{vec}(\boldsymbol{\Omega}_k^{(1)} - \boldsymbol{\Omega}_k^{(2)})\right]^\top \nabla^2 g(\mathbf{Z})\left[\mathrm{vec}(\boldsymbol{\Omega}_k^{(1)} - \boldsymbol{\Omega}_k^{(2)})\right], \end{aligned}$$

where $\mathbf{Z} = t\boldsymbol{\Omega}_k^{(1)} + (1-t)\boldsymbol{\Omega}_k^{(2)}$ with $t \in [0,1]$. So an equivalent expression for $II$ is given below:

$$II = \frac{1}{2}\left[\mathrm{vec}(\boldsymbol{\Omega}_k^{(1)} - \boldsymbol{\Omega}_k^{(2)})\right]^\top \nabla^2 g(\mathbf{Z})\left[\mathrm{vec}(\boldsymbol{\Omega}_k^{(1)} - \boldsymbol{\Omega}_k^{(2)})\right].$$

By the definition of function $g$ we construct, the negative Hessian matrix of function $g$ is

$$-\nabla^2 g(\mathbf{Z}) = \frac{1}{2n}\sum_{i=1}^{n} L_{\boldsymbol{\Theta},k}(\boldsymbol{x}_i)\mathbf{Z}^{-1} \otimes \mathbf{Z}^{-1}.$$

According to the analysis in the proof of Lemma 3.5, $\sigma_{\min}\left(\mathbf{Z}^{-1} \otimes \mathbf{Z}^{-1}\right) = \left[\sigma_{\min}(\mathbf{Z}^{-1})\right]^2 \geq 0$. Therefore, $\nabla^2 g(\mathbf{Z})$ is a negative semi-definite matrix, which implies that $II \leq 0$ holds a.s. for any pair of points $(\boldsymbol{\Theta}^{(1)}, \boldsymbol{\Theta}^{(2)})$. Incorporating with the fact that $I < 0$, it implies that

$$Q_n(\boldsymbol{\Theta}_k^{(1)}|\boldsymbol{\Theta}^{(t-1)}) - Q_n(\boldsymbol{\Theta}_k^{(2)}|\boldsymbol{\Theta}^{(t-1)}) - \left\langle \nabla_{\boldsymbol{\Theta}_k} Q_n(\boldsymbol{\Theta}_k^{(2)}|\boldsymbol{\Theta}^{(t-1)}), \boldsymbol{\Theta}_k^{(1)} - \boldsymbol{\Theta}_k^{(2)}\right\rangle \leq 0,$$

holds a.s. for pair points $(\boldsymbol{\Theta}^{(t)}, t^*\boldsymbol{\Theta}^{(t)} + (1-t^*)\boldsymbol{\Theta}^*)$, $(\boldsymbol{\Theta}^{(t)}, t^*\boldsymbol{\Theta}^{(t)} + (1-t^*)\boldsymbol{\Theta}^*)$. After doing the summation from 1 to $K$, we finish the proof of Lemma B.1. ■

## S.VII Variable Selection Consistency

**Theorem S.2.** Denote the final precision matrix estimator as $\widetilde{\boldsymbol{\Omega}}_k$ and the set of its nonzero off-diagonal elements as $\widetilde{\mathcal{V}}_k$. Under minimal signal condition, we have, with probability tending to 1, $\widetilde{\mathcal{V}}_k = \mathcal{V}_k$ for any $k = 1, \ldots, K$.

*Proof:* We prove it in two steps. In Step 1, we show that $\widetilde{\mathcal{V}_k} \supset \mathcal{V}_k$, and in Step 2, we show that $\widetilde{\mathcal{V}_k} \subset \mathcal{V}_k$, both with high probability.

*Step 1:* In order to prove $\widetilde{\mathcal{V}_k} \supset \mathcal{V}_k$, it is sufficient to show that for any $(i,j) \in \mathcal{V}_k$ with any $k = 1, \ldots, K$, $\widetilde{\omega}_{kij} \neq 0$. Note that

$$|\omega_{kij}^{(T)}| \geq |\omega_{kij}^*| - |\omega_{kij}^{(T)} - \omega_{kij}^*| \geq |\omega_{kij}^*| - \sqrt{\sum_{i,j} (\omega_{kij}^{(T)} - \omega_{kij}^*)^2},$$

Moreover,

$$\sqrt{\sum_{i,j} (\omega_{kij}^{(T)} - \omega_{kij}^*)^2} \leq \|\mathbf{\Theta}^{(T)} - \mathbf{\Theta}^*\|_2. \tag{S.38}$$

According to Corollary 3.14 and minimal signal condition we have

$$|\omega_{kij}^{(T)}| > r_n.$$

Therefore, we see that $\widetilde{\omega}_{kij} \neq 0$, which implies $\widetilde{\mathcal{V}_k} \supset \mathcal{V}_k$.

*Step 2:* In order to show $\widetilde{\mathcal{V}_k} \subset \mathcal{V}_k$, we need to check that, for any $(i,j) \in \mathcal{V}_k^c$, the estimator $\widetilde{\omega}_{kij} = 0$. Note that, the estimator before the thresholding step satisfies,

$$|\omega_{kij}^{(T)}| = |\omega_{kij}^{(T)} - \omega_{kij}^*| \leq \sqrt{\sum_{i,j} (\omega_{kij}^{(T)} - \omega_{kij}^*)^2}.$$

From (S.38), it is known that $|\omega_{kij}^{(T)}| \leq r_n$. Therefore, the thresholding step will set $\widetilde{\omega}_{kij} = \omega_{kij}^{(T)} \mathbb{1}\{|\widehat{\omega}_{kij}| > r_n\} = 0$ with high probability. This ends the proof of Theorem S.2. ■

# B Updates steps of our SCAN algorithm

## S.I Proof of Lemma 2.2:

The KKT conditions for $\mu_{kj}$ to be a maximizer of $Q(\mathbf{\Theta}|\mathbf{\Theta}^{(t-1)}) - \mathcal{R}(\mathbf{\Theta})$ are

$$\frac{1}{n}\sum_{i=1}^{n} L_{\mathbf{\Theta}^{(t-1)},k}\Big(\sum_{l=1}^{p}(x_{il} - \mu_{kl})\omega_{klj}\Big) = \lambda_1 \mathrm{sign}(\mu_{kj}), \text{ when } \mu_{kj} \neq 0,$$

$$\left|\frac{1}{n}\sum_{i=1}^{n} L_{\mathbf{\Theta}^{(t-1)},k}\Big(\sum_{l=1,l\neq j}^{p}(x_{il} - \mu_{kl})\omega_{klj} + x_{ij}\omega_{kjj}\Big)\right| \leq \lambda_1, \text{ when } \mu_{kj} = 0.$$

Therefore, the update of $\mu_{kj}^{(t)}$ is given as:

$$\text{If } \left|\frac{1}{n}\sum_{i=1}^{n} L_{\mathbf{\Theta}^{(t-1)},k}(\boldsymbol{x}_i)\Big(\sum_{l=1,l\neq j}^{p}(x_{il} - \mu_{kl}^{(t-1)})\omega_{klj}^{(t-1)} + x_{ij}\omega_{kjj}^{(t-1)}\Big)\right| \leq \lambda_1,$$

then $\mu_{kj}^{(t)} = 0$; Else

$$\mu_{kj}^{(t)} = \Big(\omega_{kjj}^{(t-1)} \frac{1}{n}\sum_{i=1}^{n} L_{\boldsymbol{\Theta}^{(t-1)},k}(\boldsymbol{x}_i)\Big)^{-1}\Big\{\frac{1}{n}\sum_{i=1}^{n} L_{\boldsymbol{\Theta}^{(t-1)},k}(\boldsymbol{x}_i)\big(\sum_{l=1}^{p} x_{il}\omega_{klj}^{(t-1)}\big) -$$
$$\big(\frac{1}{n}\sum_{i=1}^{n} L_{\boldsymbol{\Theta}^{(t-1)},k}(\boldsymbol{x}_i)\big)\Big(\sum_{l=1}^{p}\mu_{kl}^{(t-1)}\omega_{klj}^{(t-1)} - \mu_{kj}^{(t-1)}\omega_{kjj}^{(t-1)}\Big) - \lambda_1 \mathrm{sign}(\mu_{kj}^{(t-1)})\Big\}$$

Using the definitions of $g_{1,j}(\boldsymbol{x}; \boldsymbol{\Theta}_k^{(t-1)})$ and $g_{2,j}(\boldsymbol{x}_i; \boldsymbol{\Theta}_k^{(t-1)})$, we finish the proof of Lemma 2.2. ■

## S.II Proof of Lemma 2.3:

Recall that in (2.7)

$$Q_n(\boldsymbol{\Theta}|\boldsymbol{\Theta}^{(t-1)}) := \frac{1}{n}\sum_{i=1}^{n}\sum_{k=1}^{K} L_{\boldsymbol{\Theta}^{(t-1)},k}(\boldsymbol{x}_i)[\log \pi_k + \log f_k(\boldsymbol{x}_i; \boldsymbol{\Theta}_k)] - \mathcal{R}(\boldsymbol{\Theta}),$$

Then,

$$\begin{aligned}
& \max_{\boldsymbol{\Omega}_1,\ldots,\boldsymbol{\Omega}_K} \frac{1}{n}\sum_{i=1}^{n}\sum_{k=1}^{K} L_{\boldsymbol{\Theta}^{(t-1)},k}(\boldsymbol{x}_i)[\log \pi_k + \log f_k(\boldsymbol{x}_i; \boldsymbol{\Theta}_k)] - \mathcal{R}(\boldsymbol{\Theta}) \\
= & \max_{\boldsymbol{\Omega}_1,\ldots,\boldsymbol{\Omega}_K} \frac{1}{n}\sum_{i=1}^{n}\sum_{k=1}^{K} L_{\boldsymbol{\Theta}^{(t-1)},k}(\boldsymbol{x}_i)[\log \pi_k - \frac{p}{2}\log(2\pi) + \frac{1}{2}\log\det(\boldsymbol{\Omega}_k) \\
& -\frac{1}{2}(\boldsymbol{x}_i - \boldsymbol{\mu}_k)^\top \boldsymbol{\Omega}_k(\boldsymbol{x}_i - \boldsymbol{\mu}_k)] - \frac{1}{2}\mathcal{R}(\boldsymbol{\Theta}) \\
= & \max_{\boldsymbol{\Omega}_1,\ldots,\boldsymbol{\Omega}_K} \frac{1}{n}\sum_{k=1}^{K}\Big\{\frac{1}{n}\sum_{i=1}^{n} L_{\boldsymbol{\Theta}^{(t-1)},k}(\boldsymbol{x}_i)[\log\det(\boldsymbol{\Omega}_k) - (\boldsymbol{x}_i - \boldsymbol{\mu}_k)^\top \boldsymbol{\Omega}_k(\boldsymbol{x}_i - \boldsymbol{\mu}_k)]\Big\} - \mathcal{R}(\boldsymbol{\Theta}) \\
= & \max_{\boldsymbol{\Omega}_1,\ldots,\boldsymbol{\Omega}_K} \frac{1}{n}\sum_{k=1}^{K} n_k[\log\det(\boldsymbol{\Omega}_k) - \mathrm{trace}(\widetilde{S}_k\boldsymbol{\Omega}_k)] - \mathcal{R}(\boldsymbol{\Theta}),
\end{aligned}$$

where the last equality is because

$$\begin{aligned}
& \frac{1}{n}\sum_{i=1}^{n} L_{\boldsymbol{\Theta}^{(t-1)},k}(\boldsymbol{x}_i)(\boldsymbol{x}_i - \boldsymbol{\mu}_k)^\top \boldsymbol{\Omega}_k(\boldsymbol{x}_i - \boldsymbol{\mu}_k) \\
= & \frac{1}{n}\sum_{\boldsymbol{x}_i\in\mathcal{A}_k} \mathrm{trace}((\boldsymbol{x}_i - \boldsymbol{\mu}_k)(\boldsymbol{x}_i - \boldsymbol{\mu}_k)^\top \boldsymbol{\Omega}_k) \\
= & \frac{1}{n}\mathrm{trace}\Big(\sum_{\boldsymbol{x}_i\in\mathcal{A}_k}(\boldsymbol{x}_i - \boldsymbol{\mu}_k)(\boldsymbol{x}_i - \boldsymbol{\mu}_k)^\top \boldsymbol{\Omega}_k\Big).
\end{aligned}$$

Then plugging in the last update of $\boldsymbol{\mu}_k$ leads to the desirable result. ■

## C Supporting Lemma

**Lemma S.1.** Consider a finite number of independent centered sub-gaussian random variables $X_i$. Then $\sum_i X_i$ is also a centered sub-gaussian random variable. Moreover,

$$\left\|\sum_i X_i\right\|_{\psi_2}^2 \leq C\sum_i \|X_i\|_{\psi_2}^2 ,$$

where $C$ is an absolute constant.

**Lemma S.2.** Let $X, Y$ be two sub-Gaussian random variables. Then $Z = X \cdot Y$ is sub-exponential random variable. Moreover, there exits constant $C$ such that

$$\|Z\|_{\psi_1} \leq C\|X\|_{\psi_2} \cdot \|Y\|_{\psi_2} . \tag{S.1}$$

**Lemma S.3.** Let $X$ be sub-Gaussian random variable and Y be sub-exponential random variables. Then $X - \mathbb{E}[X]$ is also sub-Gaussian; $Y - \mathbb{E}[Y]$ is also sub-exponential. Moreover, we have

$$\|X - \mathbb{E}[X]\|_{\psi_2} \leq 2\|X\|_{\psi_2} ,\ \|Y - \mathbb{E}[Y]\|_{\psi_1} \leq 2\|Y\|_{\psi_1} .$$

**Lemma S.4.** Suppose $X_1, X_2, \ldots, X_n$ are $n$ iid centered sub-Gaussian random variables with $\|X_1\|_{\psi_2} \leq K$. Then for every $t \geq 0$, we have

$$\mathbb{P}\left(\left|\frac{1}{n}\sum_{i=1}^n X_i\right| \geq t\right) \geq e \cdot \exp\left(-\frac{Cnt^2}{K^2}\right),$$

where C is an absolute constant.

**Lemma S.5.** Suppose $X_1, X_2, \ldots, X_n$ are $n$ iid centered sub-expoential random variables with $\|X_1\|_{\psi_1} \leq K$. Then for every $t \geq 0$, we have

$$\mathbb{P}\left(\left|\frac{1}{n}\sum_{i=1}^n X_i\right| \geq t\right) \geq 2 \cdot \exp\left(-C\min\left\{\frac{t^2}{K^2}, \frac{t}{K}\right\}n\right),$$

where C is an absolute constant.

**Lemma S.6.** Hoeffding's inequality Suppose $X_1, X_2 \ldots X_n$ are independent random variable, $a_1 \leq X_i \leq b_i$, then we can have

$$\mathbb{P}\left(\left|\frac{1}{n}\sum_{i=1}^n (X_i - \mathbb{E}X_i)\right| > \varepsilon\right) \leq 2\exp\left\{\frac{-2n\varepsilon^2}{\frac{1}{n}\sum_{i=1}^n (b_i - a_i)^2}\right\}.$$

Moreover, if $a_i = 0$ and $b_i = 1$, then we have

$$\mathbb{P}\left(\left|\frac{1}{n}\sum_{i=1}^n (X_i - \mathbb{E}X_i)\right| > \varepsilon\right) > 1 - 2e^{-2n\varepsilon^2}.$$